\documentclass[fleqn,10pt]{wlscirep}
\usepackage[utf8]{inputenc}
\usepackage[T1]{fontenc}
\usepackage{multirow}
\RequirePackage{graphicx}
\RequirePackage{fancyhdr}
\RequirePackage{bookmark}
\RequirePackage{caption} 
\usepackage{arydshln}

\usepackage{algorithm}
\usepackage{algpseudocode}
\usepackage{xr}

\usepackage{xcolor} 
\definecolor{docnotelinkcolor}{HTML}{0000FF} 

\usepackage{hyperref}
\hypersetup{colorlinks = true, allcolors = blue}

\usepackage[nameinlink]{cleveref}
\usepackage{subfiles}

\crefname{figure}{Fig.}{Figs.}
\crefformat{equation}{Eq.~#2(#1)#3}
\crefformat{section}{Section~#2#1#3}
\AtBeginDocument{%
	\let\citet\cite
}

\usepackage[labelfont=bf]{caption}
\usepackage{babel}

\title{NOWS: Neural Operator Warm Starts for Accelerating Iterative Solvers}

\author[1]{Mohammad Sadegh Eshaghi\thanks{Corresponding author. Email: eshaghi.khanghah@iop.uni-hannover.de}} 
\author[1]{Cosmin Anitescu}
\author[1]{Navid Valizadeh} 
\author[3]{Yizheng Wang} 
\author[1]{Xiaoying Zhuang\thanks{Corresponding author. Email: zhuang@iop.uni-hannover.de}}
\author[2]{Timon Rabczuk \thanks{Corresponding author. Email: timon.rabczuk@uni-weimar.de}}

\affil[1]{Chair of Computational Science and Simulation Technology, Institute of Photonics, Department of Mathematics and Physics, Leibniz University Hannover, 30167 Hannover, Germany}
\affil[2]{Institute of Structural Mechanics, Bauhaus-Universität Weimar, Germany}
\affil[3]{Department of Engineering Mechanics, Tsinghua University, Beijing, China}

\begin{abstract}
Partial differential equations (PDEs) underpin quantitative descriptions across the physical sciences and engineering, yet high-fidelity simulation remains a major computational bottleneck for many-query, real-time, and design tasks. Data-driven surrogates can be strikingly fast but are often unreliable when applied outside their training distribution. Here we introduce Neural Operator Warm Starts (NOWS), a hybrid strategy that harnesses learned solution operators to accelerate classical iterative solvers by producing high-quality initial guesses for Krylov methods such as conjugate gradient and GMRES. NOWS leaves existing discretizations and solver infrastructures intact, integrating seamlessly with finite-difference, finite-element, isogeometric analysis, finite volume method, etc. Across our benchmarks, the learned initialization consistently reduces iteration counts and end-to-end runtime, resulting in a reduction of the computational time of up to 90\%, while preserving the stability and convergence guarantees of the underlying numerical algorithms. By combining the rapid inference of neural operators with the rigor of traditional solvers, NOWS provides a practical and trustworthy approach to accelerate high-fidelity PDE simulations. 
\end{abstract}

\begin{document}

\flushbottom
\maketitle 

\section{Introduction}

From predicting the evolution of weather systems and modeling blood flow in the human body to designing aircraft and simulating the behavior of materials under stress, PDEs form the mathematical foundation of much of modern science and engineering. Advances in computational methods over the past half-century have made it possible to simulate increasingly complex systems with remarkable accuracy, transforming both fundamental research and industrial practice \cite{quarteroni1994numerical}. Yet, the growing demand for high-fidelity models, driven by real-time and design tasks, and involving large numbers of degrees of freedom and intricate geometries, has pushed traditional numerical solvers to their computational limits \cite{menghal2012real}. This challenge is particularly acute in applications such as optimization \cite{biegler2010nonlinear}, uncertainty quantification \cite{smith2024uncertainty}, and digital twins \cite{fuller2020digital, es2024methods}, where even state-of-the-art algorithms can be prohibitively slow. Classical numerical methods, such as Finite Element Method (FEM), Finite Difference Method (FDM), and Finite Volume Method (FVM), combined with iterative solvers \cite{hageman2012applied, greenbaum1997iterative, van2003iterative, Xu1034116Iterative} such as conjugate gradient \cite{4307698, KERSHAW197843, stranden1999solving} or GMRES \cite{10363502, 9462418, ThomasIterated, AmestoyFive}, remain the gold standard for robustness and accuracy. However, as the scale of problems grows, these solvers often require large numbers of iterations to converge, particularly for ill-conditioned systems \cite{qiu2015efficient}, or complex geometries \cite{mahesh2004numerical}. High-resolution meshes further amplify the computational burden, making acceleration strategies increasingly essential \cite{kronbichler2012generic}.

In recent years, neural operators have emerged as a promising alternative by learning mappings between function spaces and approximating entire families of PDE solution operators. Early work includes the Graph Kernel Network \cite{li2020neural}, which uses integral kernels implemented via graph message-passing. DeepONet \cite{lu2021learning}, inspired by the universal approximation theorem for operators, approaches the problem through a branch–trunk network decomposition, enabling flexible representation of nonlinear operators across diverse domains. The Fourier Neural Operator (FNO) \cite{li2020fourier} introduced spectral convolution layers to capture global dependencies, demonstrating strong performance on a range of parametric PDEs. Notably, the authors report FNO as the first ML-based method to successfully model turbulent flows with zero-shot super-resolution. The Variational Physics-informed Neural Operator (VINO) \cite{eshaghi2025variational} further extends this framework by incorporating variational inference and physics-informed constraints, enabling robust learning in settings without any data. Variants based on transformers \cite{pmlr-v202-hao23c, SHIH2025117560, guibas2021adaptive} and graph neural networks \cite{xu2024equivariant} have extended these capabilities to irregular meshes \cite{pmlr-v202-hao23c}, complex geometries \cite{FU2025109754}, and coupled multi-physics systems \cite{zheng2025muti, li2025multi}. While these models often achieve inference speeds far exceeding classical solvers once trained, their training cost and limited robustness to distribution shifts remain significant challenges.

Parallel to these developments, there is a growing body of work exploring hybrid methods that integrate learned surrogates with traditional numerical solvers. Neural networks have been employed to correct coarse-grid results in multigrid and solver-in-the-loop schemes \cite{um2020solver}, to design improved iterative strategies with convergence guarantees \cite{hsieh2019learning, he2019mgnet, chen2022meta}, and to generate better initial guesses for nonlinear problems \cite{huang2020int}. They have also been used to construct problem-dependent prolongation and restriction operators \cite{pmlr-v119-luz20a, greenfeld2019learning}, and to learn multigrid-augmented or data-driven preconditioners \cite{azulay2022multigrid, tan2025scalable, lee2025fast, herb2025accelerating, giraud2025neural, zhang2024blending, han2024ugrid, huang2023reducing, kopanivcakova2025leveraging, rubio2025preconditioning, song2025matrix}. While these approaches can significantly accelerate convergence by addressing low-frequency error components, they often require architectural modifications to the solver, may introduce stability concerns, and sometimes lack theoretical convergence guarantees. In addition, there is a growing body of work exploring learned warm starts for iterative solvers. For example, neural networks have been used to provide initial guesses for specialized steady-state simulations \cite{zhou2025neural} and for AI-enhanced iterative solvers in large-scale parametric systems \cite{nikolopoulos2024ai}. While these approaches demonstrate the potential of learning-based initialization, they are typically tailored to particular problem classes or solver setups. 

Here we introduce Neural Operator Warm Starts (NOWS), a hybrid paradigm that unites the speed of learned operators with the reliability of classical iterative solvers. Rather than replacing numerical methods, NOWS employs a neural operator to generate high-quality initial guesses that sharply reduce the initial residual, thereby lowering the iteration count required for full convergence. The neural operator rapidly eliminates the bulk of the error, leaving the classical solver to perform only fine-scale corrections. Crucially, this strategy preserves the stability, interpretability, and rigorous convergence guarantees of the underlying numerical method while leveraging the expressive power and speed of modern neural architectures. Because it requires no changes to the iterative scheme or preconditioning, NOWS is solver-agnostic and compatible with a wide range of existing solvers, including FEM, FDM, IGA, FVM, etc.

We demonstrate NOWS across diverse static and dynamic PDEs, showing iteration count reductions of two- to ten-fold and total runtime savings of 25–90\%, without sacrificing accuracy or robustness. Comparative results against purely classical and purely neural approaches reveal that NOWS offers a distinctive balance of efficiency and reliability, performing well even under mesh variation and changes in domain geometry. This inversion of perspective, from substituting numerical methods to complementing them, bridges the gap between data-driven learning and physics-based computation. The resulting framework is broadly applicable in domains where PDEs are the computational bottleneck, including climate and geophysical modeling, biomedical simulation, structural engineering, and reactive flow design. More generally, NOWS illustrates a principled pathway for integrating machine learning into scientific computing, enhancing rather than supplanting the foundational strengths of traditional numerical methods.

While classical deflation and Krylov recycling methods are effective in repeated-solve regimes, NOWS offers a complementary acceleration mechanism. By generating high-quality, learned initial guesses via neural operators, NOWS reduces the initial residual and total iterations across a wide range of PDEs, solver types, and discretizations. Importantly, it preserves convergence guarantees and can be combined with recycling strategies for further speed-up. This hybrid approach broadens the scope of accelerated PDE solvers, particularly for parametric and multi-query problems where subspace recycling alone may not suffice.

\section{Method} \label{sec:Method}

\subsection{Problem formulation}
We consider the generic family of parametric partial differential equations of the form
\begin{equation}
(L_a u)(x) = f(x), \quad x \in D, 
\qquad u(x) = 0, \quad x \in \partial D,
\label{eq:pde}
\end{equation}
for parameters $a \in \mathcal{A}$, right-hand side $f \in \mathcal{U}^\ast$, and a bounded domain $D \subset \mathbb{R}^d$. 
The solution $u : D \to \mathbb{R}$ is assumed to live in a Banach space $\mathcal{U}$, 
and the operator $L_a : \mathcal{A} \to \mathcal{L}(\mathcal{U}; \mathcal{U}^\ast)$ maps the parameter space into a space of linear operators from $\mathcal{U}$ to its dual, $\mathcal{U}^\ast$. 
The solution operator naturally associated with this PDE is
\[
\mathcal{G}^\dagger := L_a^{-1} f : \mathcal{A} \to U,
\]
which maps the parameter $a$ to the solution $u$. 
When needed, we assume that the domain $D$ is discretized into $K \in \mathbb{N}$ points and that we observe 
$N \in \mathbb{N}$ pairs of coefficients and approximate solutions 
$\{a^{(i)}, u^{(i)}\}_{i=1}^N$, which are used for model training. 
Here, the coefficients $a^{(i)}$ are independent and identically distributed samples 
from a probability measure $\mu$ supported on $\mathcal{A}$, 
and the corresponding solutions $u^{(i)}$ are pushforward under $\mathcal{G}^\dagger$.

\subsection{Neural Operator Approximation}
At the first stage, we train a neural operator offline. The training task is to approximate the map between two infinite-dimensional spaces using only a finite collection of input–output pairs, and possibly physical laws when the network is physics-informed. More concretely, let $\mathcal{A}$ and $\mathcal{U}$ denote Banach spaces of functions defined on 
bounded domains $D \subset \mathbb{R}^d$ and $D' \subset \mathbb{R}^{d'}$, respectively, and let $\mathcal{G}^\dagger : \mathcal{A} \to \mathcal{U}$ be the (possibly nonlinear) solution operator. Given observations $\{a^{(i)}, u^{(i)}\}_{i=1}^N$, where $a^{(i)} \sim \mu$ are drawn from a probability distribution on $\mathcal{A}$ 
and $u^{(i)} = \mathcal{G}^\dagger(a^{(i)})$ are possibly corrupted by noise, the goal is to construct a parametric surrogate
\begin{equation}
\mathcal{G}_\theta : \mathcal{A} \to \mathcal{U}, \quad \theta \in \mathbb{R}^p,
\label{eq:neuralop}
\end{equation}
and determine parameters $\theta^\dagger \in \mathbb{R}^p$ such that $\mathcal{G}_{\theta^\dagger} \approx \mathcal{G}^\dagger$. 
Neural operators such as FNO, DeepONet, VINO, and their extensions provide flexible frameworks for this approximation. 

In physics-informed configurations, the energy-based loss function provides physics-consistent training without requiring labeled data:
\begin{equation}
    \mathcal{L}(\phi) = 
    \frac{1}{|\mathcal{B}|} 
    \sum_{(P, F, b) \in \mathcal{B}} 
    \big| \Psi(\mathcal{G}_\phi(P, F, b); P, b) - \Psi_{\mathrm{ref}} \big|,
\end{equation}
where $\mathcal{B}$ is a batch of training examples, $\Psi_{\mathrm{ref}}$ is a reference energy (often zero for equilibrium problems).  

\subsection{Neural Operator Warm Starts (NOWS)}
\begin{figure}[ht!]
    \centering
    \includegraphics[width=0.8\textwidth]{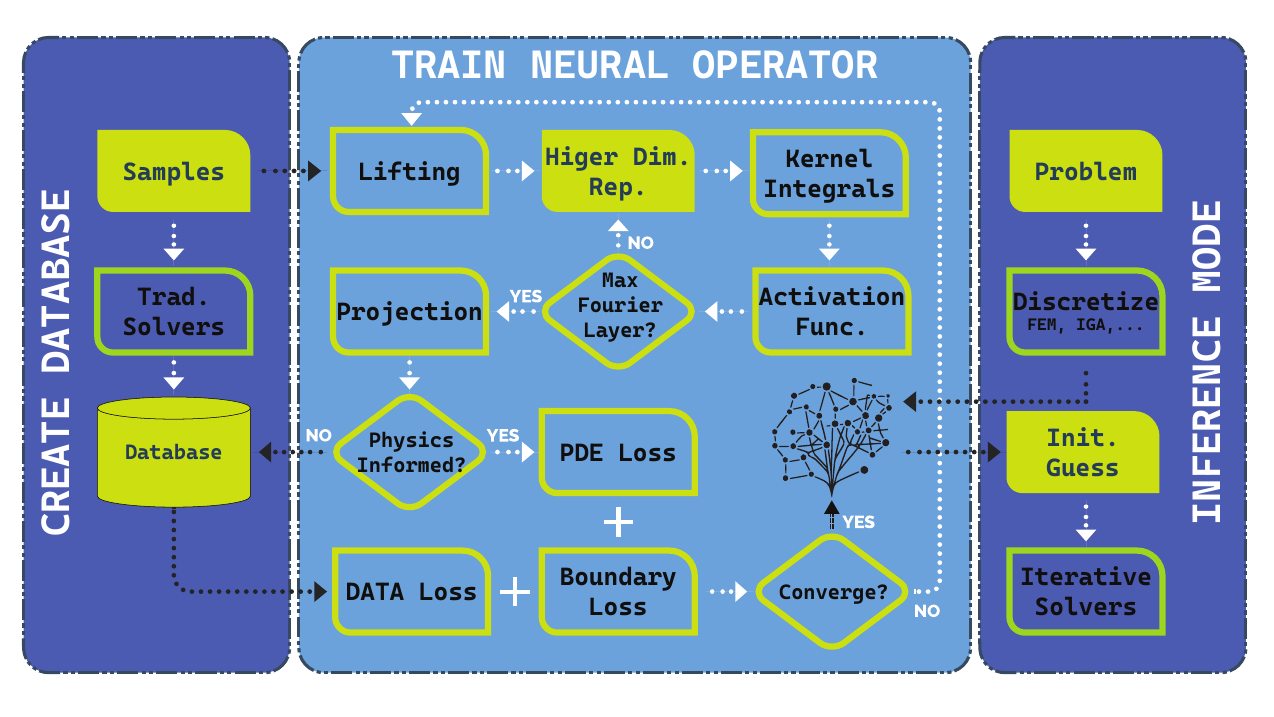}
    \caption{\textbf{Workflow of the Neural Operator Warm Start (NOWS) framework.}
    Schematic overview of the NOWS methodology integrating neural operators and numerical solvers. The left branch illustrates the \textit{data creation phase}, where data are generated, if needed (not for PI versions), using traditional solvers. The middle part describes \textit{training phase}, where a Neural Operator is trained through a combination of data, physics, and boundary losses. The architecture cycles through lifting, integral kernels (Fourier layers), activation, and projection steps until convergence. The right branch shows the \textit{inference phase}, in which parametric PDE problems are discretized 
    (e.g., using FEM, IGA, FVM). The trained Neural Operator provides an initial guess for a conventional iterative solver (e.g., CG, GMRES), which then refines the solution to full numerical accuracy. 
    This hybrid workflow leverages the efficiency of learned models and the robustness of classical solvers 
    to accelerate PDE solvers.}
    \label{fig:nows_workflow}
\end{figure}

Once the neural operator is trained, it is integrated into the NOWS workflow (Fig.~\ref{fig:nows_workflow}). The essential idea is simple: instead of relying on the neural operator for the entire solution process, we use it to provide an accurate initial guess for a conventional iterative solver. This initialization reduces the magnitude of the initial residual, allowing the solver to converge in significantly fewer iterations while maintaining its native guarantees of stability and accuracy. In other words, the NOWS framework uses a trained neural operator to provide an initial guess for a conventional iterative solver rather than replacing the solver itself.

Given a discretized system
\begin{equation}
A u = b,
\end{equation}
the neural operator produces an initial approximation
\begin{equation}
u_0 = \mathcal{G}_\theta(a),
\end{equation}
which is supplied to an iterative method such as Conjugate Gradient or GMRES. The solver then proceeds according to its standard update rule, possibly with preconditioning,
\begin{equation}
u_{k+1} = u_k + \alpha_k P^{-1}(b - A u_k), \quad 
u_0 = \mathcal{G}_\theta(a),
\end{equation}
where $P$ denotes a preconditioner and $\alpha_k$ is the step size.

By reducing the norm of the initial residual $r_0 = b - A u_0$, NOWS significantly lowers the number of iterations required for convergence. Importantly, the classical solver retains its original convergence guarantees, and no modifications to the numerical method, preconditioner, or stopping criteria are required. The approach is therefore solver-agnostic and compatible with a wide range of discretizations and iterative schemes.

The neural operator component of NOWS can be selected according to the problem's requirements. Fourier Neural Operators are particularly well-suited for problems with periodic domains or smooth spectral structures, DeepONets offer flexibility across diverse geometries, and VINO integrates physics into loss functions, thereby enhancing generalization capability.
Importantly, neural operators enjoy resolution invariance; models trained on coarse discretizations 
can be applied to finer meshes without retraining. This property allows us to train efficiently at modest resolution and later deploy the operator to accelerate solvers at scales where traditional training would be infeasible. We demonstrate this capability in the rising hot-smoke plume problem, where a neural operator trained on a coarse grid was successfully used to initialize solvers on high-resolution meshes, yielding consistent acceleration.

NOWS thus acts as a bridge between the speed of learned models and the reliability of numerical solvers. The neural operator eliminates the expensive early phase of iterative refinement, while the classical solver ensures full accuracy even under conditions not represented in the training set. The combination results in substantial runtime savings, robustness across problem classes, and seamless compatibility with established computational infrastructures. 

\subsection{Numerical discretization and solvers}

The PDEs considered in this work are discretized using standard numerical techniques appropriate to each application, including Finite Difference, Finite Element, Isogeometric Analysis, and Finite Volume methods. The resulting systems are solved using iterative solvers, primarily CG. 

Preconditioners such as Jacobi, SSOR, Incomplete Cholesky (ICC), and Incomplete LU (ILU) are employed where appropriate. For all comparisons, discretizations, boundary conditions, solver tolerances, and stopping criteria are kept identical between baseline solvers and their NOWS-accelerated counterparts.

Crucially, the approach is solver-agnostic: any iterative scheme with any preconditioner, including conjugate gradient, GMRES,  and multigrid methods, can be combined with NOWS. Similarly, the framework is independent of the underlying numerical discretization, and can be applied with finite element, finite difference, isogeometric, or finite volume methods.

It is important to distinguish between the roles of numerical discretization and linear solvers in PDE simulations. Methods such as the FEM, FDM, and FVM are used to discretize the governing PDE, resulting in a system of algebraic equations of the form (Au=b). The solution of this system is then performed using either direct solvers (e.g., sparse LU factorization) or iterative methods such as conjugate gradient (CG) or GMRES. The proposed NOWS framework does not replace the discretization step; instead, it augments the solver stage by providing a high-quality initial guess for the iterative method. Consequently, the relevant comparison is between a standard discretization–solver pipeline (e.g., FEM + CG) and the same pipeline with a neural warm start (FEM + CG + NOWS). In this sense, NOWS is complementary to classical numerical methods and can be integrated with a wide range of discretizations without modifying the underlying solver algorithms.

\section{Results} \label{sec:Results}

This section presents numerical results demonstrating the effectiveness of Neural Operator Warm Starts across a range of PDEs, discretizations, and solver configurations. The focus is on iteration count reduction, robustness with respect to mesh resolution and geometry, and overall computational efficiency. In the subsequent subsections, we will evaluate NOWS's performance through a range of numerical examples.

The numerical experiments cover elliptic and parabolic PDEs, static and dynamic problems, structured and unstructured discretizations, varying mesh resolutions, and domain geometries. The following subsections report detailed results for each scenario, beginning with a general demonstration of how NOWS bridges neural operator predictions and classical iterative solvers, followed by analyses of resolution robustness, physics-informed training, interaction with Krylov preconditioners, and applicability to complex geometries and time-dependent problems.

The runtime reported for NOWS includes both the neural network inference and the subsequent CG solver execution. The reported speedups therefore reflect the total online runtime, ensuring a fair comparison with the baseline solver. The database generation and training phases constitute offline costs, which are incurred once when constructing the neural operator. After training, the model can be reused for many solves of the same PDE class and can also be applied across different grid resolutions. Consequently, the computational benefit of NOWS is most pronounced in many-query scenarios, such as parameter studies, optimization, or uncertainty quantification, where the one-time offline cost can be amortized over a large number of simulations.

An additional trend visible in the violin plots is that the relative time savings provided by NOWS may decrease slightly as the convergence tolerance becomes more stringent. This behavior is expected because the neural operator primarily reduces the magnitude of the initial residual and removes large-scale error components, thereby accelerating the early phase of the iterative process. When very small tolerances are required, the solver must still perform fine-scale corrections to reach the prescribed accuracy. In this regime, the later stages of the iteration—dominated by the classical solver dynamics—constitute a larger fraction of the total runtime, which reduces the relative contribution of the neural initialization. Nevertheless, substantial acceleration is still observed across all tolerance levels.

\subsection{Robustness of NOWS across resolutions}

We first consider the canonical Poisson equation, a second-order elliptic PDE with a source term $p(x)$, given by
\begin{equation}
\begin{aligned}
    \nabla^2 s(x) + p(x) &= 0, && x \in \Omega = [0,1]^2, \\
    s(x) &= 0, && x \in \partial \Omega,
\end{aligned}
\label{eq:poisson-2d}
\end{equation}
where the objective is to learn the solution operator $\mathcal{G}$ that maps the source function $p(x)$ to the corresponding solution $s(x)$. To construct the neural operator, we trained a VINO using pairs $\{p(x), s(x)\}$ and PDE itself, where the forcing terms $p(x)$ are generated from Gaussian random fields (GRFs). The trained operator serves as the initialization mechanism in the NOWS framework. Once training is complete, we evaluate the method on 2000 previously unseen realizations of $p(x)$. For the iterative solver, we employ the conjugate gradient (CG) method, chosen for its efficiency in handling symmetric positive-definite systems arising from discretizations of the Poisson problem. 

\begin{figure}[ht!]
    \centering
    \includegraphics[width=1.0\textwidth]{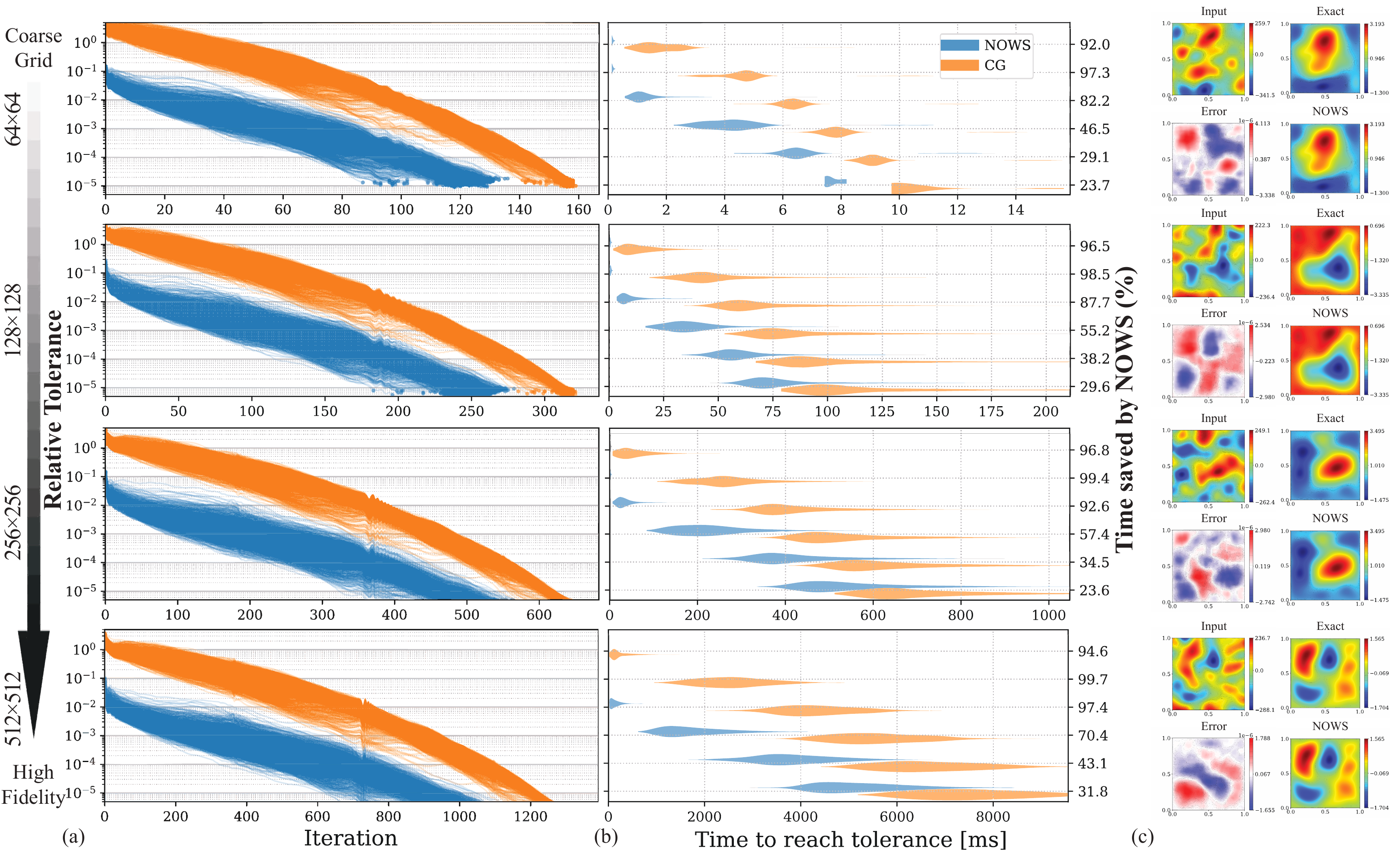}
    \caption{\textbf{NOWS accelerates iterative solvers in various resolutions.}
 (a) Overlaid residual-vs-iteration trajectories for the two methods on the 2000-sample test set, showing that NOWS reduces both the initial residual and the number of iterations to convergence. (b) Violin plots of wall-clock time distributions (per test instance) required to reach several relative residual tolerances, comparing the baseline and NOWS. (c) An example test case comparing the NOWS solution, the reference obtained with the direct solver. The NOWS removes the large-scale error, while the classical solver enforces exactness. Results are aggregated over 2000 unseen GRF realizations.}

    \label{fig:poisson_results}
\end{figure}

The experiments are conducted independently at multiple mesh resolutions, ranging from coarse to high-fidelity discretizations. This design allows us to assess the performance of NOWS across discretization scales, demonstrating that the method consistently delivers effective initializations and accelerates convergence regardless of mesh density. Figure~\ref{fig:poisson_results} summarizes the performance of NOWS compared with the standalone iterative solver over the 2000 test cases. Across all resolutions, NOWS achieves a consistent reduction in iteration counts and total runtime while retaining the final accuracy of the classical solver. The results confirm that using a neural operator solely for initialization provides a reliable and scalable acceleration strategy for elliptic PDEs.

The results in Fig.~\ref{fig:poisson_results} illustrate the impact of NOWS across four discretization levels, ranging from coarse meshes ($64 \times 64$) to high-fidelity grids ($512 \times 512$). 
Each row corresponds to a specific resolution, with the left column reporting the relative tolerance versus iteration count, 
and the right column summarizing the runtime to convergence across the 2000 test cases. 
The gradient arrow on the left highlights the transition from coarse discretizations at the top to high-fidelity meshes at the bottom, providing a visual link between mesh density and solver behavior.  

The violin plots summarize the distribution of wall-clock times required to reach the prescribed relative residual tolerance across the test cases. To quantify the benefit of neural initialization, we report the percentage of time saved by NOWS relative to the baseline iterative solver. This quantity is defined as the relative reduction in runtime,

\begin{equation}
S = \frac{T_{\mathrm{CG}} - T_{\mathrm{NOWS}}}{T_{\mathrm{CG}}} \times 100\% ,
\end{equation}

where $T_{\mathrm{CG}}$ denotes the wall-clock time required for the baseline solver to reach the target tolerance, and $T_{\mathrm{NOWS}}$ is the corresponding runtime when the solver is initialized with the neural operator warm start. Positive values therefore indicate a reduction in computational time achieved by NOWS. The percentages reported on the right axis of the plots correspond to this relative time saving.

In all cases, the conjugate gradient (CG) method initialized with NOWS (blue) consistently outperforms the standalone CG solver (orange). The left panels clearly show that NOWS reduces the initial residual magnitude, shifting the convergence curve downward and resulting in faster attainment of target tolerances. This improvement translates directly into reduced iteration counts: at coarse resolution, savings of roughly $30\%$--$40\%$ are observed, while for finer meshes the savings become even more pronounced, exceeding $50\%$ in many cases.  

The right panels quantify these improvements in terms of runtime. Across all resolutions, the violin plots indicate a clear reduction in the mean solution time when NOWS is used to initialize the solver. While the distributions generally shift toward shorter runtimes due to the improved initial guess, the variance of the runtime distribution is not consistently lower for NOWS across all resolutions and tolerance levels. In particular, for the coarsest grid the CG baseline exhibits slightly narrower distributions. These results nonetheless confirm that NOWS reliably reduces the average computational cost and iteration count by lowering the initial residual and accelerating convergence toward the target tolerance. The annotated percentages on the right axis represent the fraction of computational time saved by NOWS, demonstrating robust acceleration benefits that persist even as the problem size grows. Importantly, the advantage of NOWS scales favorably with resolution: as mesh density increases, the gap between the two methods widens, underscoring the resolution-invariance property of neural operators.  

Together, these results confirm that NOWS provides reliable acceleration across discretization levels, delivering substantial iteration and runtime savings without compromising accuracy. 
This robustness with respect to mesh resolution is critical for practical applications, where simulations often demand high fidelity but cannot afford the prohibitive computational cost of classical iterative refinement alone.

\subsection{Physics-Informed NOWS}

We evaluate the role of physics-informed supervision in training neural operators for steady-state Darcy flow. The benchmark problem involves two-dimensional subsurface flow through heterogeneous porous media, where the conductivity field \(p(x)\) is modeled as a piecewise-constant random function and the hydraulic head \(s(x)\) satisfies
\begin{equation}
    -\nabla \cdot (p(x) \nabla s(x)) = f(x), \quad x \in [0,1]^2, \qquad
    s(x) = 0, \quad x \in \partial [0,1]^2,
\end{equation}
with \(f(x) = 1\). The objective is to learn the operator \(\mathcal{G}: p \mapsto s\), mapping the heterogeneous conductivity field to the solution of the PDE. Although the PDE is linear, the solution operator is inherently nonlinear due to the piecewise nature of \(p(x)\).

To enforce physics consistency, the neural operator is trained via the variational (energy) form of the Darcy equation, minimizing the corresponding functional derived from the PDE. Three training strategies are compared: purely data-driven supervision, physics-informed training based on the variational residual, and a hybrid approach combining data and physics losses. The operator is coupled with a classical finite element method (FEM) solver in this study, providing a reference solution for error evaluation; this contrasts with previous experiments using finite difference discretizations. All models share the same backbone structure and are trained on permeability fields drawn from Gaussian random processes with varying correlation lengths.

\begin{figure}[ht!]
    \centering
    \includegraphics[width=1.0\textwidth]{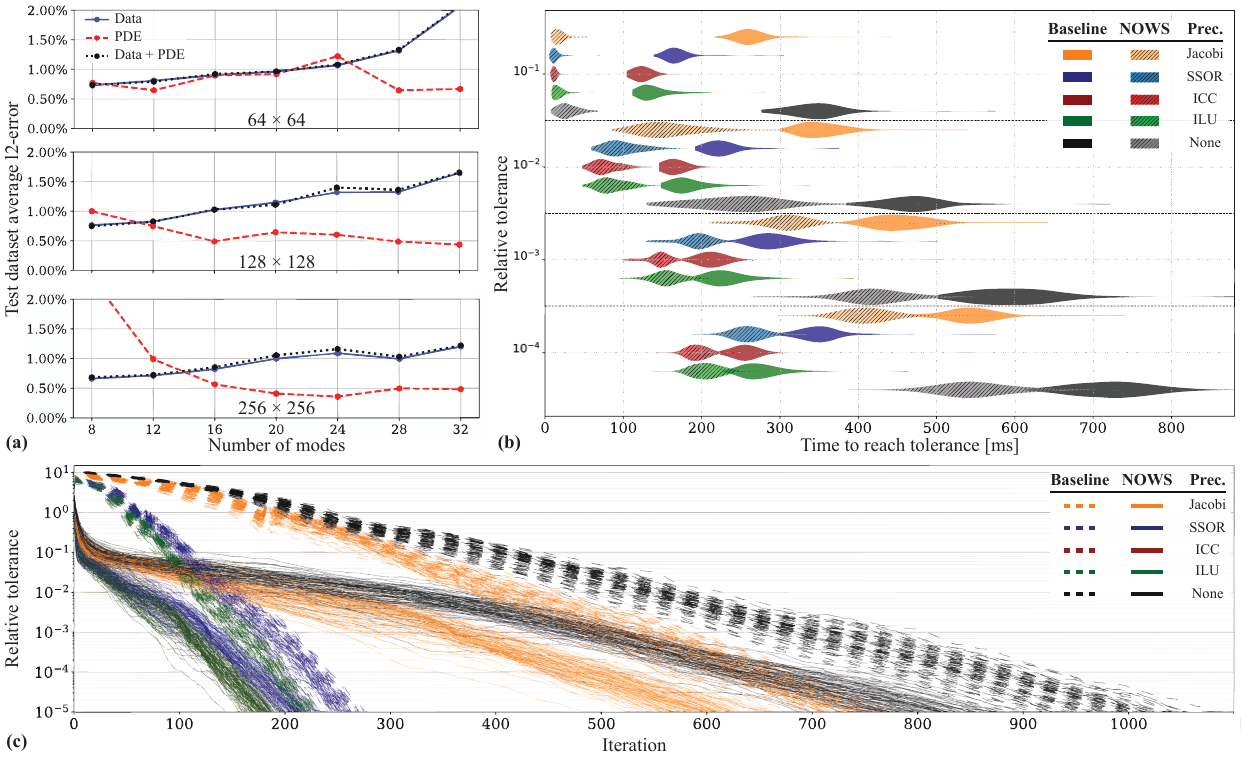}
    \caption{\textbf{Impact of physics-informed training and neural-operator warm starts (NOWS) on Darcy flow simulations.} 
\textbf{(a)} Test error as a function of network complexity, comparing data-only, physics-only, and hybrid training strategies. Physics-informed supervision yields the most robust generalization, with test error decreasing as the number of Fourier modes or network parameters increases, whereas data-only and hybrid approaches exhibit overfitting. Embedding the PDE directly in the loss produces smoother and more stable operator predictions that respect the underlying physics across resolutions and heterogeneous permeability contrasts. 
\textbf{(b)} Wall-clock runtime distributions for various preconditioners and relative tolerance levels, with and without NOWS initialization. Across all preconditioners, NOWS consistently reduces solver runtime, demonstrating measurable acceleration even for simple Jacobi preconditioning, with more pronounced improvements for ICC and ILU schemes. 
\textbf{(c)} Relative residuals versus iteration count for different preconditioners. NOWS systematically lowers the initial residual and reduces the number of iterations required to achieve the prescribed convergence tolerance, illustrating solver-agnostic acceleration and robustness across preconditioning strategies.}
    \label{fig:darcy_results}
\end{figure}

Figure~\ref{fig:darcy_results}a demonstrates that the physics-informed neural operator achieves the most robust generalization. As the number of Fourier modes or network parameters increases, the test error decreases under physics-only supervision, while it rises for data-only and mixed models, indicating susceptibility to overfitting in the absence of explicit PDE constraints. Embedding the PDE structure directly in the training loss produces smoother, more stable operator predictions that respect the underlying physics across resolution scales and heterogeneous permeability contrasts.

These results highlight that the incorporation of variational PDE constraints enables neural operators to capture essential physical structure, improving both predictive accuracy and robustness. The synergy between energy-based loss formulations and classical FEM solutions establishes a framework for physics-consistent operator learning that is both generalizable and compatible with established numerical methods. Embedding the underlying physics reduces the reliance on large labeled datasets and promotes smoother, more stable operator learning, which are key characteristics for PDE acceleration within the NOWS framework.

\subsection{Preconditioning of Krylov Methods in NOWS}

We next assess the performance of NOWS framework when integrated with various preconditioning techniques for Krylov subspace methods. The analysis is carried out on the steady-state Darcy flow problem, using VINO via the variational (energy-based) formulation of the governing PDE. In this configuration, the neural operator provides physics-consistent initial conditions for the iterative solver.

To investigate the interplay between data-driven initialization and algebraic preconditioning, we consider several standard preconditioners for the conjugate gradient (CG) method, including the unpreconditioned baseline, Jacobi, SSOR (Symmetric Successive Over-Relaxation), ICC (Incomplete Cholesky), and ILU (Incomplete LU) schemes. Each preconditioner alters the spectral characteristics of the discrete Darcy system and thereby affects convergence behavior.

For each configuration, we compare the baseline CG solver with its NOWS-augmented counterpart, in which the neural operator prediction serves as the initial iterate \(u_0\). Performance is evaluated over a range of relative residual tolerances, and both iteration count and total runtime are recorded. The experiments are conducted across 2000 unseen permeability realizations, ensuring statistical robustness.

Figure~\ref{fig:darcy_results}b summarizes the runtime distributions for different preconditioners and tolerance levels, with and without NOWS initialization. Across all settings, a consistent pattern emerges: NOWS reduces solver runtime irrespective of the preconditioner employed. Even for simple diagonal preconditioners such as Jacobi, a measurable reduction in computational cost is observed, while more advanced schemes like ICC and ILU exhibit pronounced acceleration when combined with neural initialization.

The complementary nature of preconditioning and NOWS is further evident in Figure~\ref{fig:darcy_results}c, which reports the relative residuals as a function of iteration count. The NOWS initialization systematically decreases both the initial residual magnitude and the number of iterations required to meet the prescribed convergence criterion. This effect remains consistent across all preconditioners, demonstrating the robustness and solver-agnostic character of NOWS.

The results establish that the benefits of neural initialization are invariant to preconditioner selection. Regardless of the underlying conditioning strategy, the integration of NOWS accelerates convergence and reduces total runtime without modifying the solver’s structure. The neural operator provides a physically consistent approximation that narrows the solution search space, while preconditioners refine the spectral properties of the linear system. The preconditioning analysis thus reinforces the generality of NOWS as a physics-grounded, solver-independent enhancement for PDE computations.

While the present study focuses on classical algebraic preconditioners, large-scale production codes often rely on multilevel strategies such as algebraic multigrid or domain decomposition methods. These approaches primarily improve the spectral conditioning of the linear system, whereas NOWS reduces the magnitude of the initial residual. Because these mechanisms act on different aspects of the iterative process, they are expected to remain complementary. A systematic evaluation of NOWS in combination with multilevel preconditioners is an interesting direction for future work.

\subsection{Applicability to irregular geometries and Isogeometric Analysis}

To demonstrate the flexibility of NOWS in complex domains, we consider a two-dimensional plate with arbitrarily shaped internal voids. The plate is modeled using a parametric linear elasticity problem, discretized with an isogeometric analysis (IGA) solver, where the displacement field \(\mathbf{u}(x,y)\) satisfies
\begin{equation}
\begin{split}
    & \frac{\partial \sigma_{xx}}{\partial x} + \frac{\partial \sigma_{xy}}{\partial y} + f_x = 0, \\
    & \frac{\partial \sigma_{yx}}{\partial x} + \frac{\partial \sigma_{yy}}{\partial y} + f_y = 0, \\
    & \mathbf{u}(x,y) = 0 \quad \text{for } x=0, \, L, \\
    & \boldsymbol{\sigma} \cdot \mathbf{n} = \hat{\mathbf{t}}(y) \quad \text{for } y=D,
\end{split}
\end{equation}
with stress defined via the constitutive relation \(\boldsymbol{\sigma} = E \mathbf{D} \boldsymbol{\varepsilon}\). Here, \(E\) denotes the spatially varying elasticity modulus, \(\mathbf{D}\) is the plane-stress constitutive matrix, and \(\boldsymbol{\varepsilon}\) is the strain vector. The matrix \(\mathbf{D}\) and vector definitions are given by
\begin{align}
\mathbf{D} &= \frac{1}{1-\nu^2}
\begin{bmatrix}
1 & \nu & 0 \\
\nu & 1 & 0 \\
0 & 0 & \frac{1-\nu}{2}
\end{bmatrix}, \\
\boldsymbol{\sigma} &= [\sigma_{xx}, \sigma_{yy}, \sigma_{xy}]^T, \quad 
\boldsymbol{\varepsilon} = [\varepsilon_{xx}, \varepsilon_{yy}, \varepsilon_{xy}]^T.
\end{align}

\begin{figure}[ht!]
    \centering
    \includegraphics[width=1.0\textwidth]{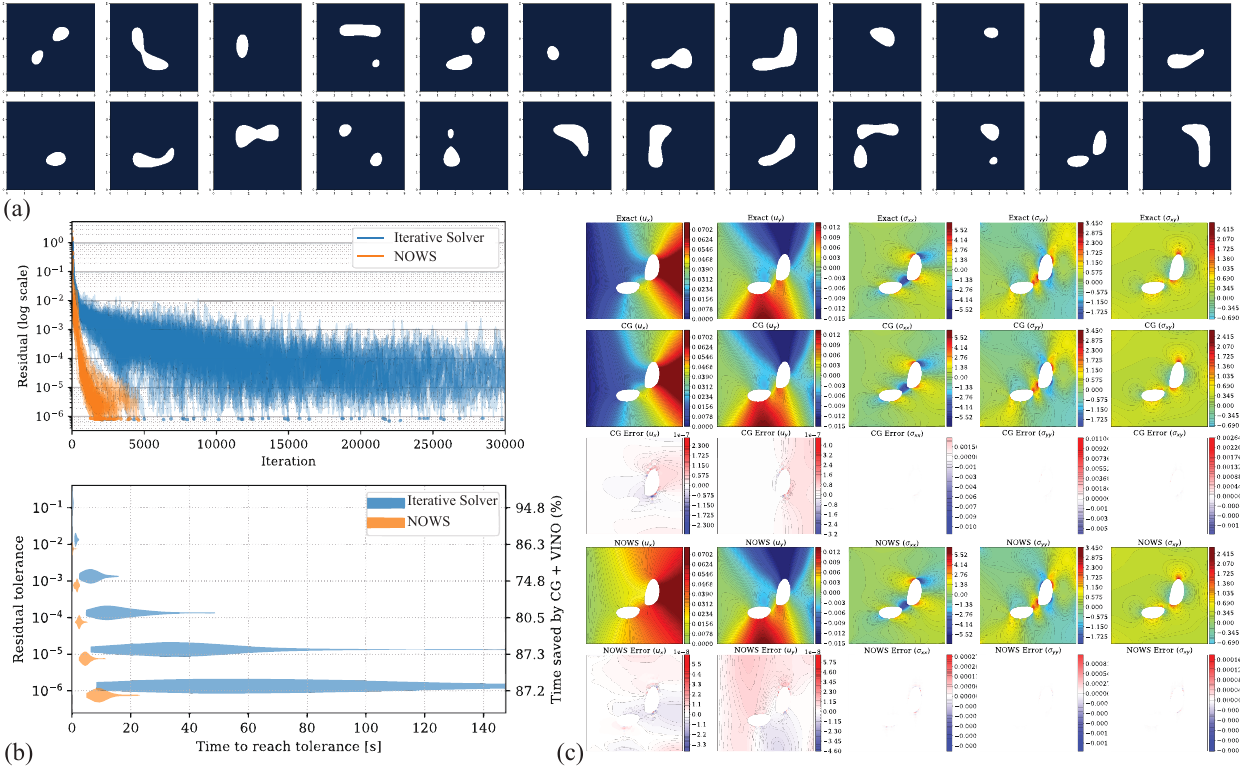}
\caption{\textbf{NOWS accelerates the iterative solution of PDEs on irregular domains.} 
\textbf{(a)} Random samples of allowable domain configurations with arbitrary voids. The plate is clamped on the left edge and subjected to constant traction on the right edge. 
\textbf{(b)} Relative residuals versus iteration count for the conjugate gradient (CG) solver initialized from zero or using the neural operator warm start (NOWS). NOWS consistently reduces the initial residual magnitude, lowering the convergence curve and decreasing the number of iterations required to reach the target tolerance. 
\textbf{(c)} Wall-clock runtime to convergence for 50 test instances. Across all geometries, NOWS yields ~10-fold speed-up and reduces iteration counts by roughly 90\%. Annotated percentages indicate the fraction of computational time saved, demonstrating robust acceleration benefits even for complex domain configurations.}
    \label{fig:plate_results}
\end{figure}

The domain \(\Omega_p = [0,5]^2 \setminus \mathcal{V}\) is parameterized by a set of design parameters \(p \in \mathcal{P}\), where \(\mathcal{V}\) represents the collection of arbitrary holes within the central region \([1,4]^2\). Dirichlet boundary conditions are imposed on the left edge, and a constant traction is applied on the right edge. The neural operator is trained to approximate the mapping \(\mathcal{G}: \Omega_p \mapsto \mathbf{u}\), capturing the solution field across variable geometries.

Random samples of allowable domain configurations are illustrated in Figure~\ref{fig:plate_results}a. The CG solver is initialized either from zero or using the neural operator prediction, NOWS. The results in Figure~\ref{fig:plate_results} summarize the performance of NOWS for 50 random test instances. Relative residuals versus iteration count (Fig.~\ref{fig:plate_results}b) and wall-clock runtime to convergence (Fig.~\ref{fig:plate_results}c) clearly demonstrate the acceleration benefits.

Across all geometries, NOWS consistently reduces the initial residual magnitude, lowering the convergence curve and enabling faster attainment of target tolerances. This improvement corresponds to a reduction of approximately 90\% in iteration counts and yields a speed-up of roughly tenfold in runtime. The annotated percentages in Fig.~\ref{fig:plate_results}c indicate the fraction of computational time saved, highlighting the robustness of NOWS even as the domain complexity increases. 

These results confirm that NOWS effectively accelerates iterative solvers for IGA-based simulations on irregular domains, providing substantial savings in both iterations and runtime without compromising solution accuracy.

\subsection{Applicability to dynamic problems}

To evaluate the generalization and robustness of NOWS in time-dependent settings, we apply the framework to the one-dimensional viscous Burgers’ equation, a canonical nonlinear PDE that models one-dimensional fluid motion under diffusion and advection. The governing equation is  
\begin{equation}
\partial_t u(x, t) + \partial_x \left(\frac{u^2(x, t)}{2}\right) = \nu \, \partial_{xx} u(x, t), 
\quad x \in (0,1), \, t \in (0,1],
\end{equation}
subject to periodic boundary conditions and an initial condition \(u(x,0) = u_0(x)\), where 
\(u_0 \in L^2_{\mathrm{per}}((0,1);\mathbb{R})\) and \(\nu > 0\) is the viscosity. 
The objective is to learn the nonlinear operator
\begin{equation}
\mathcal{G}^\dagger : L^2_{\mathrm{per}}((0,1);\mathbb{R}) \rightarrow 
H^r_{\mathrm{per}}((0,1);\mathbb{R}), 
\quad u_0 \mapsto u(\cdot, t),
\end{equation}
which maps the initial field to the solution at later times within the time interval [0,1].

\begin{figure}[ht!]
    \centering
    \includegraphics[width=1.0\textwidth]{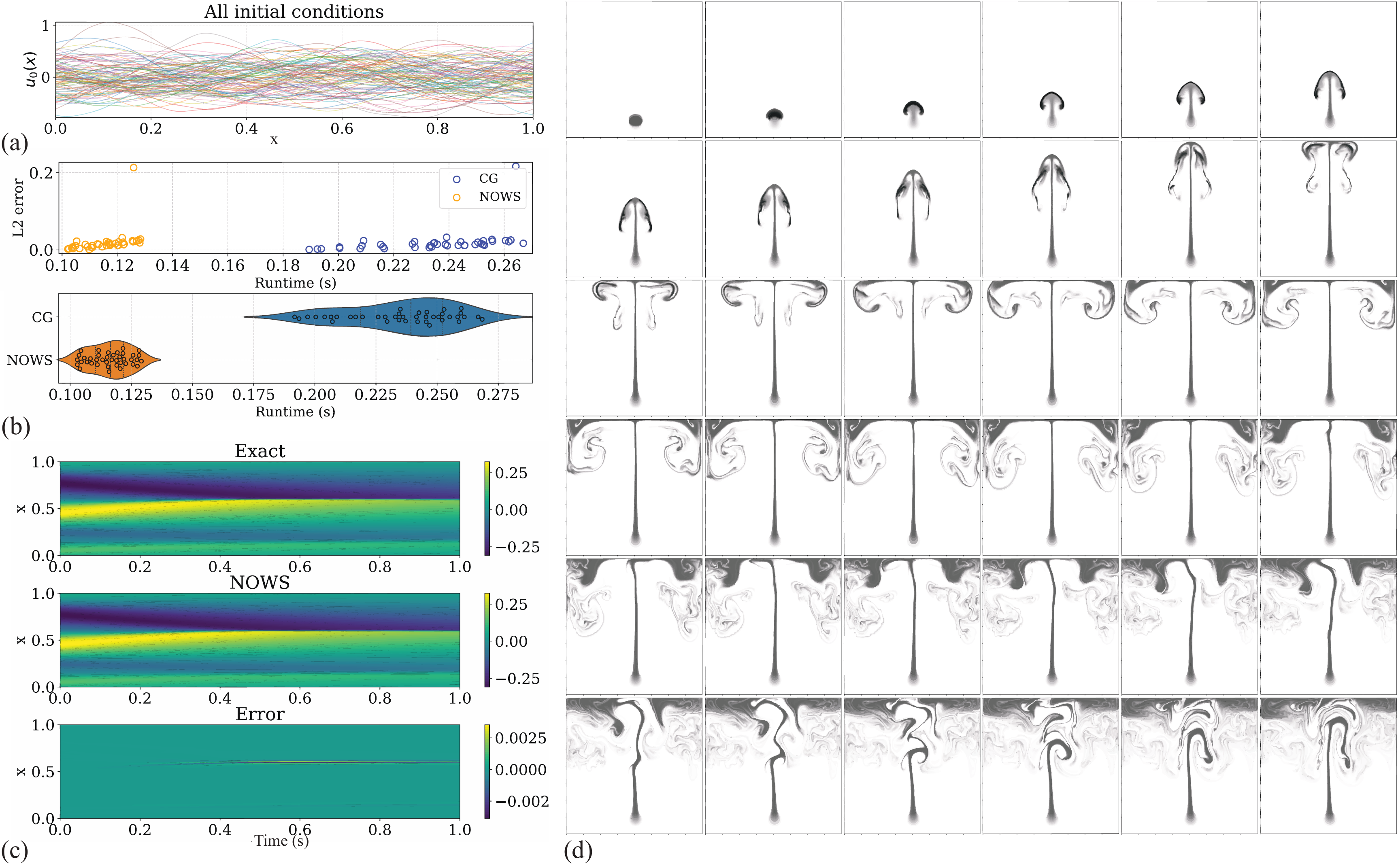}
    \caption{\textbf{NOWS for dynamic problems} 
    (a) Ensemble of initial conditions in the test dataset used for the Burgers’ equation. 
    (b) Comparing runtime distributions of CG and NOWS: scatter plot of runtime versus $L_2$ error for each sample (top),
    violin plots comparing the runtime distributions of the CG and NOWS solvers (bottom).  
    (c) Spatiotemporal evolution of the solution filed for a representative test sample in the Burgers' equation: exact reference (top), NOWS result (middle), and the corresponding error field (bottom).
    (d) \textbf{Resolution-invariant acceleration of coupled fluid–smoke simulations using NOWS.}
    A rising smoke plume governed by buoyancy-driven Navier–Stokes dynamics is simulated using NOWS. 
    The neural operator, trained on coarse 64×64 grids, is applied directly to 256×256 simulations without retraining. 
    Shown are representative smoke density snapshots at multiple time steps, illustrating stable and accurate evolution. 
    The NOWS initialization reduces total solver iterations by over 70\% and accelerates runtime by a factor of three, confirming robust resolution-invariant generalization across grid scales. }

    \label{fig:dynamic_results}
\end{figure}

In this experiment, NOWS is coupled with a classical finite-difference solver accelerated using the CG method, and we have used Multi-Head Neural Operator (MHNO) \cite{eshaghi2025multi} for operator learning. Figure~\ref{fig:dynamic_results}a illustrates the representative input profile, the corresponding exact and NOWS-predicted solutions, and their pointwise difference. The runtime statistics, summarized in Figure~\ref{fig:dynamic_results}b, confirm the efficiency gains provided by NOWS. The average wall-clock time per solve decreases from 9.62~s for the baseline CG method to 4.75~s with NOWS initialization—corresponding to approximately \textbf{50\% time savings}. Detailed statistics indicate consistent acceleration across all test cases, with mean, median, minimum, and maximum time reductions of 50.5\%, 50.9\%, 41.7\%, and 52.8\%, respectively. These results demonstrate that NOWS generalizes effectively to nonlinear, transient problems and retains its acceleration capability in dynamic PDE settings.

\subsection{Smoke Plume – Resolution Invariance}

To further assess the scalability and generalization of NOWS, we consider a coupled, nonlinear flow problem governed by the incompressible Navier–Stokes equations with buoyancy-driven smoke transport. The system is representative of large-scale multiphysics simulations typically solved using the FVM. The governing equations are
\begin{align}
\nabla \cdot \mathbf{u} &= 0, \\
\frac{\partial \mathbf{u}}{\partial t} + (\mathbf{u} \cdot \nabla)\mathbf{u} &= 
-\nabla p + \nu \nabla^2 \mathbf{u} + \mathbf{g}\,\beta (\rho - \rho_0), \\
\frac{\partial \rho}{\partial t} + (\mathbf{u} \cdot \nabla)\rho &= D \nabla^2 \rho,
\end{align}
where $\mathbf{u}$ denotes velocity, $p$ the pressure, $\rho$ the smoke density, $\nu$ the kinematic viscosity, $\beta$ the buoyancy coefficient, and $D$ the diffusion coefficient. The system couples the smoke advection–diffusion process with incompressible fluid motion, and the pressure projection step is solved iteratively using a CG solver.

In this setting, NOWS is employed to provide initial guesses for the CG-based pressure projection at each simulation step, thereby accelerating convergence while maintaining physical fidelity. The underlying numerical scheme follows a semi-Lagrangian advection and buoyancy update, followed by a pressure correction step enforcing incompressibility. The training process in NOWS ensures that the learned operator generalizes across flow configurations without requiring retraining for each grid.

To examine the mesh resolution invariance of NOWS, the neural operator was trained exclusively on coarse-grid simulations (64×64) and directly deployed on high-resolution grids (256×256) without further fine-tuning. The network has been trained to predict the pressure field at the current time step from the pressure field from the previous time step. Remarkably, the coarse-trained model provided physically consistent initializations for the fine-grid solver, retaining accuracy while significantly accelerating convergence. Quantitative performance comparisons highlight the impact of neural initialization. Without NOWS, the solver required 69,703 total iterations across 150 successful runs, yielding an average runtime of 294.74~s. When initialized with NOWS, total iterations reduced to 17,438, corresponding to a mean runtime of 93.85~s—an overall reduction of approximately \textbf{68\%} in computational time. Notably, convergence was achieved in all cases, indicating that the neural initialization preserves solver stability while enhancing efficiency.

Figure~\ref{fig:dynamic_results} illustrates representative smoke density fields at various time steps. The rising hot plume demonstrates that NOWS enables rapid convergence even for large-scale coupled PDE systems. The resolution-invariant behavior further confirms the ability of NOWS to generalize across spatial discretizations, a crucial property for scientific computing workflows where mesh adaptivity or refinement is required. The integration of NOWS with the finite volume solver thus represents a scalable strategy for accelerating multiphysics simulations without compromising physical accuracy.



\section{Conclusion}\label{sec:Conclusion}

The results presented across diverse PDE systems demonstrate that the NOWS framework provides a broadly applicable and robust acceleration mechanism for numerical solvers. By learning problem-specific operator mappings that approximate the inverse of the governing equations, NOWS delivers accurate initializations that significantly reduce the computational cost of iterative schemes. This hybridization between classical numerical methods and machine learning yields several key insights and implications.

First, the empirical evidence confirms that the physics-informed training of neural operators leads to superior generalization compared with purely data-driven approaches. In the Darcy flow experiments, enforcing the variational energy of the PDE during training resulted in models that remained stable and resistant to overfitting, even as network capacity or Fourier mode count increased. Using a physics-based loss function ensures that the network adheres to conservation laws and boundary constraints, maintaining fidelity across unseen material configurations and discretization scales.

Second, coupling NOWS with standard Krylov subspace preconditioner establishes a flexible and solver-agnostic framework. 
Across all examined preconditioners—from simple Jacobi to incomplete factorizations such as ICC and ILU—the neural warm start consistently lowered the initial residual and reduced iteration counts by up to an order of magnitude. 
These benefits are achieved without altering the solver’s mathematical formulation or convergence guarantees, distinguishing NOWS from traditional learned-solver replacements that may sacrifice interpretability or robustness.

Third, the integration of NOWS with Finite Difference, Finite Element, IGA, and Finite Volume frameworks highlights its modularity. The smoke plume experiment illustrates the resolution-invariance of the learned operator: a model trained on a coarse grid directly generalized to a four-times finer mesh without retraining, maintaining accuracy and achieving a threefold speed-up. This capability is particularly relevant for multiscale simulations and adaptive mesh refinement, where solver warm starts must remain consistent across varying resolutions.

While NOWS provides substantial acceleration across a wide range of PDE problems, it is important to acknowledge its limitations. First, the approach requires an offline investment in database generation and neural operator training; for scenarios involving only a few simulations, this cost may not be fully amortized. Second, although inference times are typically negligible compared to solver runtimes, in problems with extremely inexpensive solves or very complex neural architectures, inference could become a non-negligible contribution. Finally, cases where dataset generation itself is computationally expensive may reduce the relative benefit of NOWS. Despite these considerations, for most large-scale or many-query scenarios, NOWS delivers robust and significant runtime acceleration.

Beyond raw performance, NOWS changes how machine learning can interact with physics-based computation. Rather than substituting classical solvers, the framework augments them, retaining deterministic convergence while learning efficient priors from data or physical principles. This paradigm offers a pragmatic path toward physics-aware scientific computing, enabling acceleration and scalability without compromising trustworthiness. Future research directions include extending NOWS to graph-based operator architectures for unstructured meshes and integrating uncertainty quantification to assess confidence in neural initializations. Collectively, these results position NOWS as a unifying approach for embedding machine learning into numerical solvers, bridging the gap between data-driven models and first-principles computation.

\section*{Data availability}
The paper's datasets were created using the accompanying code. The dataset is available for free access at https://github.com/eshaghi-ms/NOWS

\section*{Code availability}
The code for all numerical experiments is publicly accessible on GitHub at https://github.com/eshaghi-ms/NOWS. 

\section*{Acknowledgement}
The authors would like to acknowledge the support provided by the German Academic Exchange Service (DAAD) through a scholarship awarded to Mohammad Sadegh Eshaghi during this research, as well as the Compute Servers of TU Ilmenau for providing computational resources.

\section*{Competing interests}
The authors declare that they have no known competing financial interests or personal relationships that could have appeared to influence the work reported in this paper.

\bibliography{mybibfile}


\makeatletter
\newcommand*{\addFileDependency}[1]{
  \typeout{(#1)}
  \@addtofilelist{#1}
  \IfFileExists{#1}{}{\typeout{No file #1.}}
}
\makeatother
\newcommand*{\myexternaldocument}[1]{
    \externaldocument{#1}
    \addFileDependency{#1.tex}
    \addFileDependency{#1.aux}}

\crefname{figure}{Fig.}{Figs.}
\crefformat{equation}{Eq.~#2(#1)#3}
\crefformat{section}{Section~#2#1#3}

\newpage
\vspace*{\baselineskip}
\centerline{\Large{\textbf{Supplementary Materials}}}
\vspace*{\baselineskip}
\centerline{\Large{\textbf{NOWS: Neural Operator Warm Starts for Accelerating Iterative Solvers}}}

\makeatletter
\renewcommand \thesection{S\@arabic\c@section}
\renewcommand\thetable{S\@arabic\c@table}
\renewcommand \thefigure{S\@arabic\c@figure}
\makeatother
\setcounter{figure}{0}
\setcounter{table}{0}
\setcounter{section}{0}
\setcounter{page}{1}

\normalsize

\section{Supplementary Methods \label{sec:SupplementaryMethods}}

\subsection{Krylov methods}

Krylov subspace methods are a broad family of iterative algorithms for solving large and sparse systems of linear equations
\[
A x = b, \qquad A \in \mathbb{R}^{n \times n}, \; x,b \in \mathbb{R}^n,
\]
as well as related eigenvalue and least-squares problems. These methods are based on the idea of constructing approximate solutions within a sequence of nested subspaces generated by repeated applications of the coefficient matrix \(A\) to the initial residual.

Given an initial approximation \(x^{(0)}\), the initial residual is defined as \(r^{(0)} = b - A x^{(0)}\). The corresponding Krylov subspace of dimension \(m\) is
\[
\mathcal{K}_m(A, r^{(0)}) = \mathrm{span}\{r^{(0)}, A r^{(0)}, A^2 r^{(0)}, \ldots, A^{m-1} r^{(0)}\}.
\]
The \(m\)-th iterate of a Krylov method, \(x^{(m)}\), is typically sought in the affine space
\[
x^{(m)} \in x^{(0)} + \mathcal{K}_m(A, r^{(0)}),
\]
subject to an optimality condition that depends on the specific algorithm, such as minimizing the residual or the energy norm of the error.

\paragraph{Conjugate Gradient (CG).}
The Conjugate Gradient method is designed for symmetric positive-definite (SPD) matrices. It can be viewed as a projection method that minimizes the quadratic functional
\[
\phi(x) = \frac{1}{2}x^\mathsf{T}A x - x^\mathsf{T}b
\]
over the Krylov subspace \(x^{(0)} + \mathcal{K}_m(A, r^{(0)})\). The method generates a sequence of mutually \(A\)-conjugate search directions and updates the solution and residual iteratively:
\[
x^{(m+1)} = x^{(m)} + \alpha_m p^{(m)}, \qquad r^{(m+1)} = r^{(m)} - \alpha_m A p^{(m)}.
\]
Here, \(\alpha_m\) is chosen to minimize \(\phi(x)\) along the search direction \(p^{(m)}\). The theoretical convergence rate depends on the condition number \(\kappa(A)\):
\[
\|e^{(m)}\|_A \le 2 \left( \frac{\sqrt{\kappa(A)} - 1}{\sqrt{\kappa(A)} + 1} \right)^m \|e^{(0)}\|_A,
\]
where \(e^{(m)} = x^{(m)} - x^\star\) denotes the current error. The rate of convergence thus improves significantly when \(A\) is well-conditioned or properly preconditioned.

\paragraph{Generalized Minimal Residual Method (GMRES).}
For non-symmetric or indefinite matrices, the Generalized Minimal Residual (GMRES) method is widely used. GMRES constructs an orthonormal basis of the Krylov subspace using the Arnoldi process and computes an approximate solution that minimizes the 2-norm of the residual:
\[
x^{(m)} = \arg\min_{x \in x^{(0)} + \mathcal{K}_m(A, r^{(0)})} \|b - A x\|_2.
\]
The method yields monotonically decreasing residual norms but requires storing all basis vectors, which leads to increasing memory and computational cost as \(m\) grows. Restarted versions, denoted GMRES(\(k\)), mitigate this by limiting the subspace dimension to \(k\), at the expense of slower convergence.

\paragraph{Biconjugate Gradient and Variants.}
Alternative Krylov methods such as BiCG, CGS (Conjugate Gradient Squared), BiCGSTAB (Bi-Conjugate Gradient Stabilized), and QMR (Quasi-Minimal Residual) are designed for non-symmetric matrices as well, typically requiring both the matrix \(A\) and its transpose \(A^\mathsf{T}\). These algorithms trade off between memory requirements, stability, and convergence smoothness.

\paragraph{Preconditioning.}
A crucial component of practical Krylov methods is preconditioning. The goal is to transform the system into an equivalent one with improved spectral properties, thereby accelerating convergence. This is achieved by introducing a preconditioner \(M \approx A^{-1}\) such that
\[
M^{-1} A x = M^{-1} b
\quad \text{(left preconditioning)}, \qquad
A M^{-1} y = b, \; x = M^{-1} y
\quad \text{(right preconditioning)}.
\]
An effective preconditioner clusters the eigenvalues of \(M^{-1}A\) and reduces the condition number. Common strategies include incomplete factorizations (ILU, IC), algebraic multigrid (AMG), domain decomposition, or sparse approximate inverses. The choice of preconditioner is problem-dependent and often dominates the performance of Krylov solvers.

\paragraph{Convergence and Stopping Criteria.}
Convergence is generally monitored using the residual norm, typically requiring
\[
\frac{\|r^{(m)}\|_2}{\|b\|_2} \le \varepsilon_{\mathrm{rel}}
\quad \text{or} \quad
\|r^{(m)}\|_2 \le \varepsilon_{\mathrm{abs}},
\]
where \(\varepsilon_{\mathrm{rel}}\) and \(\varepsilon_{\mathrm{abs}}\) are user-specified tolerances. Because Krylov methods are projection-based, they guarantee monotonic convergence in the appropriate norm for certain matrix classes (e.g., SPD for CG), though not necessarily for general matrices.

\paragraph{Computational Considerations.}
Each Krylov iteration typically requires one matrix--vector product, several vector inner products, and vector updates. The matrix--vector product is often the dominant cost, particularly for large-scale PDE discretizations where \(A\) represents a sparse operator. Memory requirements depend on the number of stored search directions: linear in the number of iterations for CG, and increasing linearly with the subspace size for GMRES and Arnoldi-based schemes.

\subsection{Neural Operators}

Neural operators are a class of machine learning architectures that aim to learn mappings between function spaces, providing a data-driven surrogate for solving families of partial differential equations (PDEs). Unlike traditional neural networks that approximate finite-dimensional maps (e.g.\ an image to a label), neural operators learn \emph{infinite-dimensional} mappings of the form
\[
\mathcal{G}: a(x) \mapsto u(x),
\]
where both the input \(a\) and output \(u\) are functions defined on a domain \(D \subset \mathbb{R}^d\). This formulation enables the learned model to generalize across different discretizations, meshes, and resolutions without retraining, making neural operators discretization-invariant by design.

\paragraph{Formulation.}
A neural operator replaces the matrix multiplications of standard neural networks by integral operators acting in function space. A general layer of a neural operator can be expressed as
\[
u_{\ell+1}(x) = \sigma \Bigl( W u_\ell(x) + \int_D \kappa_\theta(x,y)\,u_\ell(y)\,dy + b(x) \Bigr),
\]
where \(u_\ell\) is the function-valued feature at layer \(\ell\), \(W\) is a point-wise linear transformation, \(\kappa_\theta(x,y)\) is a trainable kernel parameterized by neural network weights \(\theta\), \(b(x)\) is a bias function, and \(\sigma(\cdot)\) is a nonlinear activation. The kernel integral defines a global coupling across the spatial domain, allowing the model to capture long-range interactions that are typical in PDE solution operators. Since the integral can be evaluated on arbitrary point sets, the model naturally extends across different discretizations and geometries.

\paragraph{Representative Architectures.}
Several practical neural operator architectures have been developed, each providing a different approach to approximate or accelerate the kernel integral:

\begin{itemize}
    \item \textbf{Graph Neural Operator (GNO).} Approximates the integral operator on irregular point clouds using graph-based message passing. Each node exchanges information with its neighbors within a specified radius, making the approach suitable for unstructured meshes or manifolds. The computational cost scales as \(O(J'^2)\), where \(J'\) is the number of retained neighbors.
    \item \textbf{Low-Rank Neural Operator (LNO).} Approximates the kernel as a low-rank decomposition,
    \[
    \kappa(x,y) \approx \sum_{r=1}^R \phi_r(x)\psi_r(y),
    \]
    resulting in a linear-time complexity \(O(J)\) and efficient representation for smooth kernels.
    \item \textbf{Multilevel Graph Neural Operator (MGNO).} Extends GNO by constructing a hierarchy of coarser graphs, similar to multigrid or fast multipole methods. This multi-scale architecture efficiently captures both local and global interactions while maintaining \(O(J)\) complexity.
    \item \textbf{Fourier Neural Operator (FNO).} Performs convolution in the Fourier domain using the Fast Fourier Transform (FFT):
    \[
    \mathcal{F}(u_{\ell+1})(k) = R_\theta(k)\, \mathcal{F}(u_\ell)(k),
    \]
    where \(R_\theta(k)\) is a learned transformation applied to a truncated set of Fourier modes. FNO achieves spectral accuracy on regular grids with a computational cost of \(O(J \log J)\).
\end{itemize}

Each variant can be seen as implementing a discretization-agnostic approximation of the same integral operator framework, differing only in how the kernel \(\kappa_\theta(x,y)\) is parameterized and evaluated.

\paragraph{Training Procedure.}
Neural operators are trained by supervised learning from pairs \((a, u)\) that represent the mapping from input coefficients (e.g.\ material properties, boundary conditions, or forcing terms) to PDE solutions. The training loop typically proceeds as follows:
\begin{enumerate}
    \item Sample a coefficient function \(a(x)\) from a prescribed distribution.
    \item Solve the PDE numerically to obtain the corresponding solution \(u(x)\).
    \item Evaluate the neural operator on \(a(x)\) and compute the predicted \(\hat{u}(x)\).
    \item Minimize a loss function such as the relative \(L^2\) error,
    \[
    \mathcal{L}(\theta) = \frac{\|u - \hat{u}\|_{L^2(D)}}{\|u\|_{L^2(D)}}.
    \]
\end{enumerate}
Once trained, the same model can be evaluated on unseen meshes, finer resolutions, or new geometries, owing to its continuous and mesh-independent formulation.

\paragraph{Interpretation and Applications.}
Neural operators provide a unifying framework for learning solution operators of PDEs directly from data. They can approximate mappings such as parameter-to-solution, forcing-to-solution, or boundary-to-solution relations. Their discretization invariance enables transfer across spatial resolutions and geometries, making them particularly suitable for surrogate modeling, uncertainty quantification, and real-time simulation.

\subsection{Variational Physics-Informed Neural Operator (VINO)} \label{sec:VINO}

The Variational Physics-Informed Neural Operator (VINO) provides a physics-consistent, data-free approach for learning mappings between function spaces based on the variational principles of partial differential equations (PDEs). Unlike data-driven neural operators or physics-informed formulations that rely on the strong form of the governing equations, VINO reformulates the learning task in the weak or variational form, leading to improved numerical stability, convergence, and generalization.

Traditional neural operators such as the Fourier Neural Operator (FNO) or DeepONet aim to approximate the solution operator \( \mathcal{G}: \mathcal{P} \to \mathcal{S} \) using a large number of paired data samples \( \{(p_i, s_i)\}_{i=1}^N \), where \(p\) represents a coefficient field or source term and \(s\) is the corresponding solution. Generating such training data typically requires repeatedly solving the governing PDEs numerically, which can be prohibitively expensive. Physics-Informed Neural Operators (PINO) were developed to mitigate this dependence by enforcing the governing equations directly through a loss function constructed from their strong form. However, this approach requires the computation of high-order derivatives through automatic differentiation, which can cause numerical instability and significantly increase computational cost.

The VINO framework resolves these issues by embedding the physics through a variational energy principle. Rather than enforcing the PDE at discrete spatial points, it minimizes an energy functional derived from the weak formulation of the governing equations. The true solution of a PDE corresponds to a stationary point of this functional, and the neural operator is trained to approximate the mapping that minimizes it. This removes the need for high-order derivatives, allows analytical treatment of derivatives and integrals, and enables training without reference data.

Consider a general parametric PDE of the form
\begin{equation}
    \mathcal{N}(p, s) = 0, \quad \text{in } \Omega,
    \label{eq:PDE_general}
\end{equation}
where \(p\) denotes the coefficients and \(s\) the solution field. If the PDE admits a variational principle, there exists a functional \(\Pi(s)\) such that the solution \(s^\ast\) satisfies the stationary condition
\begin{equation}
    \delta \Pi(s^\ast) = 0.
\end{equation}
Given an input \(p\), the neural operator \( \mathcal{G}_\theta(p) \) provides an approximation of \(s\). The loss function is then defined directly from the energy functional as
\begin{equation}
    \mathcal{L}_{\text{VINO}}(\theta)
    = \frac{1}{N} \sum_{i=1}^{N} 
    \left[
        \int_\Omega 
        \mathcal{F}\!\left(p^{(i)}, \mathcal{G}_{\theta}(p^{(i)}), \nabla \mathcal{G}_{\theta}(p^{(i)}) \right) d\Omega
        +
        \int_{\partial\Omega}
        \mathcal{E}\!\left(p^{(i)}, \mathcal{G}_{\theta}(p^{(i)}) \right) d\Gamma
    \right],
    \label{eq:VINO_loss}
\end{equation}
where the integrands \(\mathcal{F}\) and \(\mathcal{E}\) represent the energy densities in the domain and on the boundary, respectively. This formulation eliminates any dependence on labeled data, since the physics itself provides the learning signal.

To evaluate the energy terms efficiently, VINO employs a finite-element-inspired discretization. The computational domain is divided into small elements \(\Omega_e\), and the neural field \(\mathcal{G}_\theta(p)\) is expressed within each element as a linear combination of bilinear shape functions \(N_a(x, y)\):
\begin{equation}
    \mathcal{G}_\theta^e(p) = \sum_{a=1}^4 N_a(x, y) \, \tilde{s}_a,
    \label{eq:approx_field}
\end{equation}
where \(\tilde{s}_a\) denotes the nodal values predicted by the neural operator. For a rectangular element with sides \(2a\) and \(2b\), the shape functions are defined as
\begin{equation}
\begin{aligned}
    N_1 &= \tfrac{1}{4}\left(1-\tfrac{x'}{a}\right)\left(1-\tfrac{y'}{b}\right), \quad
    N_2 = \tfrac{1}{4}\left(1+\tfrac{x'}{a}\right)\left(1-\tfrac{y'}{b}\right), \\
    N_3 &= \tfrac{1}{4}\left(1+\tfrac{x'}{a}\right)\left(1+\tfrac{y'}{b}\right), \quad
    N_4 = \tfrac{1}{4}\left(1-\tfrac{x'}{a}\right)\left(1+\tfrac{y'}{b}\right),
\end{aligned}
\label{eq:shape_functions}
\end{equation}
where \(x' = x - x_0\) and \(y' = y - y_0\) denote the local coordinates relative to the element center. This representation allows all derivatives and integrals to be computed analytically, ensuring high precision and avoiding numerical quadrature or finite difference approximations. The total energy is then assembled over all elements as
\begin{equation}
    \mathcal{L}_{\text{VINO}}(\theta)
    = \sum_e \left[
        \int_{\Omega_e} \mathcal{F}\left(p, \mathcal{G}_{\theta}(p), \nabla \mathcal{G}_{\theta}(p)\right) d\Omega_e
        +
        \int_{\partial\Omega_e} \mathcal{E}\left(p, \mathcal{G}_{\theta}(p)\right) d\Gamma_e
    \right].
    \label{eq:element_energy}
\end{equation}

A simple example is provided by the Poisson equation
\begin{equation}
    -\nabla^2 \phi = f \quad \text{in } \Omega,
    \label{eq:Poisson_equation}
\end{equation}
whose variational formulation seeks to minimize the energy functional
\begin{equation}
    \Pi(\phi)
    = \frac{1}{2} \int_{\Omega} |\nabla \phi|^2 \, d\Omega
    - \int_{\Omega} f \phi \, d\Omega.
    \label{eq:Poisson_functional}
\end{equation}
In this case, the neural operator \(\mathcal{G}_\theta\) learns the mapping \(f \mapsto \phi\) by directly minimizing \(\Pi(\phi)\), without requiring any reference solutions.

The variational structure of VINO ensures that the learned operator respects the underlying physical principles of the governing equations. Because it relies solely on the energy functional, it is independent of training data, robust to discretization, and consistent under mesh refinement. Moreover, by avoiding the computation of high-order derivatives, it achieves superior numerical stability compared with strong-form physics-informed approaches. The combination of these properties makes VINO a powerful and general framework for operator learning governed by physical laws.

\subsection{Multi-Head Neural Operator (MHNO)}

The Multi-Head Neural Operator (MHNO) extends the framework of standard neural operators by introducing time-specific projection and coupling mechanisms designed to capture temporal dependencies in dynamical systems governed by partial differential equations. Instead of relying on a single global projection operator \(\mathcal{Q}\) that maps the final latent representation to the output field, MHNO replaces this step with a collection of time-specific neural networks \(\{\mathcal{Q}_n\}_{n=1}^{N_t}\), each responsible for producing the solution field at its corresponding time step \(n\). In addition, MHNO incorporates a second family of networks \(\{\mathcal{H}_n\}_{n=2}^{N_t}\), which explicitly encode the temporal interactions between consecutive time steps, allowing the model to represent complex temporal dynamics directly within the neural operator architecture.

The shared core of the structure of MHNO consists of a neural operator that processes the input field \(a(x)\) and encodes global spatial dynamics through a sequence of integral layers parameterized by \(\mathcal{K}_\ell\) and pointwise transformations \(\mathcal{W}_\ell\), interleaved with nonlinear activations \(\sigma\). These layers produce a sequence of intermediate latent fields \(\{v_\tau\}_{\tau=0}^{\mathcal{T}}\), representing the progressive transformation of the input into a high-dimensional feature space. The final latent representation \(v_\mathcal{T}(x)\) is then passed through time-step-specific projection networks \(\mathcal{Q}_n\) to yield the solution \(u_n(x)\) corresponding to the \(n\)-th time step. Unlike traditional neural operators, which apply a single projection to all time steps, MHNO allows each projection head \(\mathcal{Q}_n\) to specialize in the representation of its corresponding temporal slice, thereby improving expressivity and temporal resolution.

Formally, the output of MHNO at time step \(t_n\) is defined as
\begin{equation}
\mathcal{G}_\theta(x,t_n)(a(x)) := \mathcal{Q}_n \circ (\mathcal{W}_L + \mathcal{K}_L) \circ \cdots \circ \sigma(\mathcal{W}_1 + \mathcal{K}_1) \circ \mathcal{P} (a(x)) + \mathcal{H}_n \circ \mathcal{G}_\theta(x,t_{n-1})(a(x)),
\label{eq:MHNO}
\end{equation}
where \(\mathcal{P}\) denotes the lifting network mapping the input field to a higher-dimensional latent space, \(\mathcal{Q}_n\) and \(\mathcal{H}_n\) are the time-specific projection and temporal coupling networks respectively, and \(\mathcal{G}_\theta(x,t_0)\) is defined as the identity operator \(I(x)\). The operator \(\mathcal{H}_1\) is set to the zero operator \(\mathcal{O}(x)\) to ensure the correct initialization of the temporal chain. The additive term \(\mathcal{H}_n \circ \mathcal{G}_\theta(x,t_{n-1})\) establishes a direct dependency between consecutive time steps, introducing a form of neural message passing where the prediction at each time step explicitly incorporates the information propagated from the preceding one.

This formulation provides a principled mechanism to model temporal evolution within a unified operator framework. Each \(\mathcal{Q}_n\) is trained to specialize in its own temporal context, while \(\mathcal{H}_n\) captures the influence of preceding dynamics. The use of distinct neural networks for each time step allows the model to represent time-dependent solution operators with greater flexibility than architectures that rely on a single, global projection layer. In effect, MHNO generalizes the standard neural operator by embedding temporal connections directly into the operator’s functional mapping, enabling accurate modeling of long-term temporal dependencies without requiring recurrent or autoregressive structures.

A restricted subclass of MHNOs can be defined by removing all temporal couplings, achieved by setting \(\mathcal{H}_n = \mathcal{O}\) for all \(n\). In this case, MHNO reduces to a collection of independent neural operators, each with its own projection head \(\mathcal{Q}_n\):
\begin{equation}
\mathcal{G}_\theta(x,t_n)(a(x)) := \mathcal{Q}_n \circ (\mathcal{W}_L + \mathcal{K}_L) \circ \cdots \circ \sigma(\mathcal{W}_1 + \mathcal{K}_1) \circ \mathcal{P} (a(x)).
\end{equation}
If all \(\mathcal{Q}_n\) are further constrained to be identical, \(\mathcal{Q}_n = \mathcal{Q}\), the architecture reduces exactly to the standard neural operator formulation.

\subsection{Pseudo-code of Neural Operator Warm Starts (NOWS)}

NOWS framework integrates learned neural operators with conventional iterative PDE solvers. The neural operator provides an accurate initial guess that rapidly reduces the residual norm, enabling the iterative solver to converge in fewer steps while maintaining the guarantees of the underlying numerical method. This hybrid paradigm requires no modification to existing solver infrastructure, preconditioning strategy, or discretization scheme, thus remaining fully compatible with standard methods such as finite element, finite difference, finite volume, and isogeometric analysis.

The key stages of the NOWS workflow are (1) data generation and training of the neural operator, (2) inference to generate warm-start initializations, and (3) iterative refinement through a classical solver. The neural operator captures global solution structures efficiently, while the iterative solver ensures local accuracy and stability. Below we summarize the general pseudo-code of the NOWS algorithm.

\begin{algorithm}[H]
\caption{Pseudo-code of Neural Operator Warm Starts (NOWS)}
\label{alg:nows}
\begin{algorithmic}[1]
\Require PDE operator $\mathcal{L}(u; \theta) = f$, discretization scheme $\mathcal{D}$, iterative solver $\mathcal{S}$, trained neural operator $\mathcal{G}_\phi$, tolerance $\varepsilon$
\Ensure Numerical solution $u^*$ approximating $\mathcal{L}(u) = f$
\Statex
\State \textbf{Offline phase: Training of neural operator}
\State Generate dataset $\{(a_i, u_i)\}_{i=1}^N$ by solving $\mathcal{L}(u_i; a_i) = f_i$ with a classical solver.
\State Train neural operator $\mathcal{G}_\phi: a \mapsto u$ to minimize loss $\mathcal{L}_{train} = \|u_i - \mathcal{G}_\phi(a_i)\|_{L^2}$.
\State Store trained weights $\phi$ for inference.
\Statex
\State \textbf{Online phase: Neural Operator Warm Start}
\State Given a new input coefficient field $a(x)$, compute neural prediction:
\[
u_0 = \mathcal{G}_\phi(a)
\]
\State Initialize the iterative solver $\mathcal{S}$ with $u_0$ as the initial guess.
\State Compute residual $r_0 = f - \mathcal{L}(u_0)$.
\While{$\|r_k\| > \varepsilon$}
    \State Update $u_{k+1} = \mathcal{S}(u_k, r_k)$
    \State Recompute residual $r_{k+1} = f - \mathcal{L}(u_{k+1})$
\EndWhile
\State \Return $u^* = u_{k+1}$
\end{algorithmic}
\end{algorithm}

\noindent
The above pseudo-code captures the solver-agnostic nature of NOWS: any iterative algorithm (e.g., conjugate gradient, GMRES, multigrid) can be integrated without modification. The neural operator $\mathcal{G}_\phi$ serves as a data-driven preconditioner that provides an informed initialization. The iterative solver then performs only fine-scale corrections to reach the final converged solution. 

This general pseudo-code can be readily adapted to various elliptic, parabolic, or hyperbolic PDEs by modifying the neural operator architecture or the underlying iterative solver. The modularity of the NOWS framework makes it a broadly applicable hybrid paradigm for accelerating high-fidelity PDE simulations while retaining full numerical rigor.

\section{Supplementary Tables \label{sec:SupplementaryTables}}

\subsection{Summary of notation}

Table~\ref{tab:notation} summarizes the main mathematical symbols and operators used throughout the manuscript. The notation follows standard conventions in partial differential equations, operator learning, and numerical linear algebra.

\begin{table}[H]
\centering
\caption{\textbf{Summary of notation.} Key symbols and their definitions used throughout the paper.}
\label{tab:notation}
\begin{tabular}{ll}
\toprule
\textbf{Symbol} & \textbf{Description} \\
\midrule
$D \subset \mathbb{R}^d$ & Spatial domain of the PDE \\
$\partial D$ & Boundary of the domain $D$ \\
$a \in \mathcal{A}$ & Coefficient or parameter field of the PDE \\
$f \in \mathcal{U}^\ast$ & Forcing or source term \\
$u : D \to \mathbb{R}$ & Solution field \\
$\mathcal{U}$ & Function (Banach) space of solutions \\
$\mathcal{A}$ & Function space of coefficients or parameters \\
$\mathcal{L}(u; \theta)$ & Parametric differential operator \\
$L_a$ & Linearized differential operator associated with $a$ \\
$\mathcal{G}^\dagger = L_a^{-1}$ & True solution operator mapping $a \mapsto u$ \\
$\mathcal{G}_\theta$ & Neural operator approximation with parameters $\theta$ \\
$N$ & Number of training samples \\
$K$ & Number of discretization points in the domain \\
$u_0$ & Initial iterate for the numerical solver \\
$u_k$ & Solution iterate at step $k$ in the iterative scheme \\
$\nu$ & Viscosity coefficient (Burgers’ and Navier–Stokes equations) \\
$p$ & Pressure field (fluid simulations) \\
$\rho$ & Density or transported scalar (smoke plume) \\
$\mathbf{u}$ & Velocity field \\
$\mathbf{g}$ & Gravitational acceleration vector \\
$\beta$ & Buoyancy coefficient \\
$D$ (in transport term) & Diffusion coefficient for scalar transport \\
$\mathbf{D}$ & Constitutive (material) matrix in elasticity \\
$\boldsymbol{\sigma}$ & Stress tensor \\
$\boldsymbol{\varepsilon}$ & Strain tensor \\
$\mathcal{G}: p \mapsto s$ & Operator mapping permeability field to pressure (Darcy flow) \\
CG & Conjugate Gradient method \\
GMRES & Generalized Minimal Residual method \\
ICC, ILU, SSOR & Standard preconditioners for Krylov methods \\
VINO & Variational physics-informed neural operator \\
FNO & Fourier Neural Operator \\
DeepONet & Deep Operator Network \\
\bottomrule
\end{tabular}
\end{table}

\subsection{Summary of hyperparameters}

Table~\ref{tab:hyperparameters} summarizes the best-performing hyperparameter configurations for each model across different problems. The hyperparameters shown, such as batch size, learning rate, network width, and depth, are the ones that yielded the highest accuracy in our experiments. 
\begin{table}[H]
    \centering
    \caption{Best-performing hyperparameter configurations for each model across various PDE problems and temporal approaches. Each entry reports the selected values for different hyperparameters. Values in parentheses correspond to different temporal approaches.}
    \label{tab:hyperparameters}
    \resizebox{\textwidth}{!}{%
    \begin{tabular}{l|ccccccccccccccc}
\multicolumn{1}{c|}{Problem} & \textbf{Neural Operator} & \textbf{Resolution} & \textbf{\(N_{\text{train}}\)} & \textbf{\(N_{\text{test}}\)} & \textbf{\(Bs\)} & \textbf{\(Lr\)} & \textbf{Epoch}   & \textbf{\(M\)} & \textbf{\(W\)} & \textbf{\(W_\mathcal{Q}\)} & \textbf{\(W_\mathcal{H}\)} & \textbf{\(N_l\)} & \textbf{\(N_\mathcal{Q}\)} & \textbf{\(N_\mathcal{H}\)} \\ \hline
\multirow{4}{*}{Poisson} 
& VINO   & 64 $\times$ 64 & 1000 & 2000 & 100 & 0.001 & 5000  & 24 & 32 & 128  & -   & 4   & 2  & -  \\
& VINO   & 128 $\times$ 128 & 1000 & 2000 & 50 & 0.001 & 5000  & 20 & 32 & 128  & -   & 4   & 2  & -  \\
& VINO   & 256 $\times$ 256 & 1000 & 2000 & 50 & 0.001 & 5000  & 28 & 32 & 128  & -   & 4   & 2  & -  \\
& VINO   & 512 $\times$ 512 & 1000 & 2000 & 25 & 0.001 & 5000  & 32 & 32 & 128  & -   & 4   & 2  & -  \\\hline

\multirow{6}{*}{Darcy}
& FNO   & 64 $\times$ 64 & 1000 & 100 & 100 & 0.001 & 2000  & 28 & 32 & 128  & -   & 4   & 2  & -  \\
& FNO & 128 $\times$ 128 & 1000 & 100 & 100 & 0.001 & 2000  & 32 & 32 & 128  & -   & 4   & 2  & -  \\
& FNO & 256 $\times$ 256 & 1000 & 100 & 50 & 0.001 & 2000  & 28 & 32 & 128  & -   & 4   & 2  & -   \\
& VINO  & 64 $\times$ 64 & 1000 & 100 & 100 & 0.001 & 2000  & 28 & 32 & 128  & -   & 4   & 2  & -  \\
& VINO & 128 $\times$ 128 & 1000 & 100 & 100 & 0.001 & 2000  & 32 & 32 & 128  & -   & 4   & 2  & -  \\
& VINO & 256 $\times$ 256 & 1000 & 100 & 50 & 0.001 & 2000  & 28 & 32 & 128  & -   & 4   & 2  & -  \\ \hline

\multirow{1}{*}{\begin{tabular}[c]{@{}l@{}} Plate with voids\end{tabular}}
& VINO   & 200 $\times$ 200 & 1000  & 200 & 20 & 0.001 & 1000  & 8 & 32 & 128 & - & 8 & 2 & - \\ \hline

\multirow{1}{*}{\begin{tabular}[c]{@{}l@{}} Burgers \end{tabular}}
& MHNO   & 1024 & 1000  & 100 & 20 & 0.001 & 500  & 16 & 64 & 128 & 64 & 4 & 2 & 2 \\ \hline

\multirow{1}{*}{\begin{tabular}[c]{@{}l@{}} Smoke Plume \end{tabular}}
& FNO   & 64 $\times$ 64 & 2400 & 600 & 100 & 0.0001 & 1000  & 16 & 32 & 32 & - & 6 & 2 & - \\ \hline

\end{tabular}%
    }
\end{table}

\subsection{Summary of computational performance}
Table \ref{tab:benchmark_timing_all} reports the detailed timing results for the considered test cases. An important feature of the proposed approach is that the neural operator
can be trained on relatively low-resolution data and subsequently applied
to significantly higher resolutions without retraining.
For example, a model trained using $64\times64$ simulations can be used to
accelerate solves on $128\times128$, $256\times256$, and $512\times512$ grids.
Consequently, the offline cost of database generation and training does not
need to be repeated for each resolution.

Using the $64\times64$ Poisson case as the training cost, the break-even
number of simulations required to amortize the offline cost is approximately
188 runs for $64\times64$, 20 runs for $128\times128$, 3 runs for $256\times256$,
and only a single simulation for $512\times512$.
This demonstrates that the method becomes highly advantageous for
large-scale simulations or many-query scenarios.

\begin{table}[H]
\centering
\caption{\textbf{Comprehensive timing analysis across all benchmark problems and resolutions.}
Each example lists representative mesh or grid resolutions along with database generation, network training, inference, and solver runtimes for the baseline CG and NOWS-accelerated simulations. 
NOWS consistently reduces solver wall-clock time across all resolutions while maintaining accuracy.}
\begin{tabular}{llccccccc}
\toprule
\textbf{Example} & \textbf{Resolution} & \textbf{Samples} & \textbf{Database (s)} & \textbf{Training (s)} & \textbf{Inference (s)} & \textbf{CG (s)} & \textbf{NOWS (s)} & \textbf{Speedup} \\
\midrule
\multirow{1}{*}{Poisson} 
 & $64 \times 64$   & 1000 & 19.95 & 511.33 & 2.1 & 11.92 & 9.09 & 23.7 \% \\
 & $128 \times 128$ & 1000 & 25.72 & 1798.31 & 2.15 & 90.34 & 63.60 & 29.6 \% \\
 & $256 \times 256$ & 1000 & 46.55 & 6416.46 & 2.31 & 827.15 & 631.94 & 23.6 \% \\
 & $512 \times 512$ & 1000 & 143.01 & 26454.66 & 2.62 & 5810.51 & 3962.76 & 31.8 \%  \\
\midrule
\multirow{1}{*}{Darcy Flow} 
 & $64 \times 64$ & 1000 & 241.61 & 668.47 & 2.37 & 13.8 & 11.7 & 15.22 \% \\
 & $128 \times 128$ & 1000 & 1033.46 & 1022.16 & 2.40 & 126.8 & 102.32 & 19.30 \% \\
 & $256 \times 256$ & 1000 & 4287.05 & 3874.47 & 2.82 & 905.79 & 736.95 & 18.64 \% \\
\midrule
\multirow{1}{*}{Plate \& Voids} 
 & $200 \times 200$ & 1000 & 180335.50 & 8480.28 & 119.90 & 20366.50 & 2675.02 & 86.86 \% \\
\midrule
\multirow{1}{*}{Burgers} 
 & $1024$   & 1000 & 697.23 & 475.91 & 1.65 & 236.40 & 116.86 & 50.57 \% \\
\midrule
\multirow{1}{*}{Smoke Plume} 
 & $64 \times 64$   & 20 & 5894.8 & 903.23 & 4.65 & 97.37 & 50.16 & 48.5 \% \\
 & $256 \times 256$ & 20 & 5894.8 & 903.23 & 4.95 & 290.38 & 74.75 & 74.3 \% \\
\bottomrule
\end{tabular}
\label{tab:benchmark_timing_all}
\end{table}

\subsection{Break-even and amortization analysis}

To assess the practical benefit of the proposed framework, we estimate the number of solves required to amortize the offline cost associated with database generation and neural operator training.

Let the total offline cost be
\[
T_{\mathrm{offline}} = T_{\mathrm{database}} + T_{\mathrm{training}},
\]
and the runtime reduction achieved per solve be
\[
\Delta T = T_{\mathrm{CG}} - T_{\mathrm{NOWS}}.
\]
The break-even number of solves is therefore
\[
N_{\mathrm{BE}} = \frac{T_{\mathrm{offline}}}{\Delta T}.
\]

A key feature of the proposed approach is that the neural operator can be trained on relatively low-resolution simulations and applied to higher-resolution problems without retraining. Consequently, the offline cost depends only on the \emph{training resolution}, while the runtime savings depend on the \emph{inference resolution}.

The following tables summarize the break-even number of solves for different combinations of training and inference resolutions.

\subsubsection*{Poisson equation}

\begin{table}[H]
\centering
\caption{Break-even number of solves for the Poisson equation. Rows correspond to training resolution, columns correspond to inference resolution.}
\begin{tabular}{lcccc}
\toprule
Training $\downarrow$ / Inference $\rightarrow$ & 64$\times$64 & 128$\times$128 & 256$\times$256 & 512$\times$512 \\
\midrule
64$\times$64 & 188 & 20 & 3 & $<$1 \\
128$\times$128 & -- & 68 & 9 & 1 \\
256$\times$256 & -- & -- & 33 & 4 \\
512$\times$512 & -- & -- & -- & 14 \\
\bottomrule
\end{tabular}
\end{table}

\subsubsection*{Darcy flow}

\begin{table}[H]
\centering
\caption{Break-even number of solves for Darcy flow. Rows correspond to training resolution, columns correspond to inference resolution.}
\begin{tabular}{lccc}
\toprule
Training $\downarrow$ / Inference $\rightarrow$ & 64$\times$64 & 128$\times$128 & 256$\times$256 \\
\midrule
64$\times$64 & 434 & 37 & 5 \\
128$\times$128 & -- & 84 & 12 \\
256$\times$256 & -- & -- & 48 \\
\bottomrule
\end{tabular}
\end{table}

\subsubsection*{Additional benchmarks}

\begin{table}[H]
\centering
\caption{Break-even number of solves for Plate with Voids, Burgers equation, and Smoke Plume.}
\begin{tabular}{lccc}
\toprule
Problem & Training resolution & Inference resolution & Break-even solves \\
\midrule
Plate with Voids & 200$\times$200 & 200$\times$200 & 11 \\
Burgers equation & 1024 & 1024 & 10 \\
Smoke Plume & 64$\times$64 & 64$\times$64 & 144 \\
Smoke Plume & 64$\times$64 & 256$\times$256 & 32 \\
\bottomrule
\end{tabular}
\end{table}

Several trends can be observed. The amortization threshold decreases significantly as the inference resolution increases, because the solver runtime grows rapidly with problem size. Training at relatively low resolution can dramatically reduce the effective break-even point when the model is applied to higher-resolution simulations. For example, in the Poisson benchmark a model trained on 64$\times$64 data requires approximately 188 solves to amortize at the same resolution, but only 3 solves when used to accelerate 256$\times$256 simulations. These results indicate that the proposed framework is particularly advantageous in many-query settings involving high-resolution simulations, such as parametric studies, uncertainty quantification, optimization, and inverse problems.

\section{Supplementary Result}  \label{sec:SupplementaryResult}
\subsection*{S1. Poisson Equation (2D)}

To evaluate the effectiveness of the proposed NOWS framework in accelerating classical iterative solvers, we first considered the canonical 2D Poisson problem:
\begin{equation}
    -\nabla^2 u(x, y) = f(x, y), \quad (x, y) \in (0,1)^2,
\end{equation}
subject to homogeneous Dirichlet boundary conditions \(u=0\) on \(\partial \Omega\).
The forcing term \(f(x,y)\) was modeled as a realization of a Gaussian Random Field (GRF) with a prescribed correlation length, and the analytical solution \(u(x,y)\) was computed using a finite-difference discretization of the Laplace operator.

\paragraph{Data generation.}
The random forcing fields were generated using GRF, which samples a mean-zero Gaussian process with an RBF kernel,
\begin{equation}
    k(\mathbf{x}, \mathbf{x}') = \exp\left(-\frac{\|\mathbf{x} - \mathbf{x}'\|^2}{2l^2}\right),
\end{equation}
where \(l\) is the length scale, here fixed to \(l = 0.1\).
A set of \(N\) realizations was produced, where each forcing field is interpolated onto an \(m\times m\) uniform grid.
For each realization, the corresponding exact solution was obtained by solving the discrete Poisson system
\begin{equation}
    \mathbf{A} \mathbf{u} = \mathbf{f},
\end{equation}
using the sparse direct solver, where \(\mathbf{A}\) is the standard 5-point finite-difference Laplacian matrix on a unit square domain. This provides reference solutions for training and testing.

\paragraph{Iterative solver and warm starts.}
For iterative benchmarking, we used a Conjugate Gradient (CG) solver.
This routine records both the residual trajectory and the total wall-clock time.
The solver can be initialized either from a zero field (cold start) or with a neural operator prediction (NOWS warm start).
Convergence is declared when the relative residual satisfies \(\|\mathbf{r}_k\| / \|\mathbf{r}_0\| < 10^{-5}\).

\paragraph{Neural operator architecture.}
The neural surrogate employed is VINO. The architecture consists of four spectral convolution layers followed by local nonlinear transformations and skip connections. Each spectral layer performs a forward 2D FFT, a learned complex-valued multiplication of the lowest Fourier modes, an inverse FFT to return to the spatial domain. The input \(f(x,y)\) is concatenated with spatial coordinates \((x,y)\) and lifted to a high-dimensional latent space via a linear layer.
The network width was set to 32, and 24 Fourier modes for resolution \(64\times64\) were retained in each direction. The total number of trainable parameters was approximately \(9.45 \times 10^6\).

\begin{figure}[h!]
    \centering
    \includegraphics[width=0.45\textwidth]{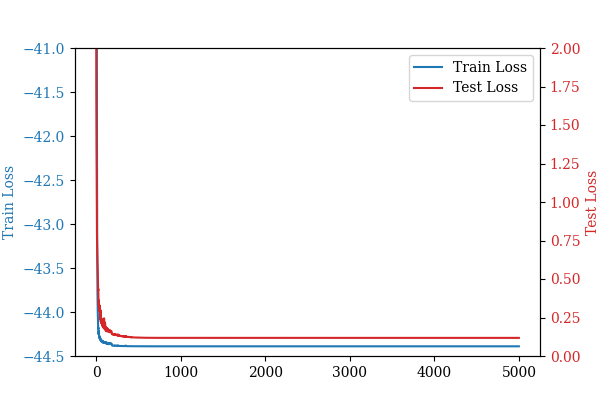}
    \includegraphics[width=0.45\textwidth]{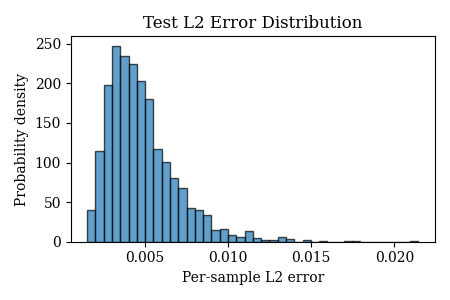}
    \caption{ 
    (a) Loss function trajectory based on iteration
    (b) Distribution of per-sample \(L^2\) test errors for the VINO model.}
    \label{fig:loss}
\end{figure}

\paragraph{Training setup.}
For \(64\times64\) grid as a sample, the model was trained on \(N_{\text{train}}=1000\) samples and tested on \(N_{\text{test}}=2000\) samples generated from the GRF ensemble. We used a learning rate of \(10^{-3}\), batch size of 100, and trained for up to 5000 epochs with a learning-rate scheduler (\(\gamma = 0.5\)) and patience of 50 epochs.
The loss function was a mixed data-PDE variational energy functional composed of the physical residual loss and the data loss:
\begin{equation}
    \mathcal{L} = \| \nabla u_{\theta}\|_2^2 - \langle f, u_{\theta} \rangle + \lambda \|u_{\theta} - u_{\text{true}}\|_2^2,
\end{equation}
The loss function reduction is shown in Fig.~\ref{fig:loss}.a.

\begin{figure}[h!]
    \centering
    \includegraphics[width=0.45\textwidth]{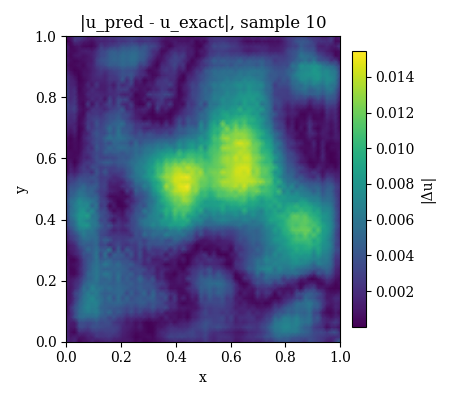}
    \includegraphics[width=0.45\textwidth]{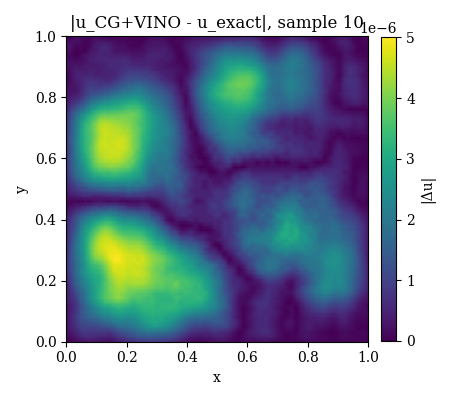}
    \caption{ 
    (a) Example spatial map of the absolute error field for VINO. 
    (b) Example spatial map of the absolute error field for NOWS.}
    \label{fig:error}
\end{figure}

\paragraph{Results.}
For the \(64\times64\) grid, the pretrained model achieved a \textbf{mean relative \(L^2\) error} of \(4.83\times10^{-3}\) on the test set, with a \textbf{standard deviation} of \(2.13\times10^{-3}\).
The minimum and maximum errors were \(1.53\times10^{-3}\) and \(2.10\times10^{-2}\), respectively.
The median relative error was \(4.36\times10^{-3}\), and the average pointwise maximum error was \(1.74\times10^{-2}\).
The error distribution (Supplementary Fig.~\ref{fig:loss}.b) is narrowly concentrated around the mean, indicating stable generalization across stochastic forcings. Spatial error field has been shown in (Supplementary Fig.~\ref{fig:error}.a).

\begin{figure}[h!]
    \centering
    \includegraphics[width=0.45\textwidth]{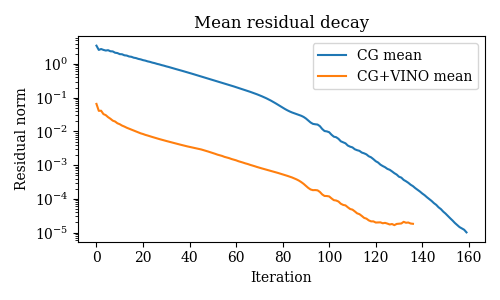}
    \includegraphics[width=0.45\textwidth]{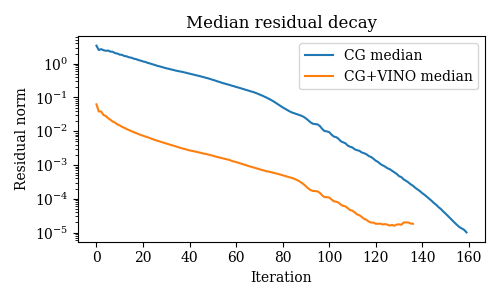}
    \caption{
    Mean and median residual-norm decay for CG and CG+NOWS.}
    \label{fig:residual}
\end{figure}

\paragraph{Iterative solver acceleration.}
When coupled with the CG solver, NOWS provided a consistent reduction in computational time.
The conventional CG solver required an average wall time of \(33.89\,\text{s}\), while initializing with NOWS reduced this to \(25.83\,\text{s}\), corresponding to a \textbf{23.7\% reduction in time}.
The solver’s core iteration time decreased similarly from \(33.71\,\text{s}\) to \(25.68\,\text{s}\).
The mean and median relative savings were both \(23.79\%\).
Residual-norm trajectories (Supplementary Fig.~\ref{fig:residual}) demonstrate that CG+NOWS starts approximately two orders of magnitude closer to convergence and reaches the prescribed tolerance roughly 25\% faster.

\begin{figure}[h!]
    \centering
    \includegraphics[width=0.45\textwidth]{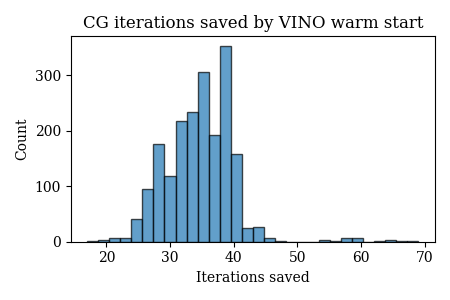}
    \includegraphics[width=0.45\textwidth]{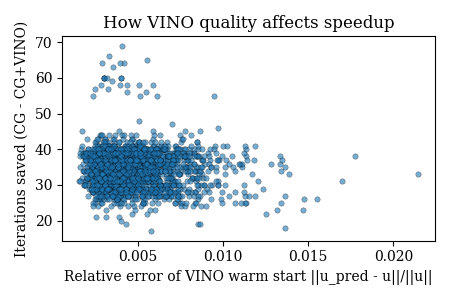}
    \caption{
    (a) Histogram of iteration savings
    (b) Correlation between warm-start error and solver iteration reduction.
    }
    \label{fig:correlation}
\end{figure}

The correlation between the neural prediction error and iteration savings was slightly negative (\(\rho=-0.114\)), suggesting that better-quality warm starts yield marginally higher speedups (Supplementary Fig.~\ref{fig:correlation}.b).
The iteration-saving histogram (Supplementary Fig.~\ref{fig:correlation}.a) confirms consistent acceleration across all samples without any loss of final accuracy.
The final relative \(L^2\) error after CG+NOWS refinement was \(1.15\times10^{-6}\), nearly identical to the fully converged solution (\(7.9\times10^{-7}\)). The pointwise error of the NOWS result for one sample is presented in Fig.~\ref{fig:error}.b).

\subsection*{Darcy Flow in Heterogeneous Media}

The second benchmark examines the performance of NOWS on steady-state Darcy flow through heterogeneous porous media. The governing equation is
\begin{equation}
    -\nabla \cdot (a(x, y) \nabla u(x, y)) = f(x, y), \qquad (x, y) \in (0,1)^2,
\end{equation}
with Dirichlet boundary conditions \(u(0,y)=u(x,0)=0\), a fixed forcing \(f(x,y)=1\), and a spatially varying diffusion coefficient \(a(x,y)\).
The objective is to learn the nonlinear operator mapping
\(\mathcal{G}: a(x,y) \mapsto u(x,y)\)
and to assess the effect of NOWS on iterative solvers for the resulting elliptic system.

\begin{figure}[h!]
    \centering
    \includegraphics[width=0.33\textwidth]{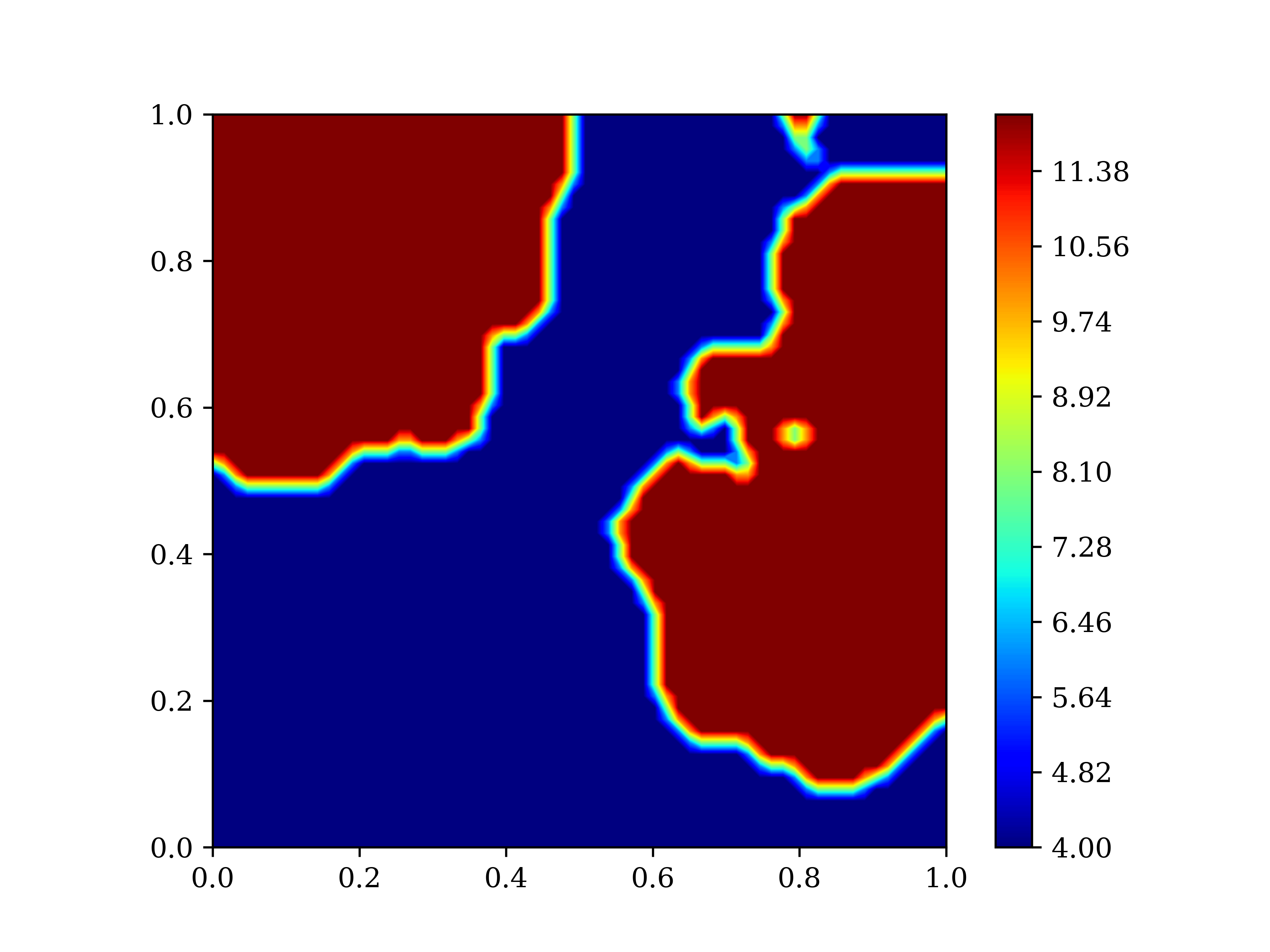}
    \includegraphics[width=0.33\textwidth]{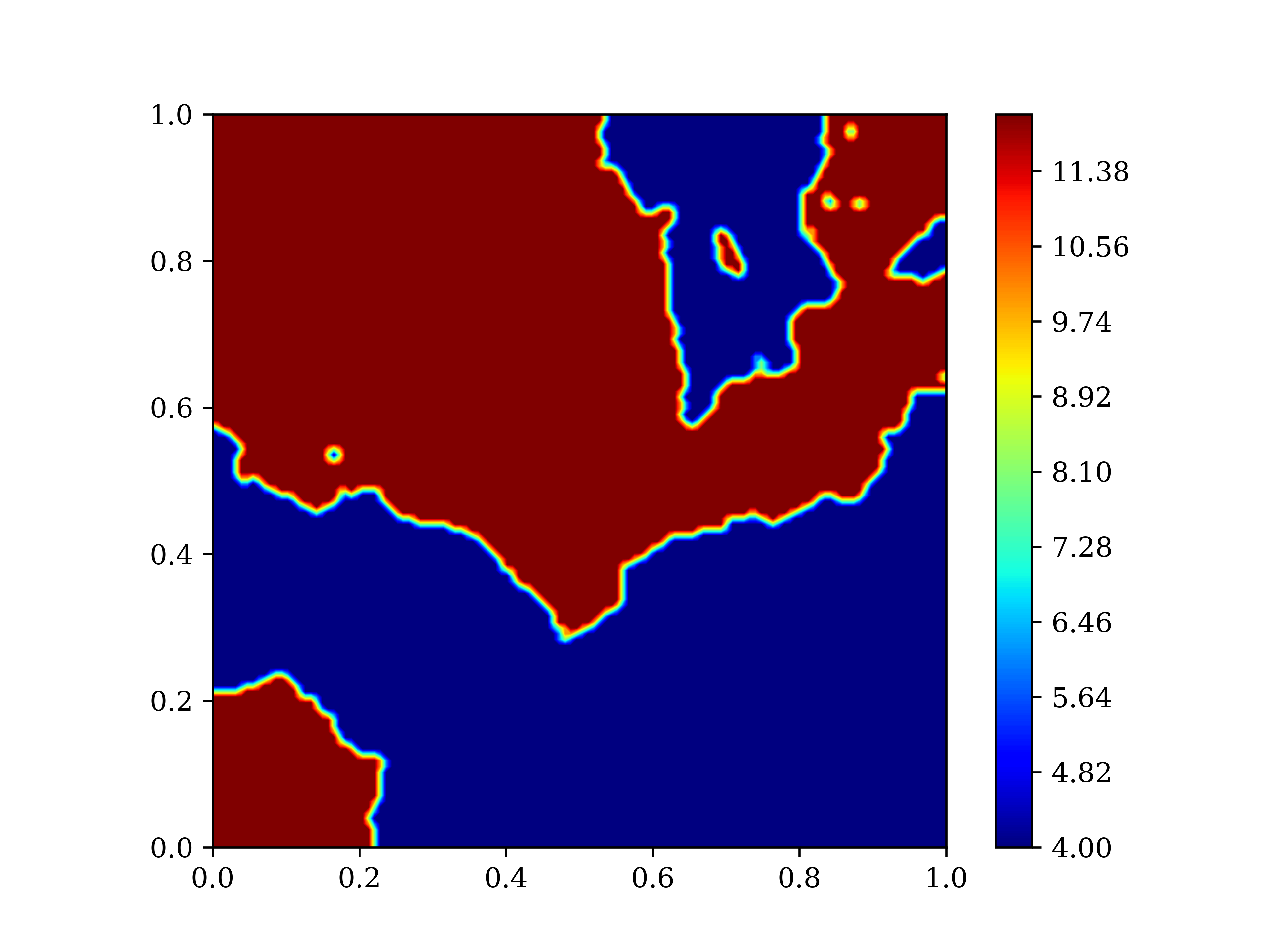}
    \includegraphics[width=0.33\textwidth]{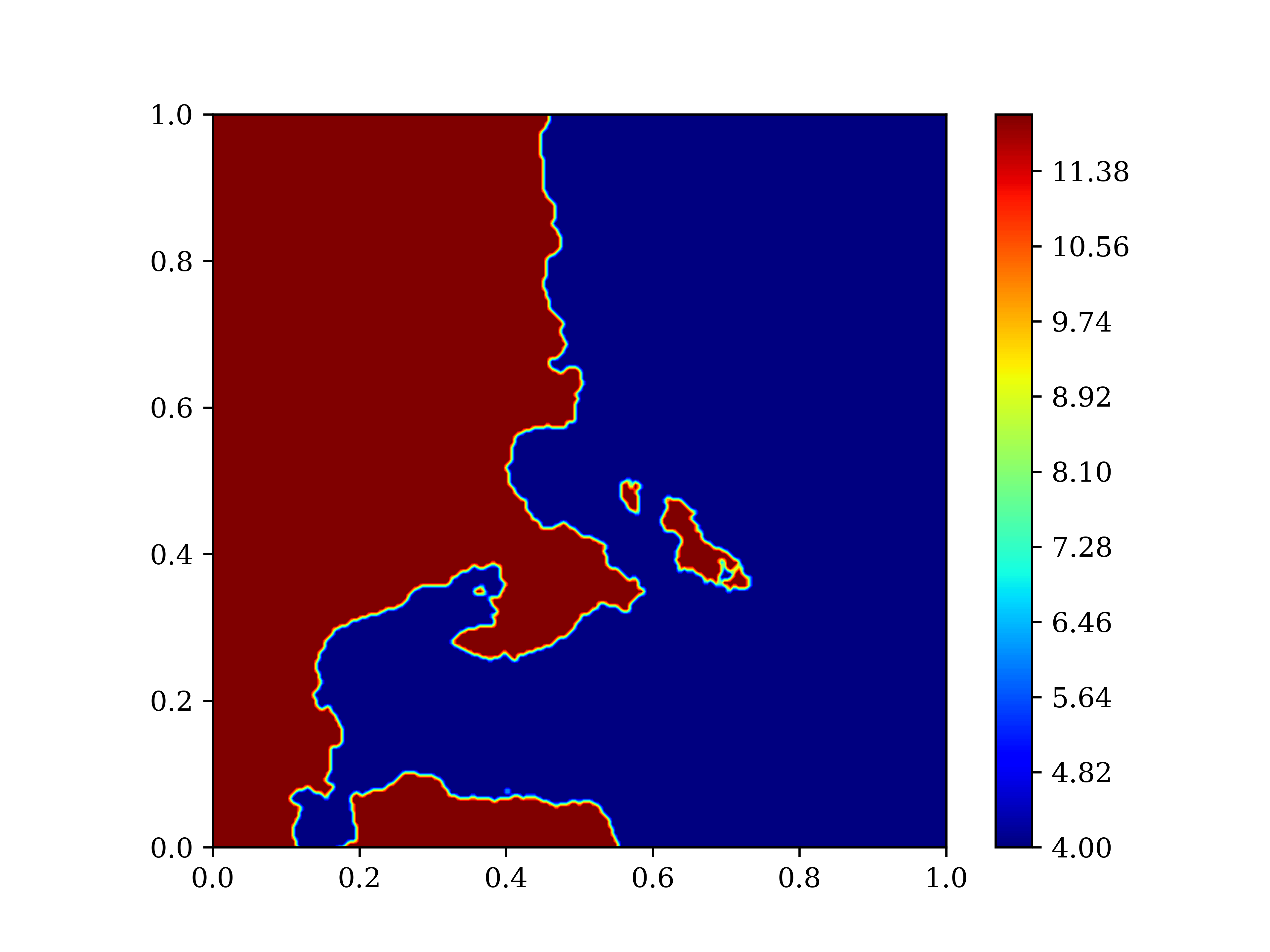}
    \caption{Representative samples for resolutions size 64, 128, and 256, respectively}
    \label{fig:darcy_input_GRF}
\end{figure}

\paragraph{Data generation.}
Random diffusion fields were generated using the function, which samples permeability distributions of the form
\[
a(x, y) = \exp(\alpha\, g_\tau(x, y)),
\]
where \(g_\tau(x,y)\) is a Gaussian random field with correlation length \(\tau\). Some samples from the test data set for different resolutions have been shown in Fig.~\ref{fig:darcy_input_GRF}. This representation yields heterogeneous media with controllable contrast ratio. The forward solutions were computed using finite element solvers, and for iterative benchmarking, we employed the conjugate gradient (CG) and preconditioned CG variants. 

\paragraph{Network architecture and training.}
We employed the variational physics-informed neural operator (VINO) that embeds boundary-condition consistency by multiplying the network output with the bilinear term
\[
m(x, y) = x\,y\,(x-1)\,(y-1),
\]
enforcing homogeneous Dirichlet constraints.
For grid \(64\times64 \), the network uses 28 Fourier modes per dimension and width 32, totaling approximately \(9.1\times10^6\) parameters.
Training was conducted for 2000 epochs with learning rate \(10^{-3}\), batch size 100, and adaptive learning rate scheduling with patience 50.
The variational (energy-based) loss combines PDE residual and data terms:
\begin{equation}
    \mathcal{L} = \int_{\Omega} \tfrac{1}{2} a(x,y) |\nabla u_\theta|^2 - f(x,y)u_\theta \, dx\,dy
    + \lambda \|u_\theta - u_{\text{true}}\|_2^2,
\end{equation}
where \(\lambda\) controls the data-physics tradeoff.

\begin{figure}[h!]
    \centering
    \includegraphics[width=0.33\textwidth]{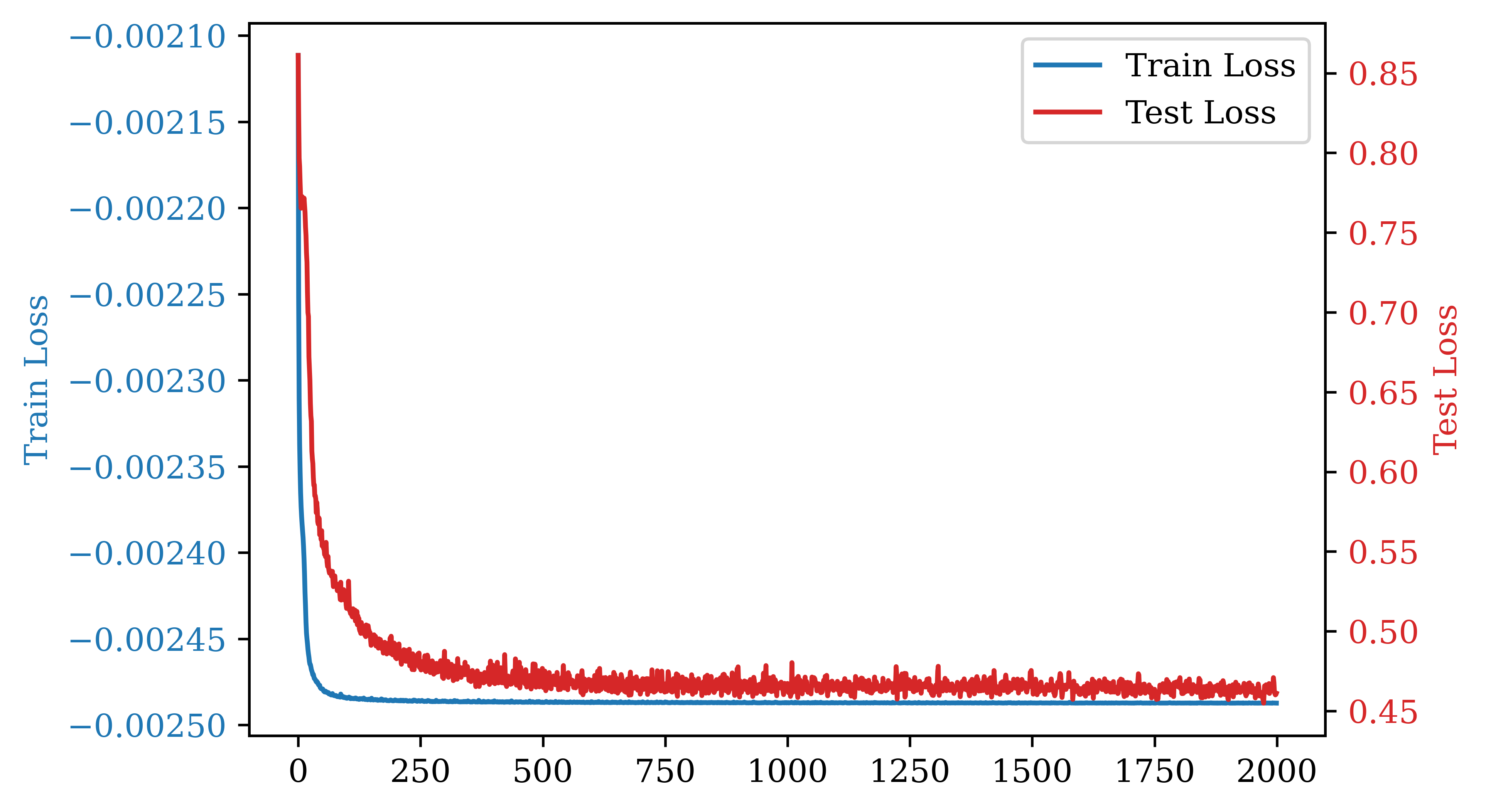}
    \includegraphics[width=0.33\textwidth]{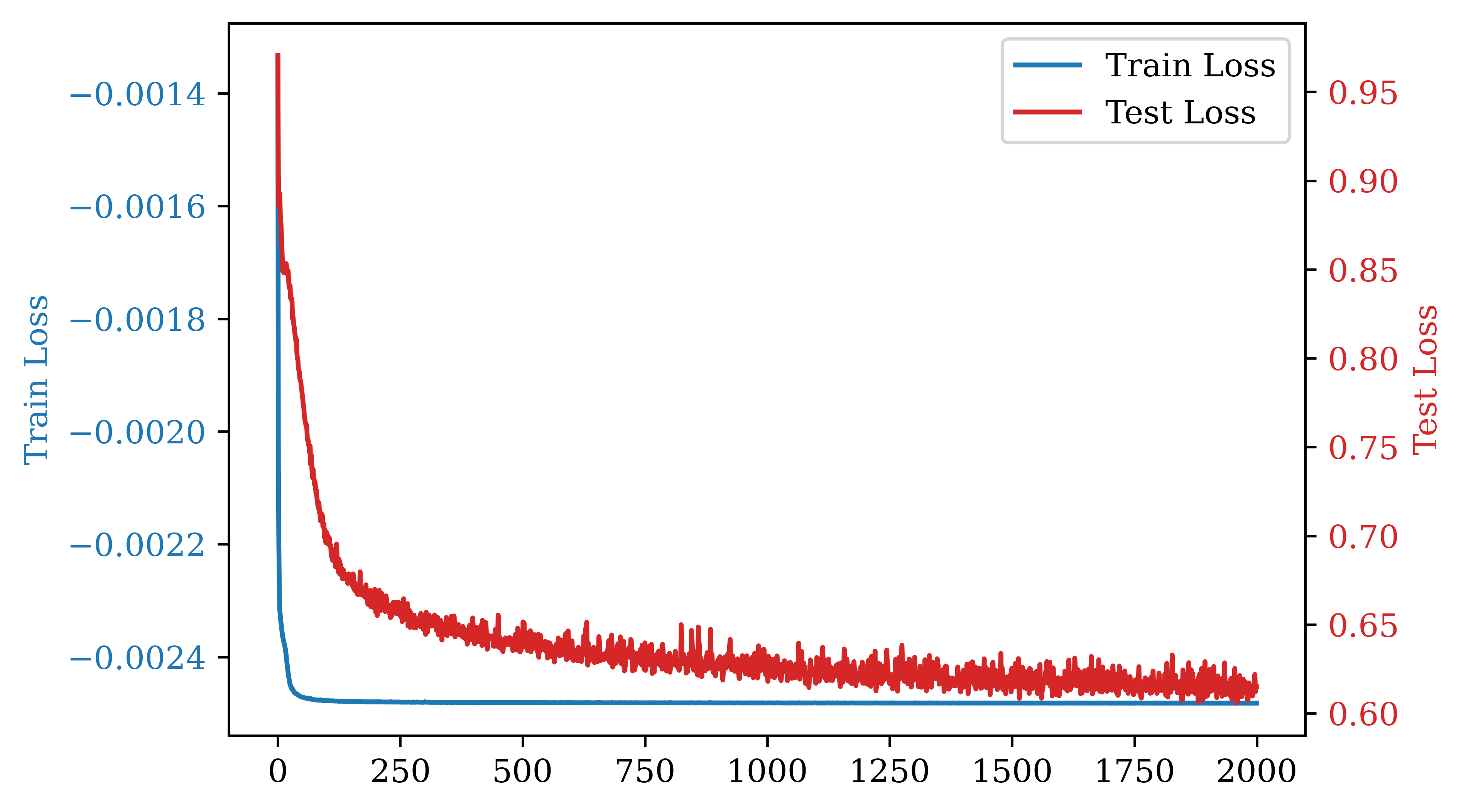}
    \includegraphics[width=0.33\textwidth]{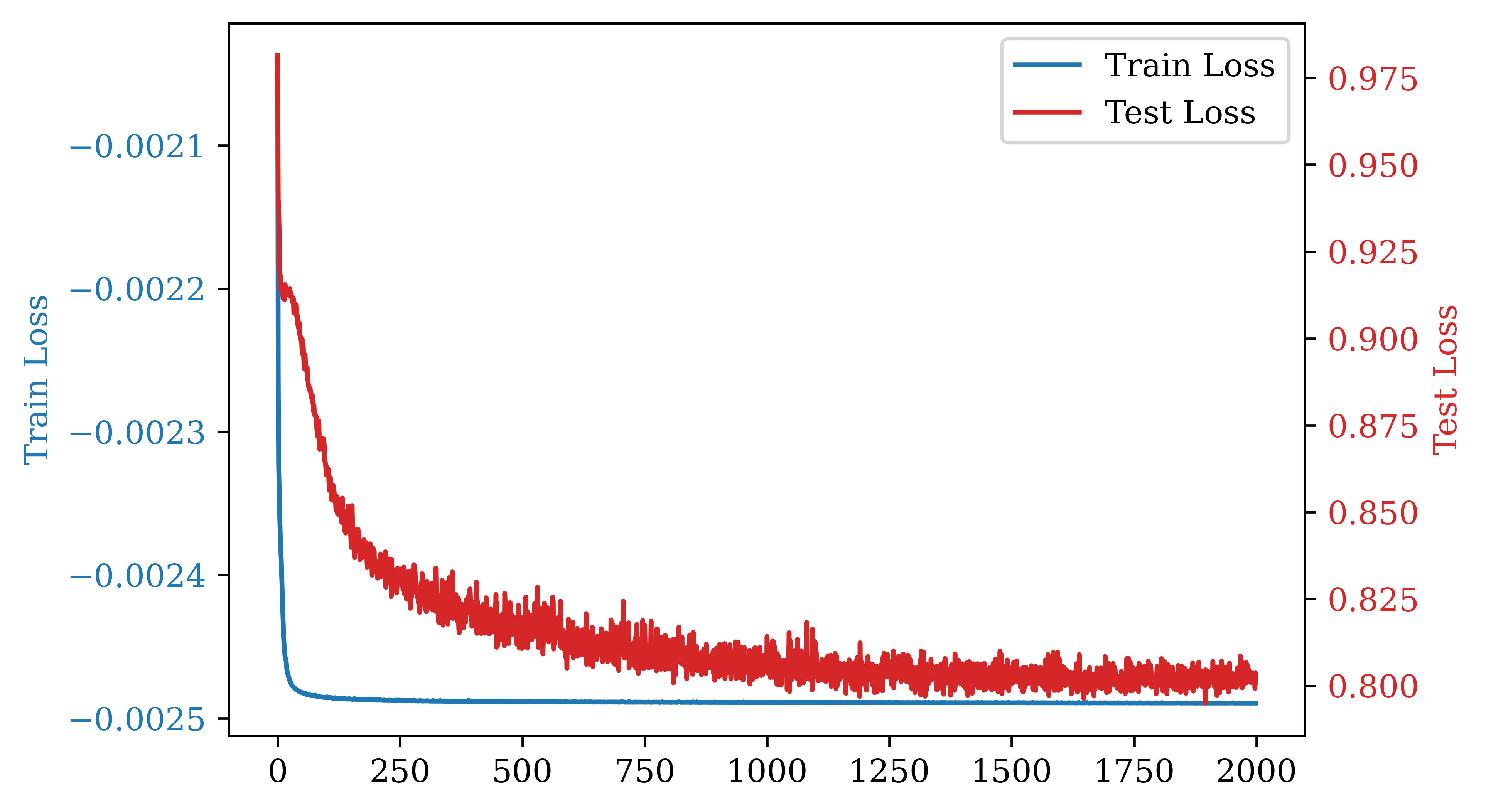}
    \caption{Loss functions over training iterations  for resolutions size 64, 128, and 256, respectively}
    \label{fig:loss_trend}
\end{figure}

\begin{figure}[h!]
    \centering
    \includegraphics[width=0.33\textwidth]{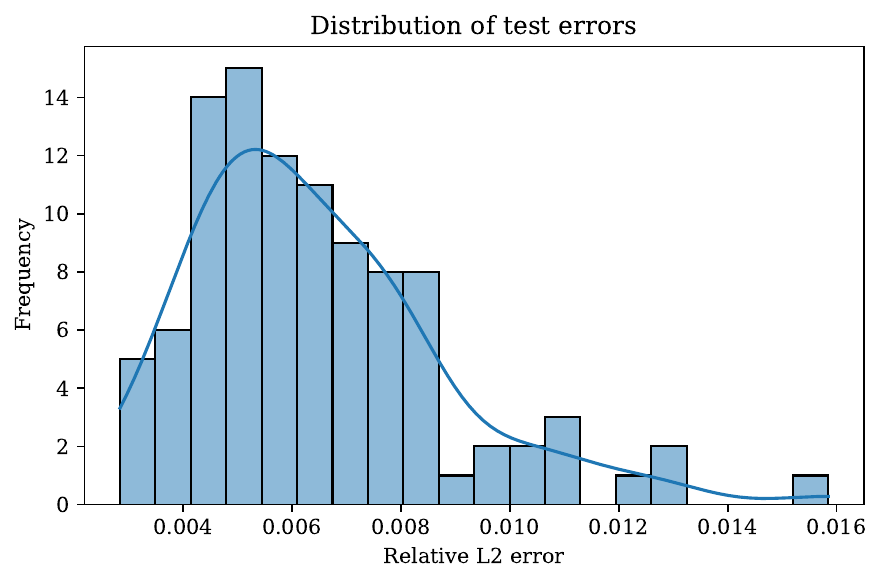}
    \includegraphics[width=0.33\textwidth]{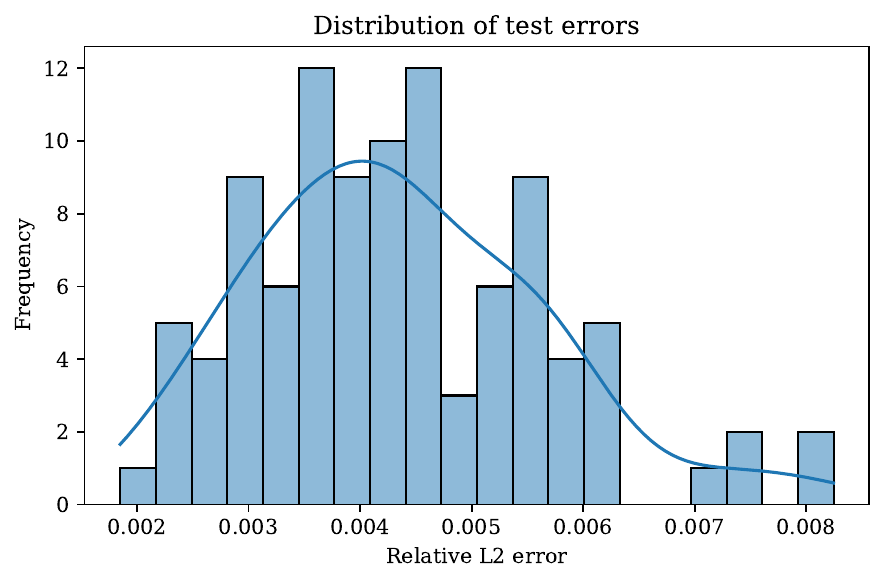}
    \includegraphics[width=0.33\textwidth]{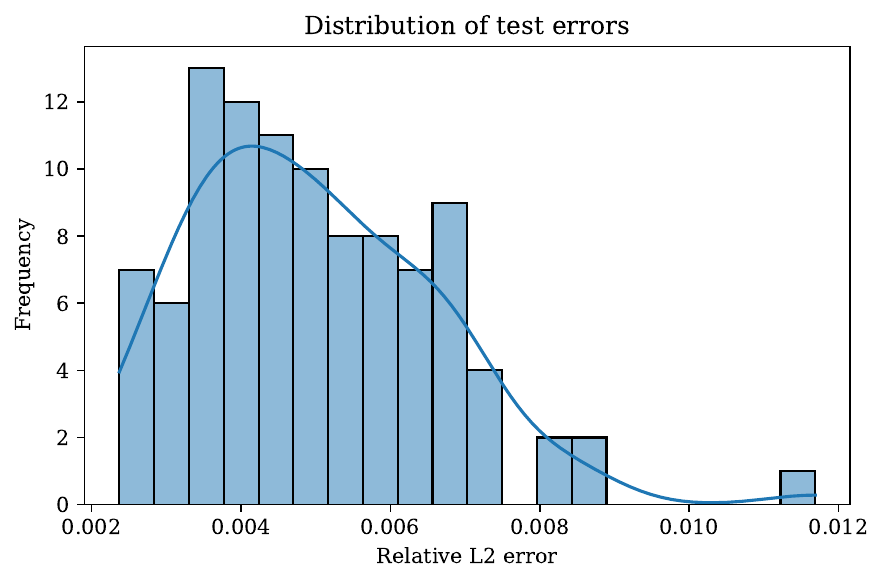}
    \caption{Distribution of per-sample test \(L^2\) errors for the VINO model trained on Darcy flow for resolutions size 64, 128, and 256, respectively.}
    \label{fig:darcy_error_distribution}
\end{figure}

\begin{figure}[h!]
    \centering
    \includegraphics[width=0.3325\textwidth]{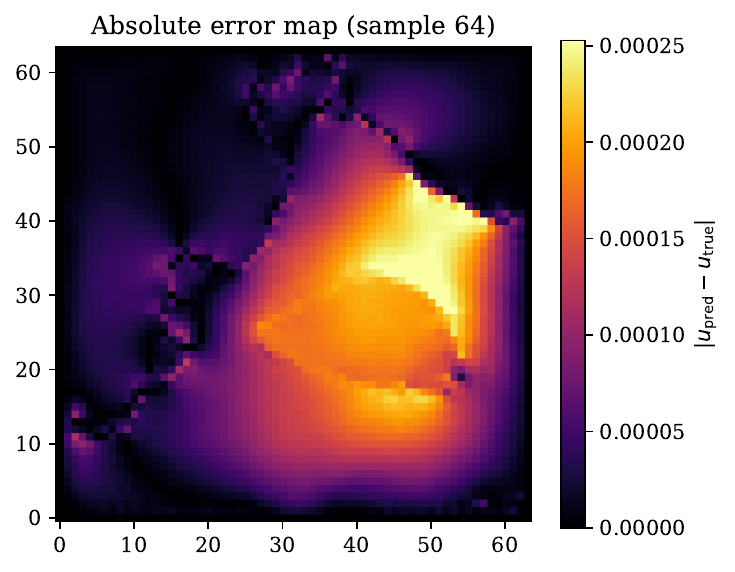}
    \includegraphics[width=0.31\textwidth]{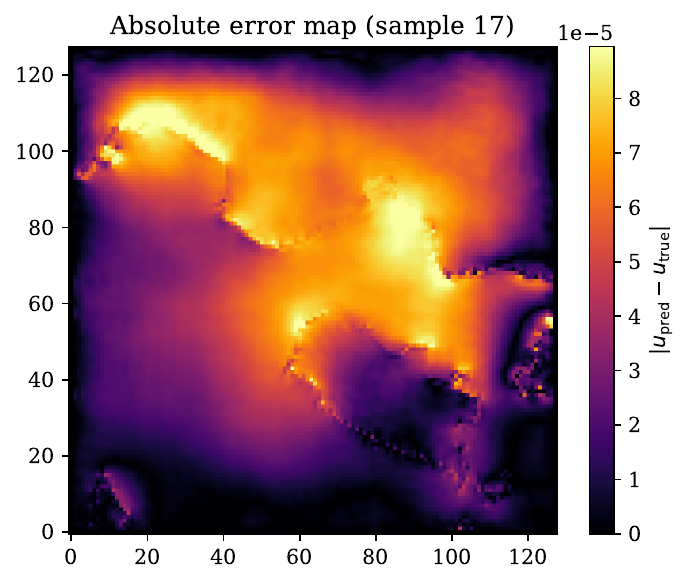}
    \includegraphics[width=0.34\textwidth]{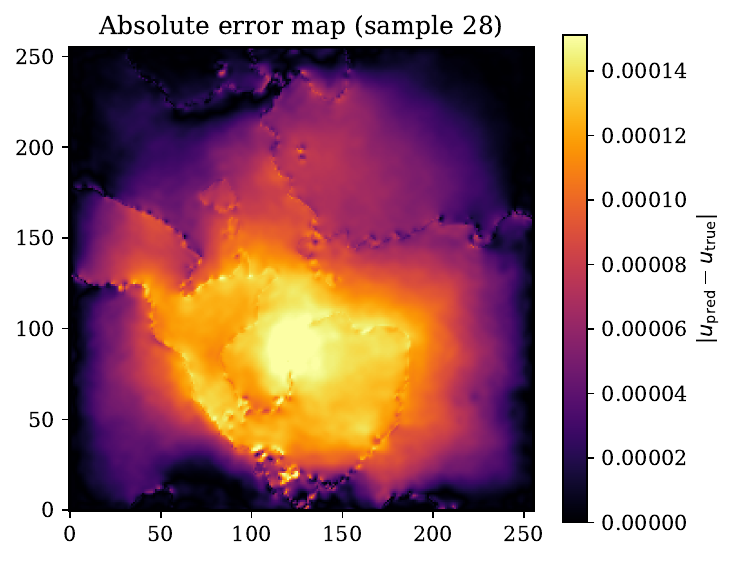}
    \caption{Example absolute error map \(|u_{\text{pred}} - u_{\text{exact}}|\) for a heterogeneous permeability realization, showing localized residuals near strong contrasts.}
    \label{fig:darcy_absolute_error}
\end{figure}

In the resolution \(64\times64 \), the trained model achieved an average test \(L^2\) error of \(6.4\times10^{-3}\) with a standard deviation of \(2.3\times10^{-3}\). The minimum and maximum test errors across all samples were \(2.8\times10^{-3}\) and \(1.5\times10^{-2}\), respectively. The loss functions based on iteration for different resolutions have been illustrated in Fig.~\ref{fig:loss_trend}. The distribution of errors (Fig.~\ref{fig:darcy_error_distribution}) is compact, showing no heavy tails.
Absolute error maps for representative permeability realizations (Fig.~\ref{fig:darcy_absolute_error}) indicate that residuals are localized near sharp permeability contrasts, confirming that the operator captures large-scale pressure response with high fidelity.

\paragraph{Iterative solver acceleration.}
The VINO output was used to initialize several iterative solvers, including the standard CG and preconditioned CG variants.
For the unpreconditioned CG solver, the average runtime decreased \(24\%\).
The relative \(L^2\) error in NOWS was \(1.7\times10^{-6}\), compared to \(8.6\times10^{-7}\) for the converged solver, confirming that accuracy is fully preserved.

\begin{figure}[h!]
    \centering
    \includegraphics[width=0.33\textwidth]{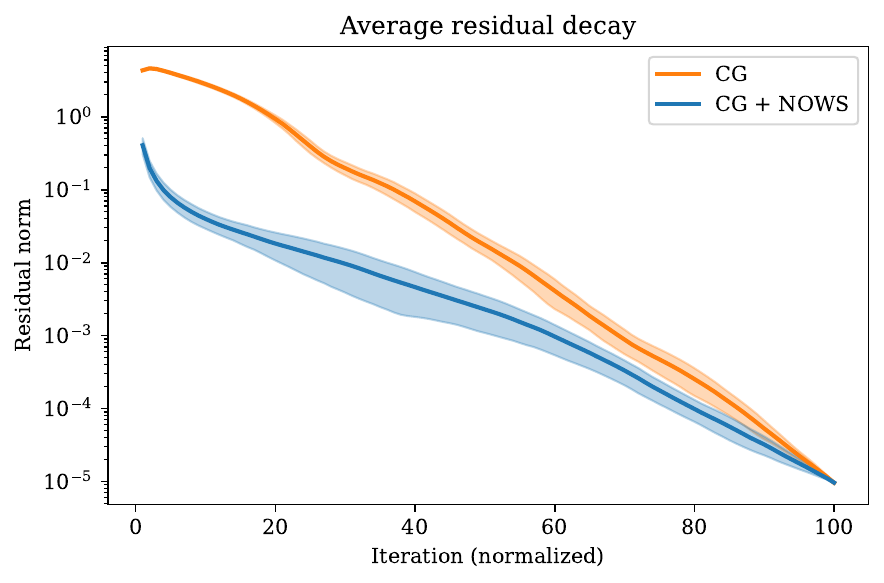}
    \includegraphics[width=0.33\textwidth]{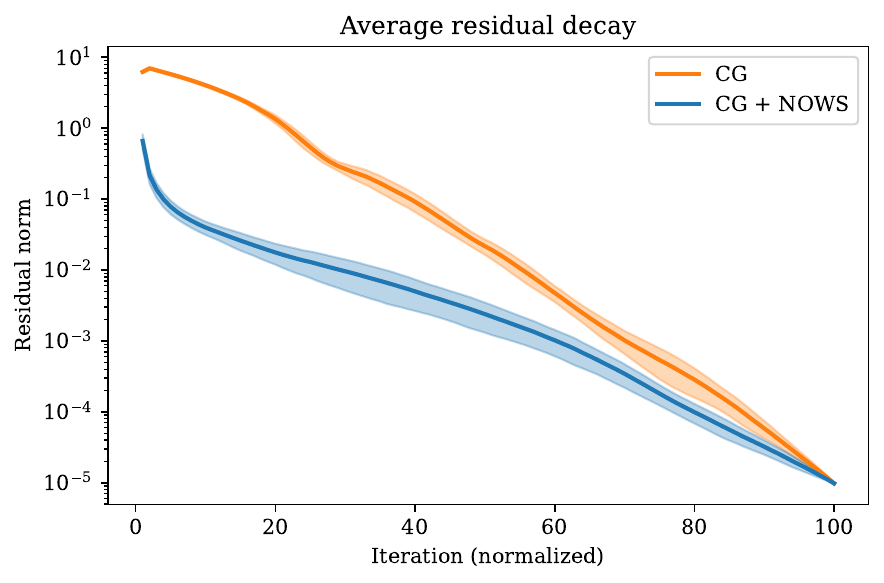}
    \includegraphics[width=0.33\textwidth]{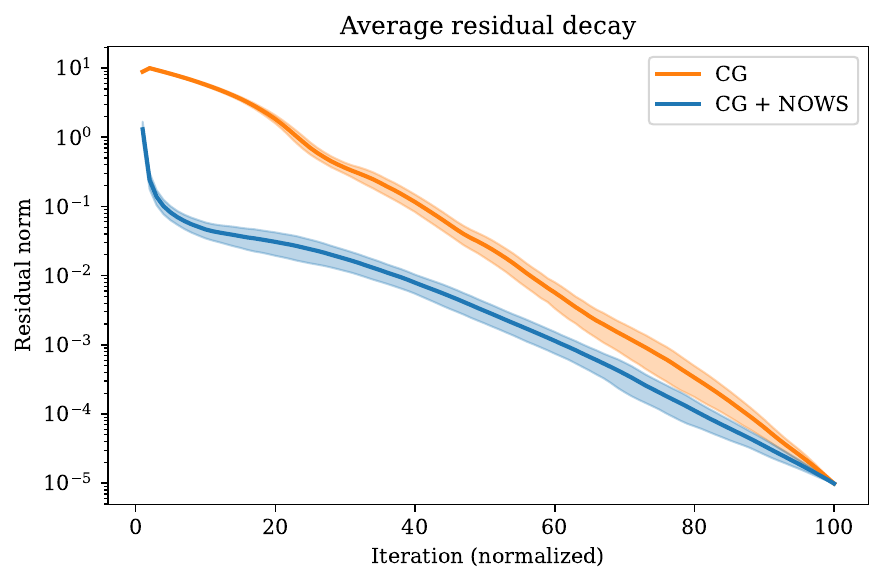}
    \caption{\textbf{Supplementary Figure S4.}
    Mean residual-norm decay of CG and CG+NOWS across all test samples.}
    \label{fig:residual_decay}
\end{figure}

Residual trajectories averaged over the test set (Fig.~\ref{fig:residual_decay}) show that NOWS initialization begins roughly two orders of magnitude closer to the converged solution and consistently shortens the iteration sequence.

\paragraph{Preconditioning analysis.}
We further examined the interplay between NOWS initialization and preconditioning strategies.
Five common preconditioners were considered: Jacobi, SSOR, ICC (Incomplete Cholesky), ILU (Incomplete LU), and the unpreconditioned baseline.
Each preconditioner modifies the spectral conditioning of the FEM stiffness matrix, impacting CG convergence.
For each preconditioner, both baseline and NOWS-initialized solvers were evaluated over 100 random permeability fields.

\begin{figure}[h!]
    \centering
    \includegraphics[width=0.33\textwidth]{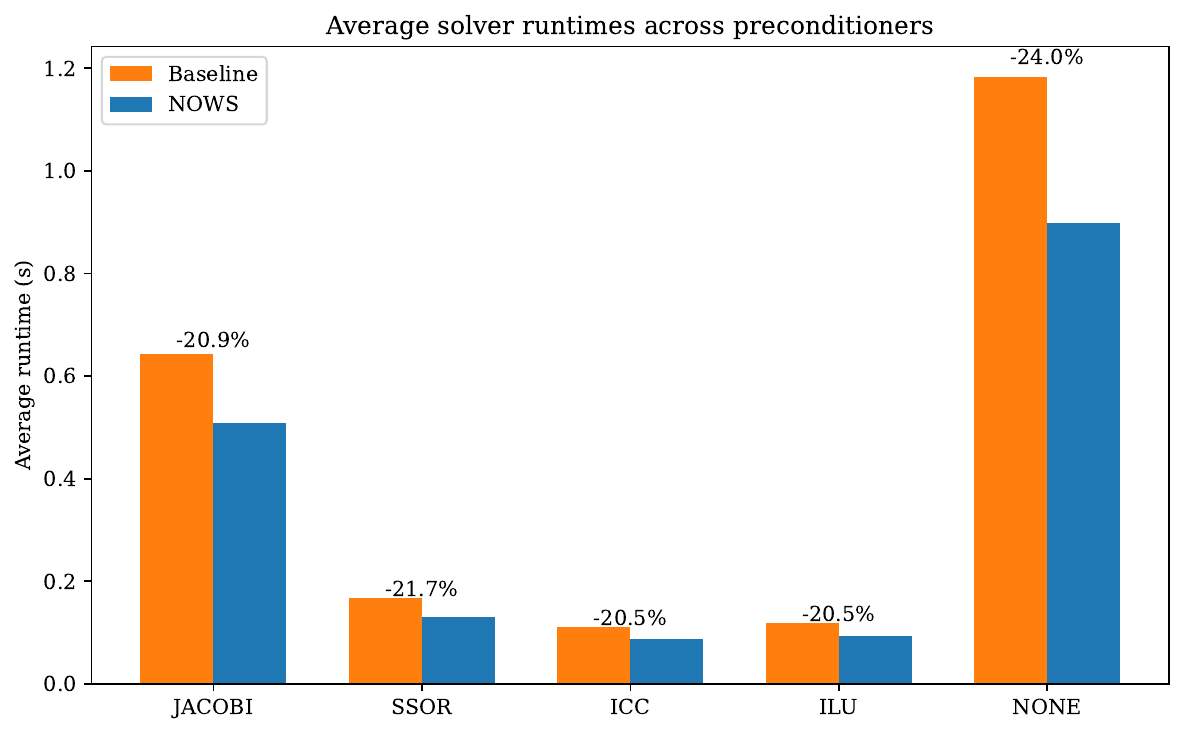}
    \includegraphics[width=0.33\textwidth]{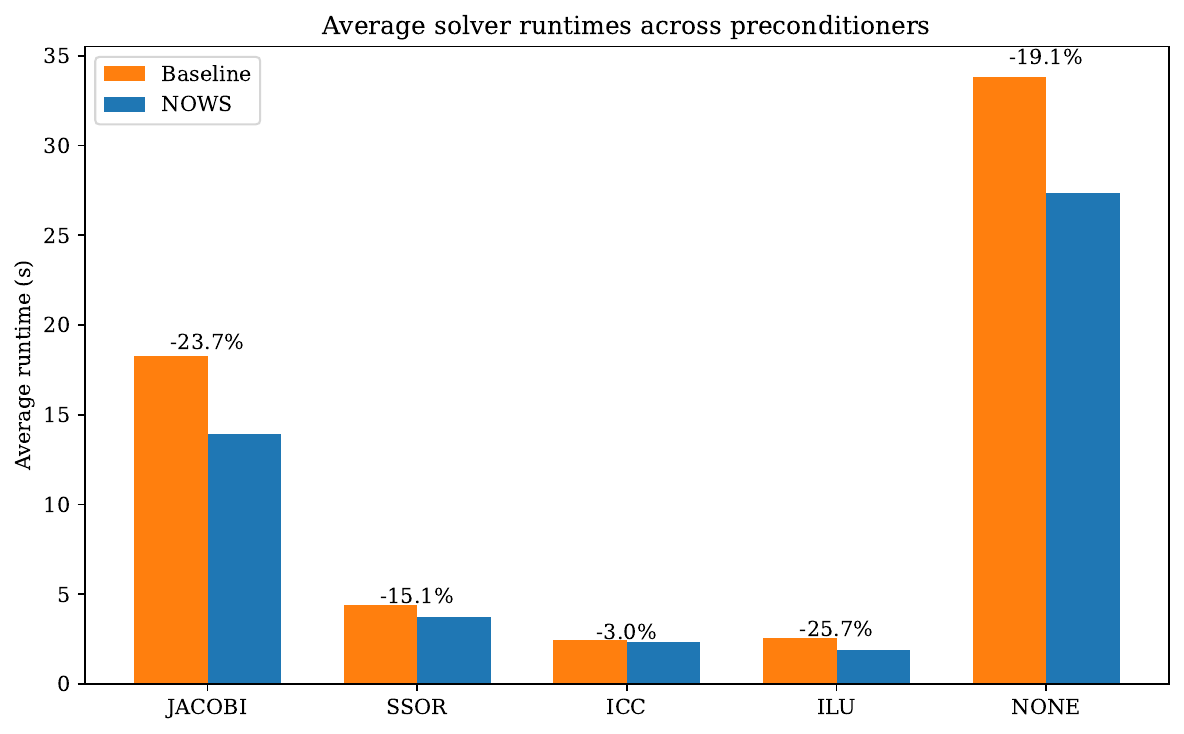}
    \includegraphics[width=0.33\textwidth]{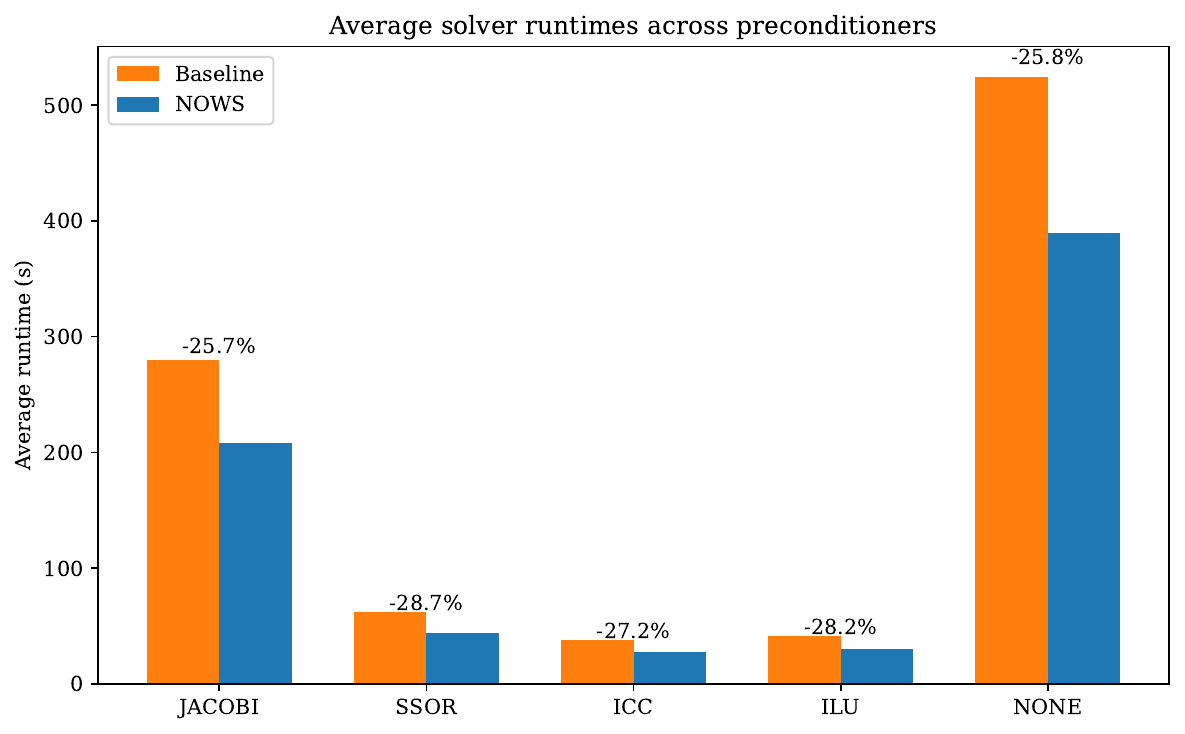}
    \caption{Comparison of average solver runtimes across preconditioners (Jacobi, SSOR, ICC, ILU, and none) with and without NOWS initialization.}
    \label{fig:runtime_bar}

\end{figure}

\begin{figure}[h!]
    \centering
    \includegraphics[width=0.33\textwidth]{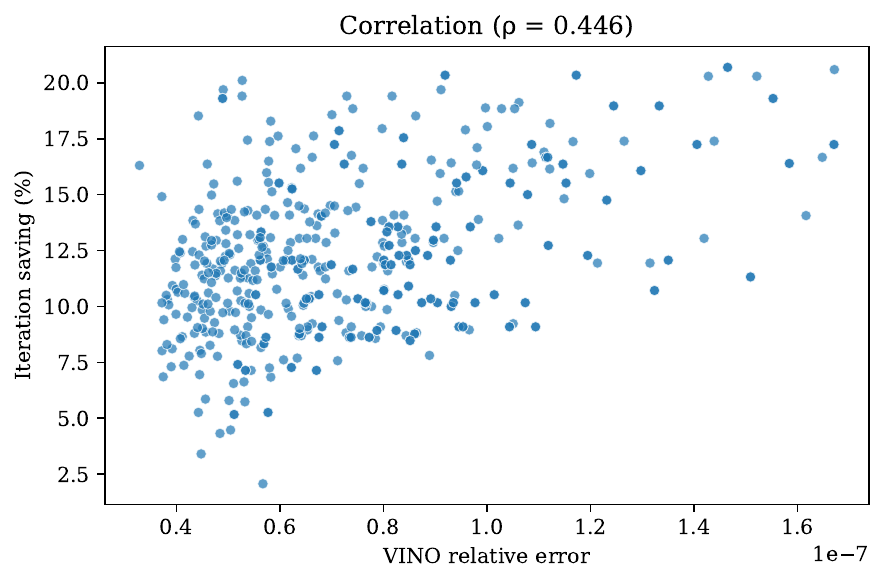}
    \includegraphics[width=0.33\textwidth]{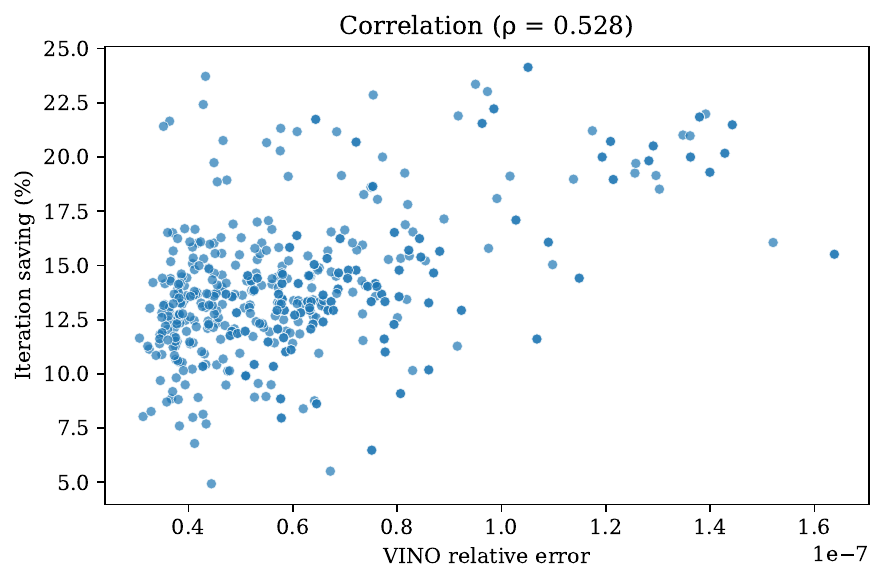}
    \includegraphics[width=0.33\textwidth]{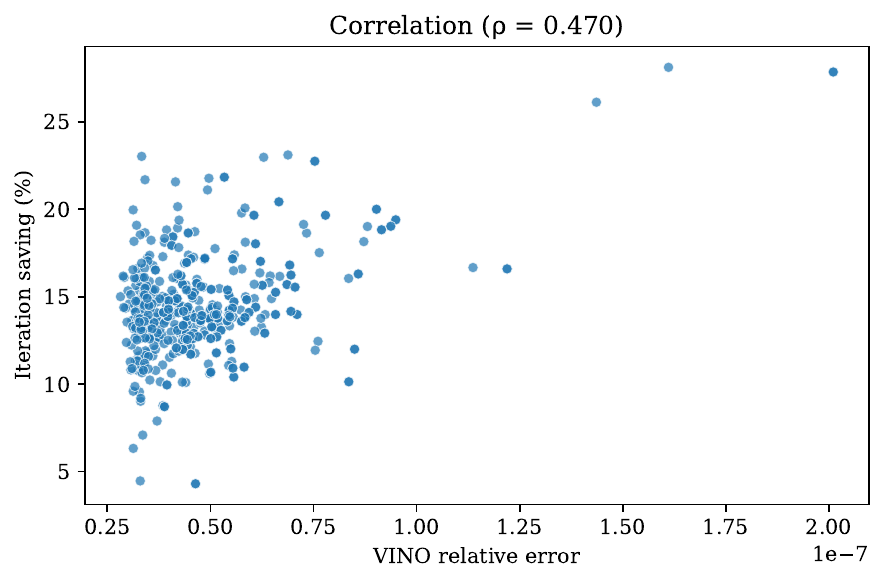}
    \caption{Scatter plot of iteration savings versus warm-start relative error, showing weak negative correlation and consistent performance improvements.}
    \label{fig:speedup_vs_error}

\end{figure}

Across all preconditioners, NOWS consistently reduced solver runtime (Fig.~\ref{fig:runtime_bar}).
Average time savings are about 25 percent.
The improvement was nearly linear in the logarithm of the warm-start accuracy, as shown in the correlation plot between VINO prediction error and iteration savings (Fig.~\ref{fig:speedup_vs_error}).

\subsection*{Plate with Arbitrary Voids - Isogeometric Analysis}

To further examine the performance and robustness of the NOWS framework on irregular domains, we perform an extended statistical and physical analysis of the plate-with-voids benchmark. The goal is to quantify runtime savings, iteration reduction, and accuracy trends across diverse geometric configurations, and to assess how these improvements relate to structural complexity and stress distribution consistency.

\begin{figure}[ht!]
    \centering
    \includegraphics[width=0.49\textwidth]{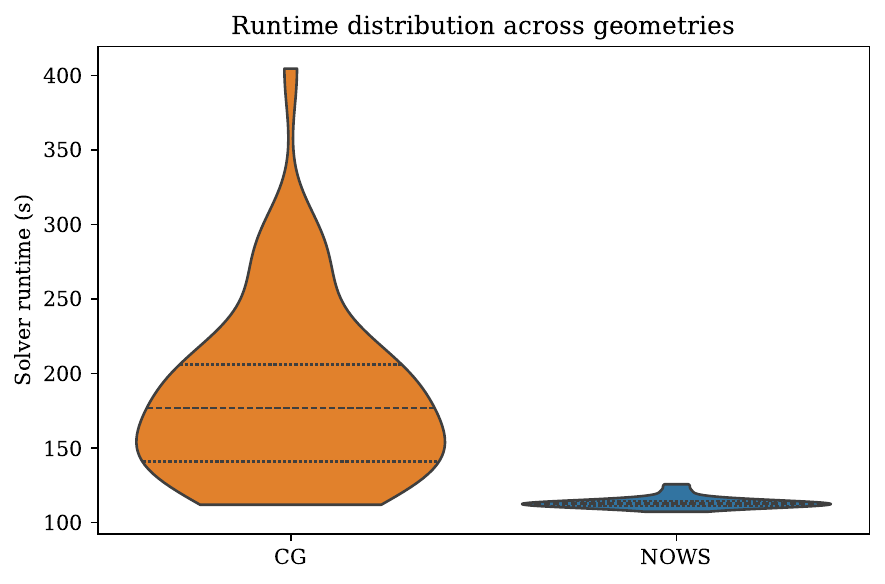}
    \includegraphics[width=0.49\textwidth]{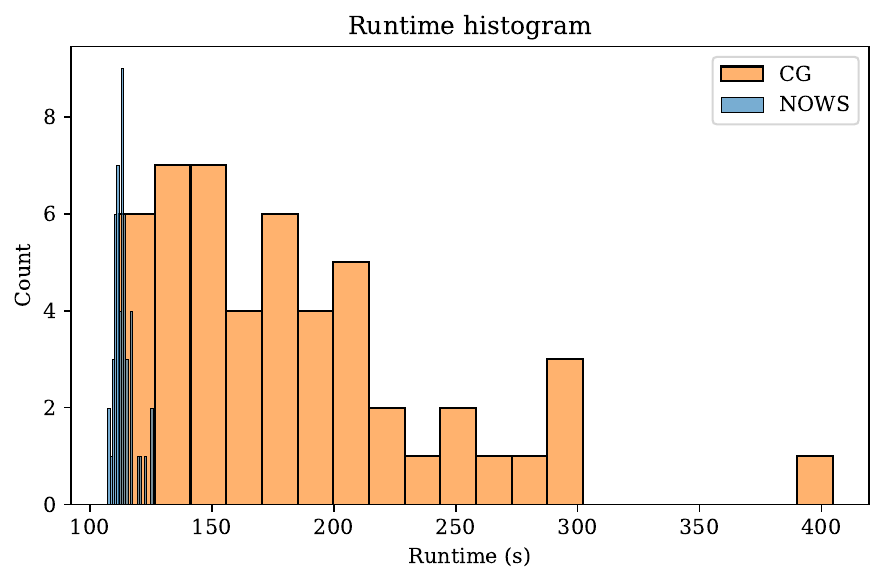}
    \caption{ 
    Violin plot and bar plot of solver runtime distributions for 50 test geometries, showing nearly order-of-magnitude speed-up.}
    \label{fig:plate_supp_1}
\end{figure}

\begin{figure}[ht!]
    \centering
    \includegraphics[width=0.49\textwidth]{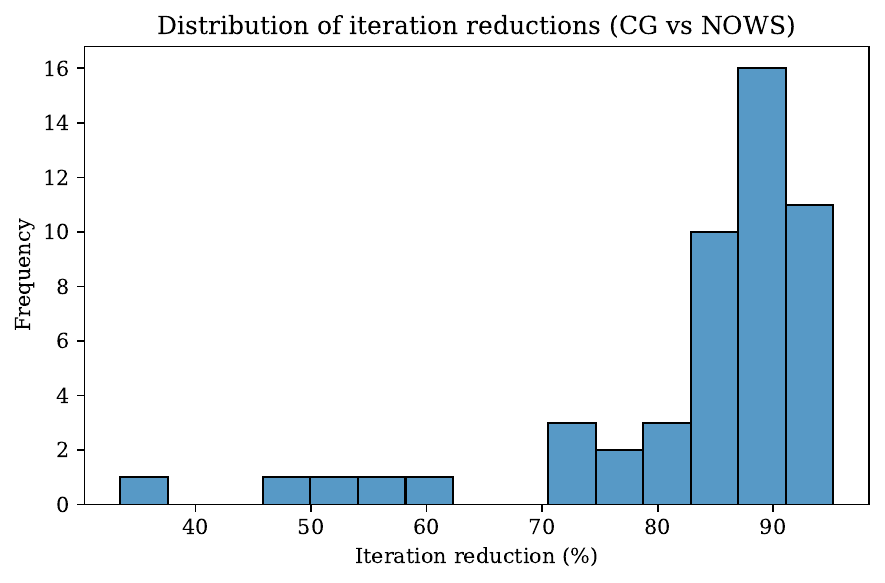}
    \includegraphics[width=0.49\textwidth]{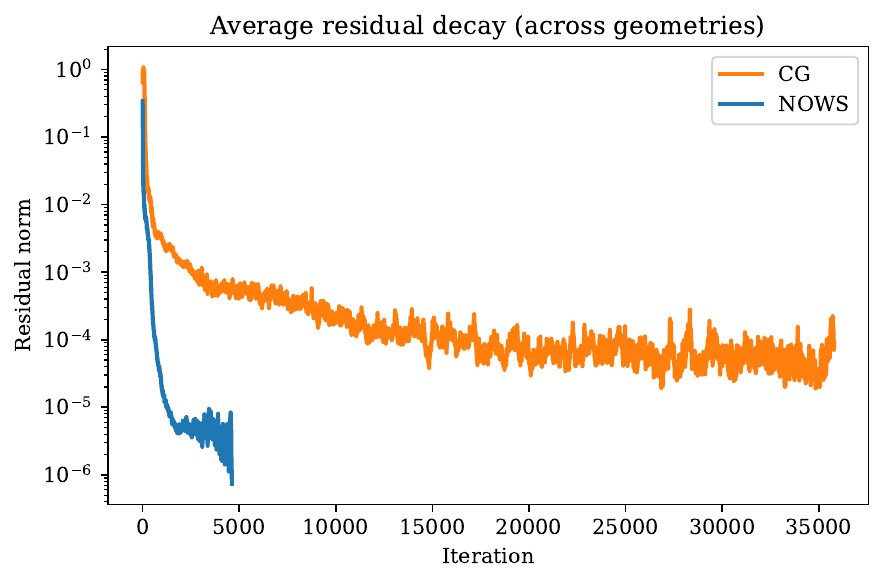}
    \caption{  
    \textbf{(a)} Histogram of iteration reductions, with a mean decrease of $\sim$91\%. 
    \textbf{(b)} Ensemble-averaged residual decay curves for CG and CG+NOWS, illustrating reduced initial residuals and faster convergence.}
    \label{fig:plate_supp_2}
\end{figure}

\paragraph{Statistical runtime and iteration analysis.}
Across the ensemble of 50 randomly generated test geometries, NOWS reduces the mean wall-clock runtime by approximately \textbf{88.3\%} relative to the baseline CG solver (Fig.~\ref{fig:plate_supp_1}). A similar pattern emerges in the iteration statistics, where NOWS initialization lowers the number of CG iterations required for convergence by an average of \textbf{91.4\%}, with a narrow variance across test instances (Fig.~\ref{fig:plate_supp_2}a). The distributions of both metrics remain unimodal and compact, indicating stable acceleration even for highly irregular geometries.

\paragraph{Residual and convergence characteristics.}
Figure~\ref{fig:plate_supp_2}b presents the ensemble-averaged residual decay curves for both solvers. The NOWS initialization substantially decreases the initial residual magnitude, yielding a consistently lower convergence trajectory and reaching the prescribed tolerance with fewer than 10\% of the baseline iterations. This convergence acceleration is achieved without modifying the solver structure or preconditioning scheme, reaffirming the solver-agnostic nature of the NOWS framework.

\begin{figure}[ht!]
    \centering
    \includegraphics[width=0.49\textwidth]{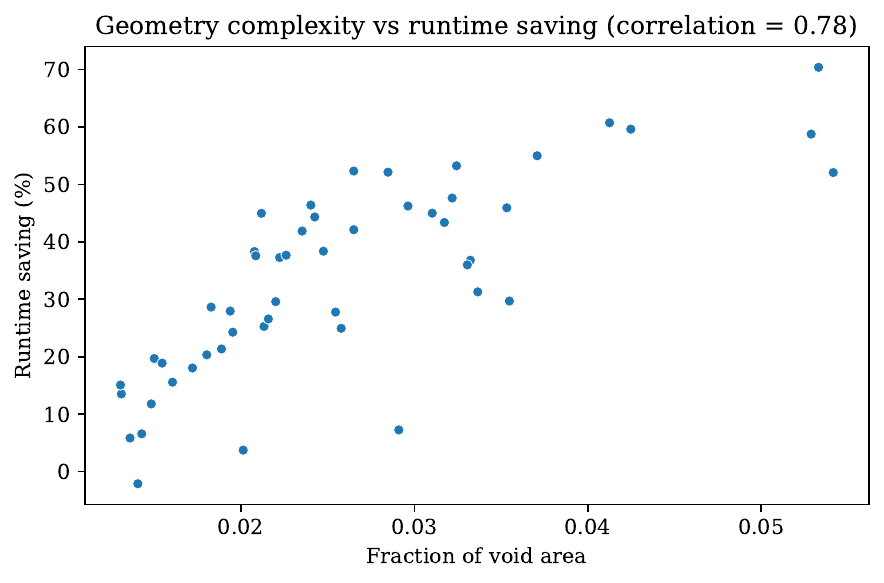}
    \includegraphics[width=0.49\textwidth]{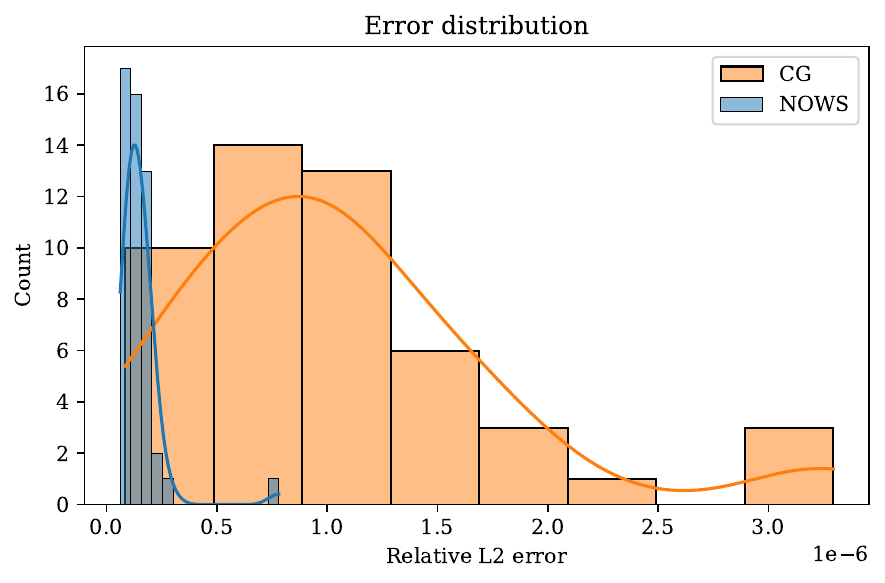}
    \caption{
    \textbf{(a)} Correlation between void fraction (geometry complexity) and runtime savings, showing weak dependence ($\rho \approx 0.18$). 
    \textbf{(b)} Distribution of relative $L^2$ errors for displacement fields, confirming identical solution accuracy.}
    \label{fig:plate_supp_3}
\end{figure}

\paragraph{Geometry complexity versus solver speed-up.}
To examine the relationship between geometric irregularity and solver acceleration, we define a \emph{complexity measure} as the fraction of void area within the plate domain. As shown in Fig.~\ref{fig:plate_supp_3}a, the runtime savings exhibit a dependence on geometric complexity (Pearson correlation coefficient $\rho = 0.78$), indicating that NOWS provides more benefits for more complex geometries.

\paragraph{Accuracy and error distribution.}
The relative $L^2$ error distributions for displacement predictions show that the NOWS-enhanced solver maintains the same (or better) level of accuracy as the baseline CG solver, while converging significantly faster (Fig.~\ref{fig:plate_supp_3}e). The mean relative error remains below $10^{-3}$ in all cases. Moreover, spatial error maps reveal that discrepancies are localized near material interfaces and void boundaries, regions of high strain gradient, while the bulk solution remains indistinguishable between methods.

\begin{figure}[ht!]
    \centering
    \includegraphics[width=0.33\textwidth]{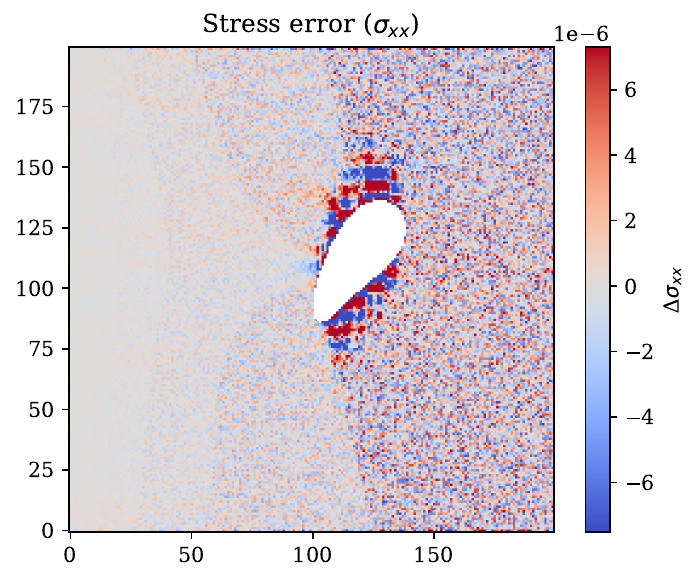}
    \includegraphics[width=0.33\textwidth]{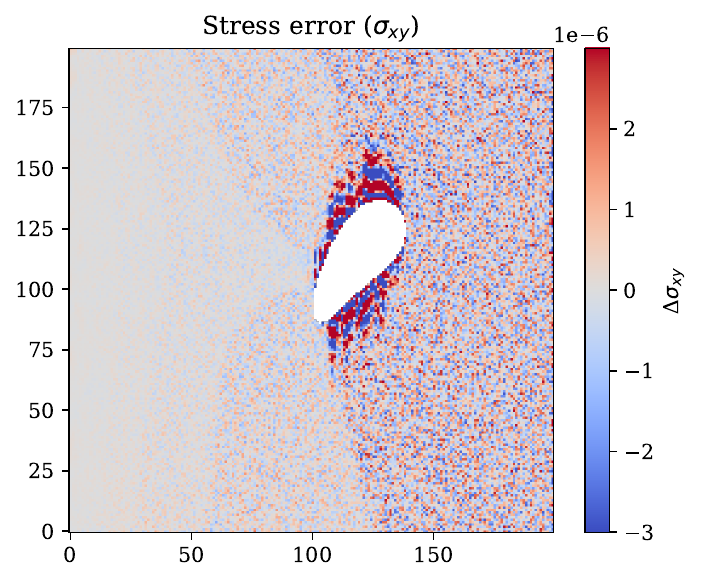}
    \includegraphics[width=0.33\textwidth]{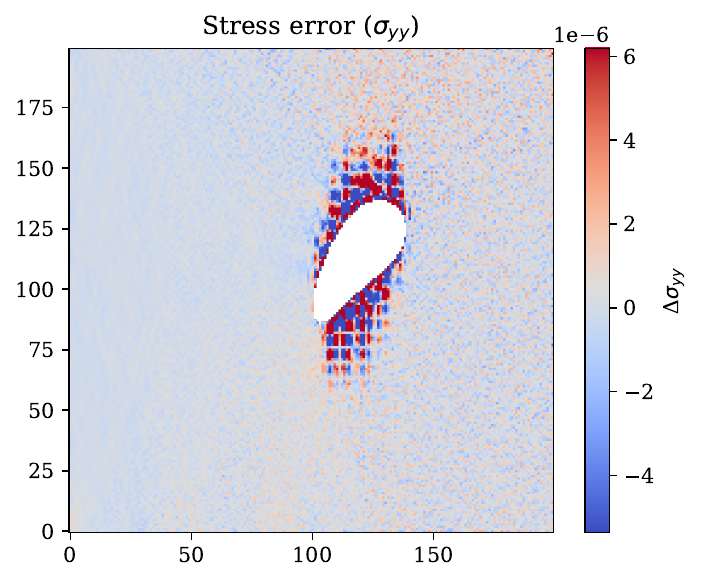}
    \caption{ 
    Stress field difference maps for $\sigma_{xx}$, $\sigma_{yy}$, and $\sigma_{xy}$ components, demonstrating negligible deviation from the reference solution.}
    \label{fig:plate_supp_4}
\end{figure}

\paragraph{Stress-based validation.}
To assess physical consistency, stress components $(\sigma_{xx}, \sigma_{yy}, \sigma_{xy})$ were computed from the displacement fields obtained via both solvers. The stress differences between CG and NOWS solutions remain below $0.001\%$ of the reference magnitude (Fig.~\ref{fig:plate_supp_4}). The high correspondence across all components confirms that the neural warm-start initialization preserves equilibrium and compatibility in the reconstructed fields, validating the physical fidelity of the accelerated solution.

These extended results reinforce the conclusions presented in the main text: NOWS reliably accelerates iterative solvers for elasticity problems on complex, irregular domains without compromising accuracy or physical consistency. The method exhibits strong generalization to varying void geometries, maintains solver stability, and achieves substantial reductions in both computational cost and iteration count, demonstrating its practical utility in large-scale IGA-based structural simulations.

\subsection*{Burgers’ Equation: Additional Analyses}

To further assess the behavior of NOWS on time-dependent problems, we conduct extended evaluations on the one-dimensional viscous Burgers’ equation described in the main text. The Multi-Head Neural operator is trained on $1{,}000$ training and $100$ test samples, each representing an independent realization of the initial field $u_0(x)$. The trained neural operator provides the warm-start initialization for the CG solver used within a finite-difference scheme.

\begin{figure}[H]
    \centering
    \includegraphics[width=0.335\textwidth]{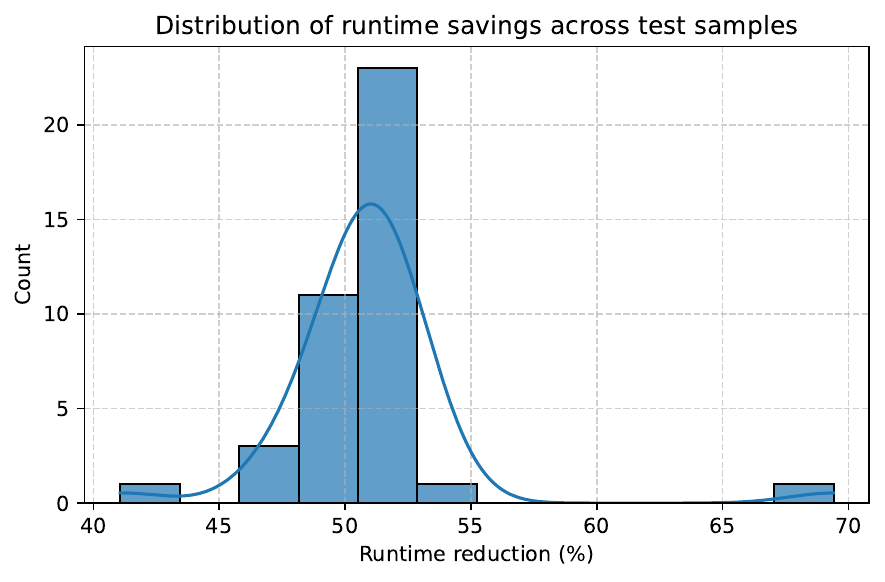}
    \includegraphics[width=0.295\textwidth]{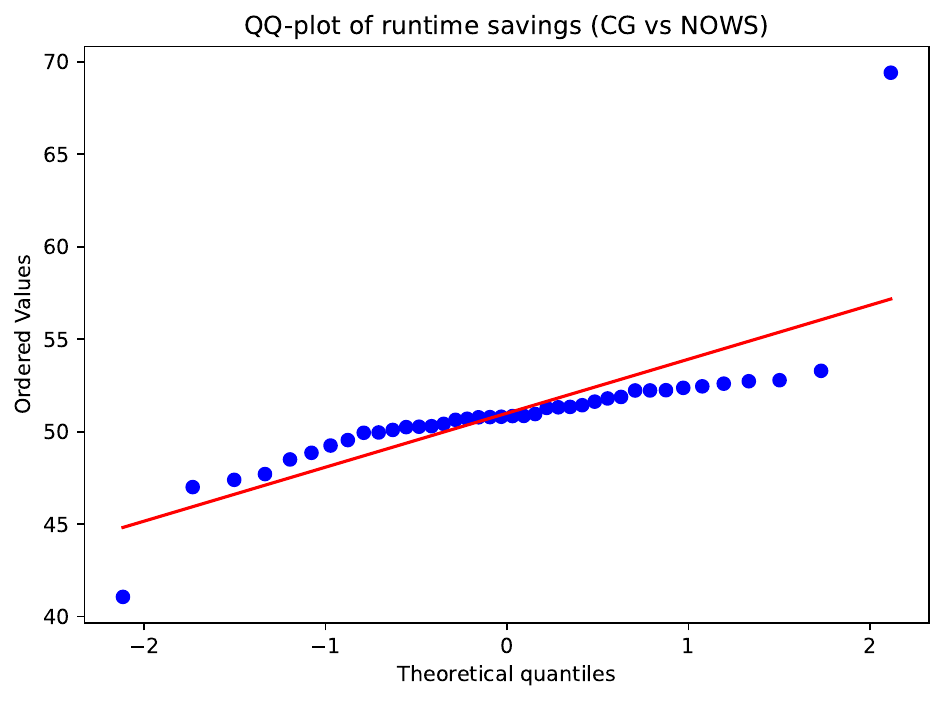}
    \includegraphics[width=0.36\textwidth]{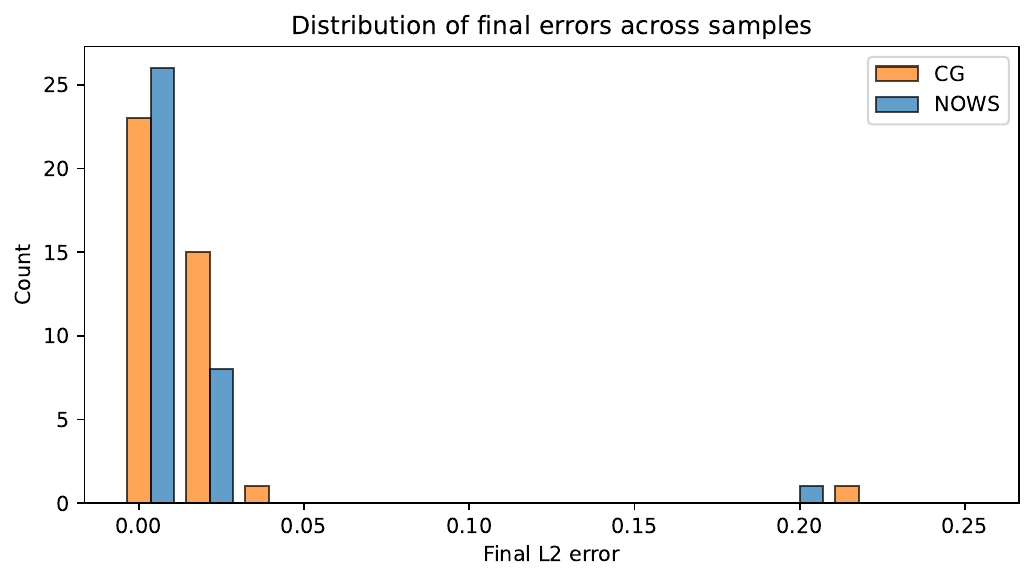}
    \caption{\textbf{a. Runtime savings distribution.}
    Histogram of relative runtime reductions (in percent) achieved by NOWS across all test samples.
    \textbf{b. Normality of runtime reduction.}
    Quantile–quantile plot comparing observed runtime savings with a normal distribution.
    \textbf{Distribution of final $L_2$ errors across test samples.}
    Comparison between the NOWS-accelerated solver and the baseline CG method. 
    Both distributions are concentrated near zero, showing that the warm-start initialization preserves final accuracy while reducing runtime.}
    \label{fig:burgers_supp}
\end{figure}

\paragraph{Runtime statistics.}
The runtime distributions in Figure~S\ref{fig:burgers_supp}a show a clear leftward shift for NOWS-initialized runs, indicating uniformly shorter wall-clock times. The mean time saving is $50.5\% \pm 3.4\%$, in agreement with the main text. The QQ-plot (Figure~S\ref{fig:burgers_supp}b) indicates that the distribution of runtime savings across test samples is approximately normal, with most values centered around 50\%. A few outliers at the lower and upper tails correspond to specific realizations of the initial condition, reflecting minor variability in solver convergence behavior.

\paragraph{Error consistency across samples.}
To evaluate the consistency of solution accuracy, we compared the final $L_2$ errors across all test samples for the CG and NOWS solvers. 
Figure~S\ref{fig:burgers_supp} shows the histogram of the final errors. 
Both methods achieve very small absolute errors, concentrated near zero, indicating high numerical fidelity. The distributions nearly overlap, confirming that the neural initialization does not compromise accuracy.

\subsection*{Smoke Plume – Resolution Invariance}

To further assess the scalability and generalization of NOWS, we consider a coupled, nonlinear flow problem governed by the incompressible Navier–Stokes equations with buoyancy-driven smoke transport. The system represents a typical multiphysics problem solved using the finite-volume method. The governing equations are
\begin{align}
\nabla \cdot \mathbf{u} &= 0, \\
\frac{\partial \mathbf{u}}{\partial t} + (\mathbf{u} \cdot \nabla)\mathbf{u} &= 
-\nabla p + \nu \nabla^2 \mathbf{u} + \mathbf{g}\,\beta (\rho - \rho_0), \\
\frac{\partial \rho}{\partial t} + (\mathbf{u} \cdot \nabla)\rho &= D \nabla^2 \rho,
\end{align}
where $\mathbf{u}$ denotes velocity, $p$ pressure, $\rho$ smoke density, $\nu$ kinematic viscosity, $\beta$ buoyancy coefficient, and $D$ diffusion coefficient. The pressure projection step is solved iteratively using a CG solver.

In this setup, NOWS provides physics-consistent warm-start initializations for the CG pressure projection, significantly reducing the number of iterations required at each timestep. The neural operator predicts the pressure field at the current step from the preceding field and velocity, supplying a high-quality initial guess that generalizes across flow states and resolutions. 

\begin{figure}[H]
    \centering
    \includegraphics[width=0.48\textwidth]{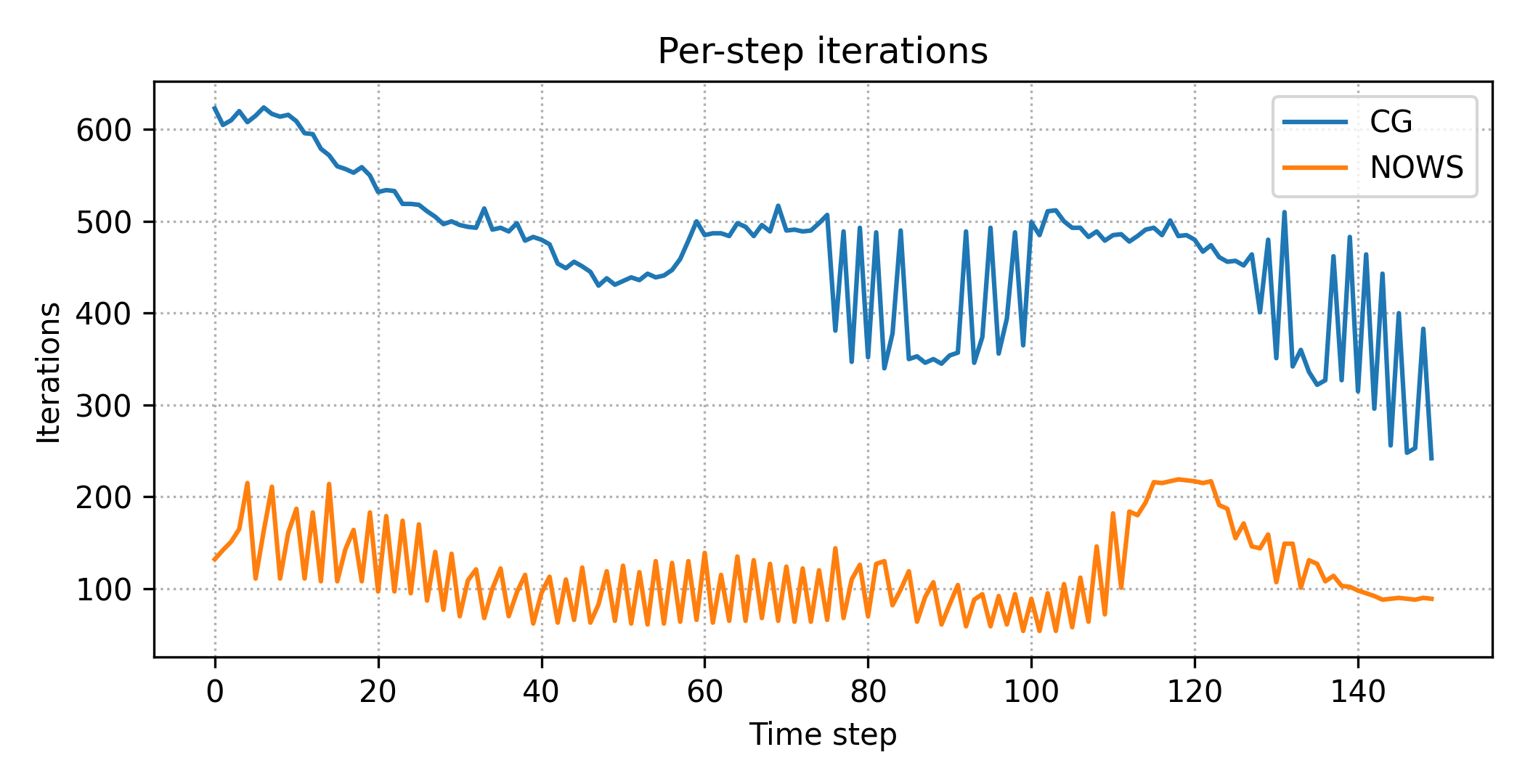}
    \includegraphics[width=0.48\textwidth]{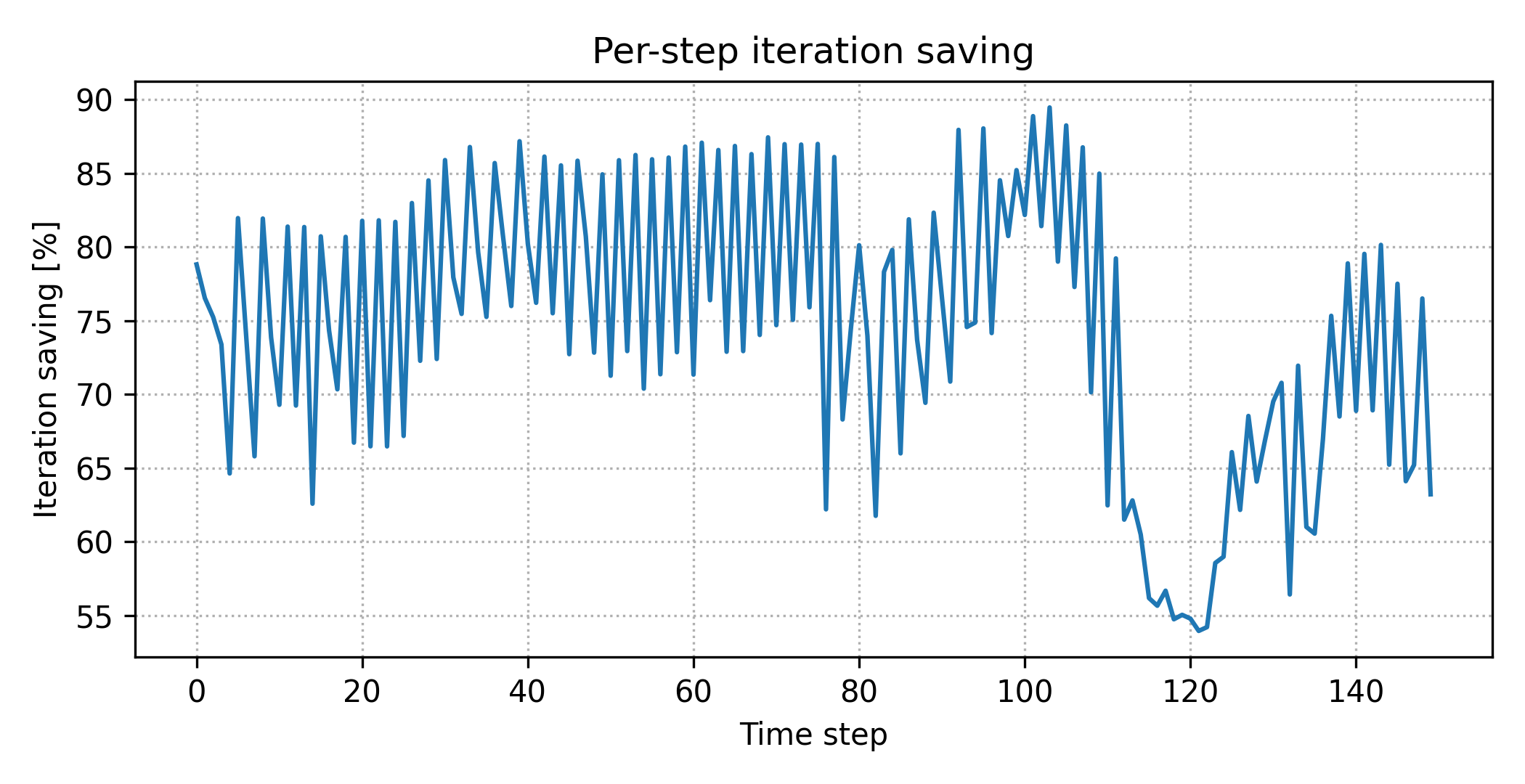}\\[4pt]
    \includegraphics[width=0.48\textwidth]{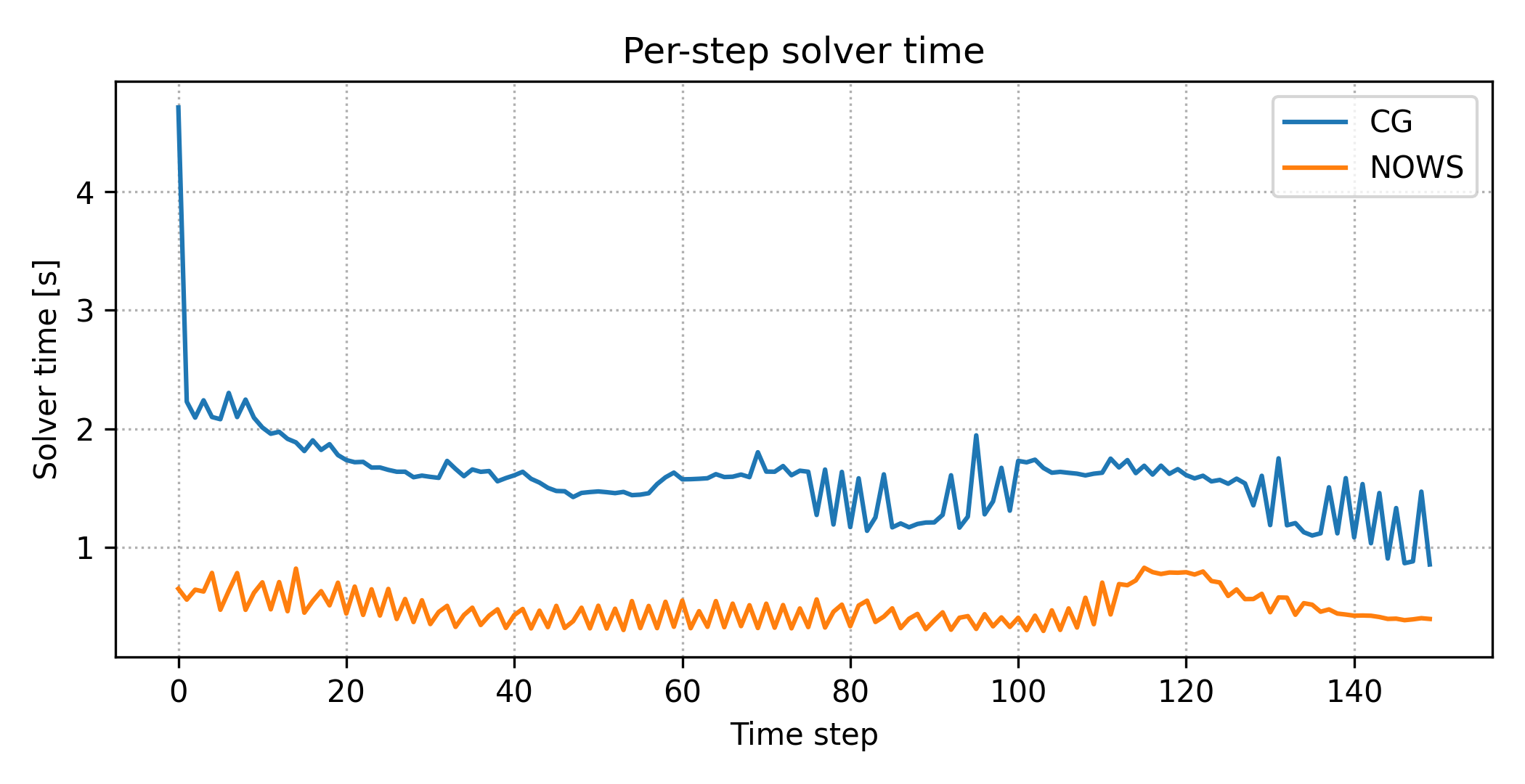}
    \includegraphics[width=0.48\textwidth]{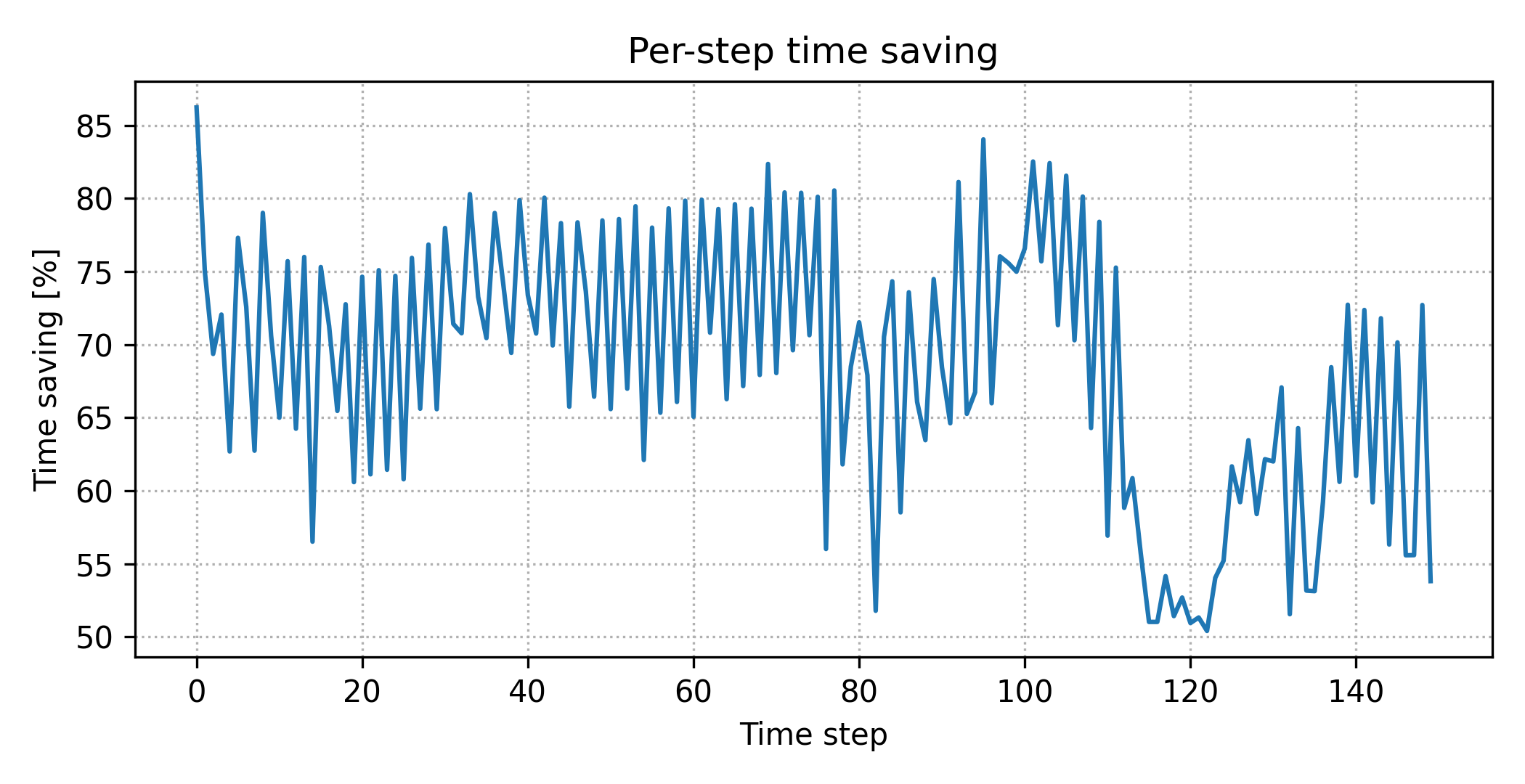}
    \caption{\textbf{Per-step convergence and efficiency trends for the buoyant smoke simulation.}
    (\textbf{a}) Iteration count per time step comparing the baseline CG solver and NOWS-initialized runs. 
    (\textbf{b}) Percentage iteration savings achieved by NOWS, consistently maintaining 70–85\% reduction.
    (\textbf{c}) Solver wall-clock time per step, showing stable acceleration across the entire trajectory.
    (\textbf{d}) Corresponding time-saving percentage, reflecting direct correlation with iteration savings.
    The consistent gap between the curves highlights the robustness and persistence of the neural warm-start benefit throughout the nonlinear transient evolution.}
    \label{fig:smoke_evolution}
\end{figure}

\paragraph{Performance and Convergence Trends.}  
Figure~\ref{fig:smoke_evolution} presents per-step convergence metrics comparing the baseline CG solver with NOWS initialization. The first panel shows the number of iterations required by each method over 150 simulation steps. While the CG solver requires between 300–600 iterations per step, NOWS consistently converges within 80–200 iterations. This reduction is uniform across the entire simulation, demonstrating stable performance even during the transient plume evolution phase. The second panel quantifies the iteration savings, revealing that NOWS achieves between 70\% and 85\% reduction per step, with minor fluctuations linked to dynamic flow features.

The third panel depicts the solver wall-clock time evolution. Consistent with iteration trends, NOWS lowers the per-step solver time from 1–2.5~s to below 0.8~s for most of the simulation. The final panel shows the corresponding time-saving percentage, closely tracking iteration savings and confirming that the efficiency gain stems directly from reduced iteration counts and faster residual convergence. The temporal smoothness and stability of these savings confirm that the neural initialization remains effective even in the presence of evolving nonlinear dynamics.

\begin{figure}[H]
    \centering
    \includegraphics[width=0.49\textwidth]{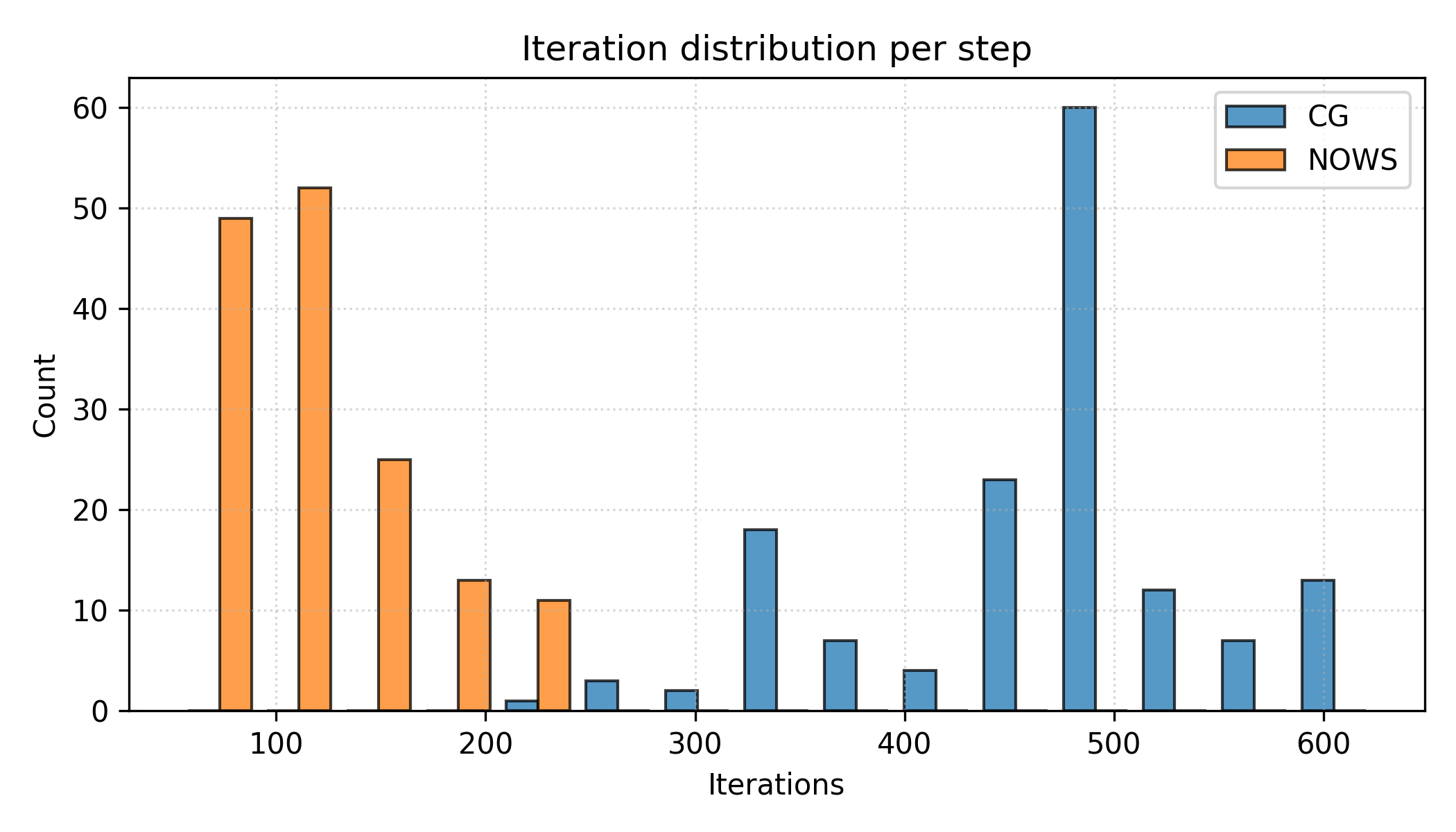}
    \includegraphics[width=0.49\textwidth]{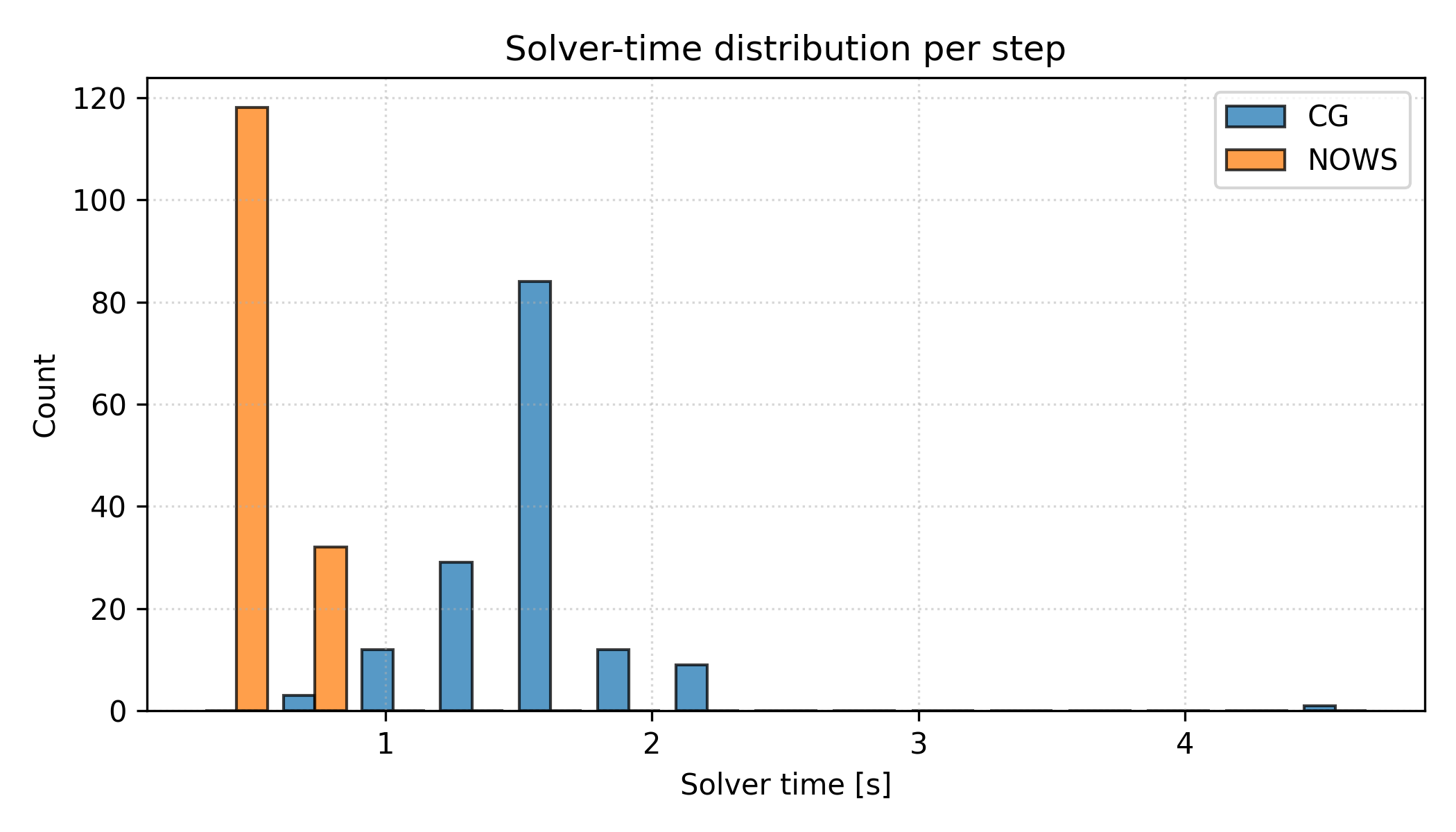}
    \caption{\textbf{Statistical comparison of per-step solver behavior.}
    (\textbf{a}) Iteration distribution per step for CG and NOWS-initialized solvers. The shift toward smaller iteration counts demonstrates the systematic impact of the neural warm start.
    (\textbf{b}) Corresponding solver-time distribution, showing consistent contraction toward lower runtimes.
    Both metrics confirm stable and repeatable acceleration under NOWS initialization.}
    \label{fig:smoke_dists}
\end{figure}

\paragraph{Statistical Distributions.}  
Complementary histogram analyses (Fig.~\ref{fig:smoke_dists}) provide statistical evidence of this acceleration. The iteration distribution shows that NOWS shifts solver behavior toward substantially lower iteration counts (80–200 range) with reduced spread, while the baseline CG solver exhibits a wide distribution centered around 400–600 iterations. The corresponding solver-time histogram mirrors this trend, with NOWS clustering near sub-second runtimes and CG spreading up to 4~s per step. These results confirm that neural initialization consistently moves the solver into a low-cost regime and reduces variability across simulation timesteps.

\begin{figure}[H]
    \centering
    \includegraphics[width=0.49\textwidth]{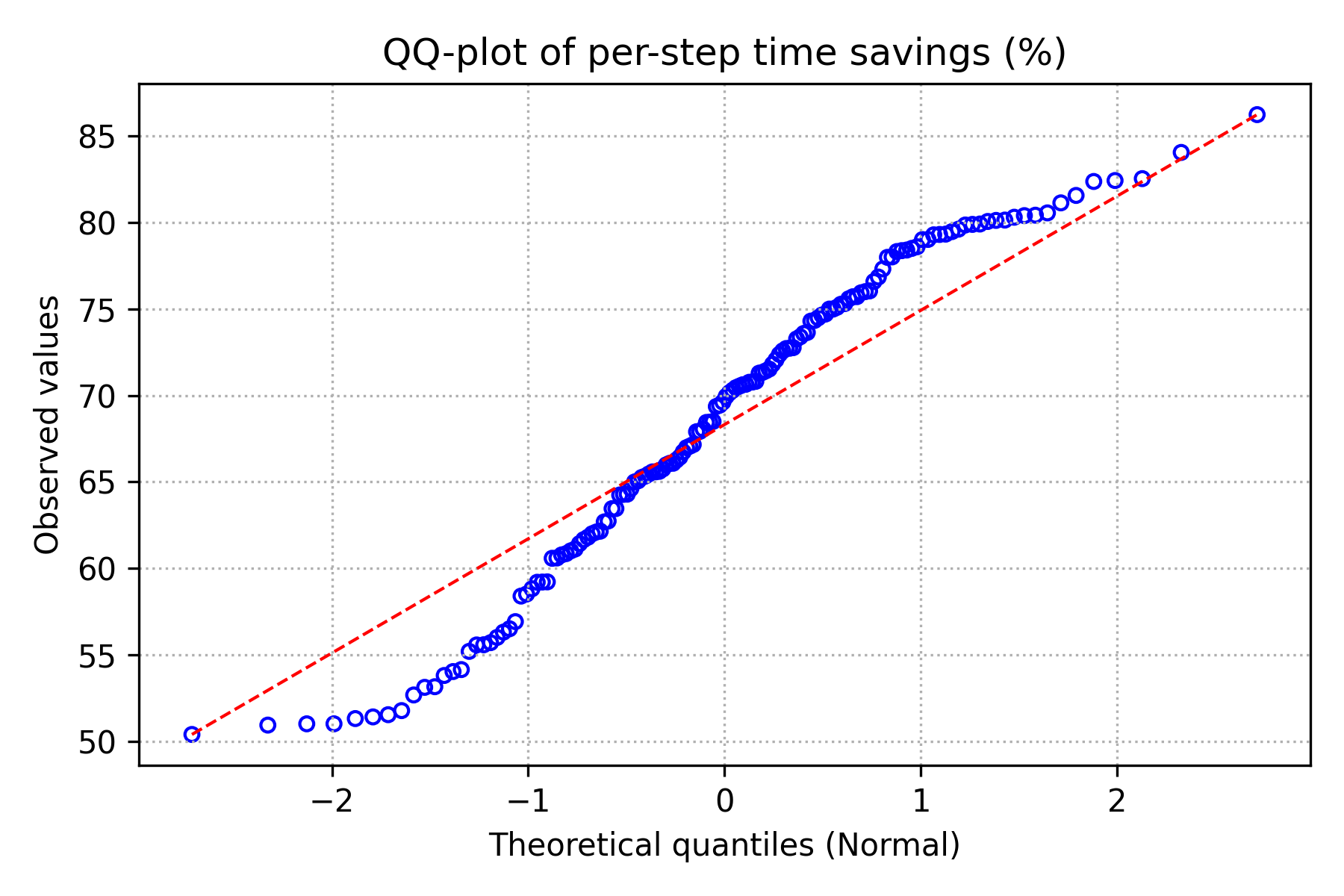}
    \includegraphics[width=0.49\textwidth]{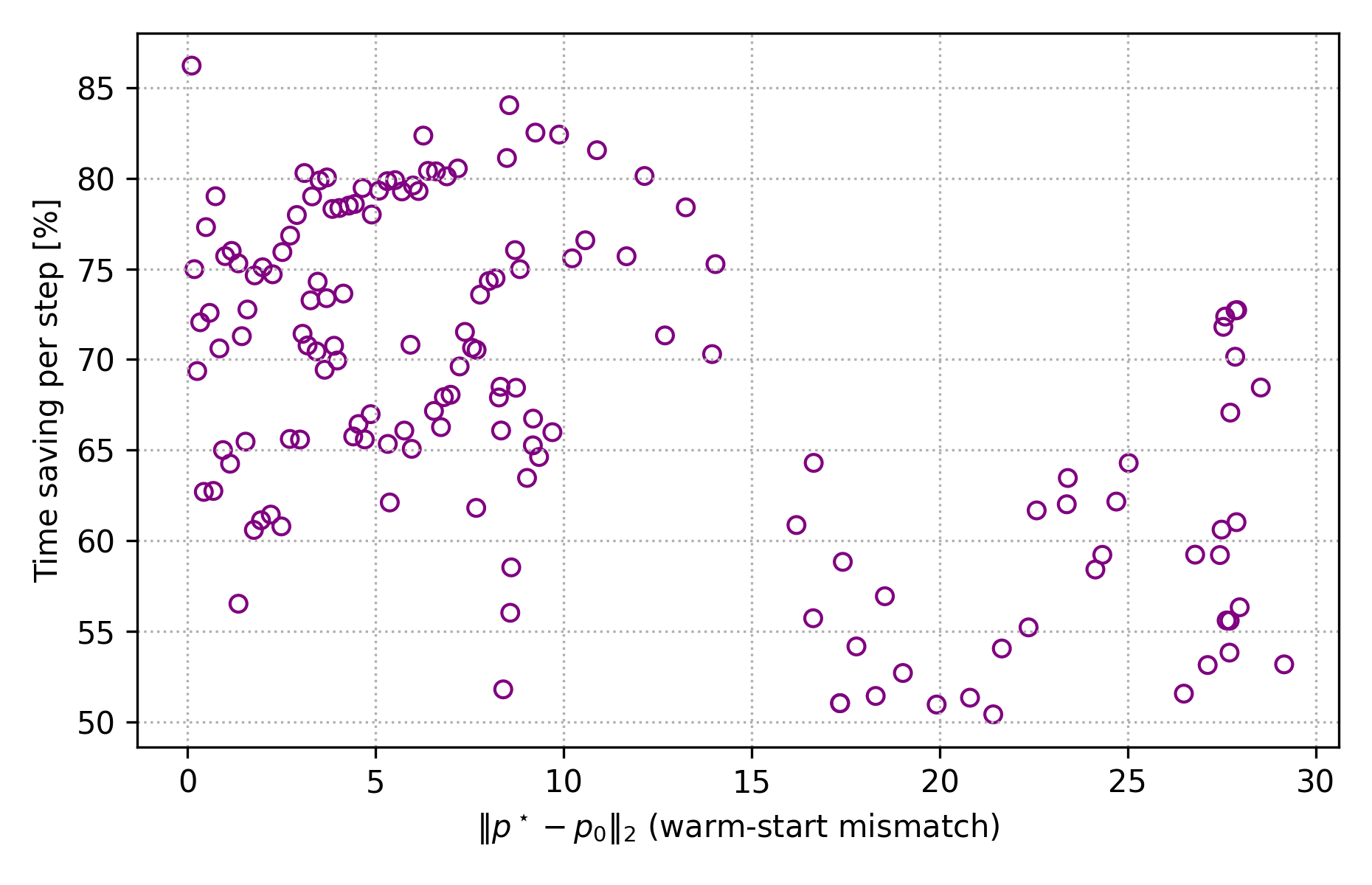}
    \caption{\textbf{Distributional and correlation analysis of NOWS acceleration.}
    (\textbf{a}) QQ-plot of per-step time savings showing an approximately normal distribution centered around 70--80\%, indicating statistically consistent acceleration across the full simulation. 
    (\textbf{b}) Scatter plot showing correlation between warm-start mismatch $\|p^\star - p_0\|_2$ and per-step time saving. 
    Steps with more accurate neural initializations yield higher time savings, demonstrating the direct link between warm-start quality and solver efficiency.}
    \label{fig:smoke_correlation}
\end{figure}

\paragraph{Distributional and Correlation Analysis.}
Beyond aggregate statistics, we examine the distribution and consistency of runtime improvements across timesteps. 
Figure~\ref{fig:smoke_correlation} presents two complementary diagnostics: a QQ-plot assessing the normality of per-step savings and a correlation analysis quantifying the relationship between initialization accuracy and speedup.

The QQ-plot (right panel) demonstrates that the empirical distribution of per-step runtime savings aligns closely with a normal distribution, with most quantiles following the red reference line. This indicates that the acceleration effect of NOWS is statistically uniform across the entire simulation, without heavy-tailed outliers or unstable cases.

The scatter plot (left panel) correlates the warm-start mismatch $\|p^\star - p_0\|_2$ with the achieved time saving per step. 
A clear negative trend is observed: smaller initialization errors correspond to higher time savings, confirming that the neural operator’s prediction quality directly governs solver acceleration. 
Quantitatively, the Pearson correlation coefficient is approximately $r = -0.67$, highlighting a strong inverse relationship between warm-start accuracy and runtime cost.

\paragraph{Resolution-Invariant Acceleration.}  
The learned operator, trained solely on coarse $64\times64$ grids, was directly applied to $256\times256$ simulations without retraining. Across both resolutions, NOWS reduced total solver iterations by over 65\% and runtime by nearly a factor of three. Table~\ref{tab:smoke_scaling} summarizes these quantitative improvements, confirming that the learned operator generalizes seamlessly across spatial discretizations. Together, these results establish NOWS as a resolution-invariant and solver-agnostic strategy for accelerating large-scale, coupled PDE systems.

\begin{table}[H]
\centering
\caption{\textbf{Quantitative summary of scalability and performance for the smoke plume problem.}}
\begin{tabular}{lcccc}
\toprule
Grid Resolution & Solver & Total Iterations & Runtime (s) & Reduction (\%) \\
\midrule
$64\times64$ & CG   & 24,860 & 97.37 & -- \\
$64\times64$ & NOWS & 10,523 & 50.16  & 48.5 \\
$256\times256$ & CG   & 90,199 & 290.38 & -- \\
$256\times256$ & NOWS & 17,783 & 74.75  & 74.3 \\
\bottomrule
\end{tabular}
\label{tab:smoke_scaling}
\end{table}

\end{document}